# GeoJEPA: Towards Eliminating Augmentation- and Sampling Bias in Multimodal Geospatial Learning

**Theodor Lundqvist, Ludvig Delvret**

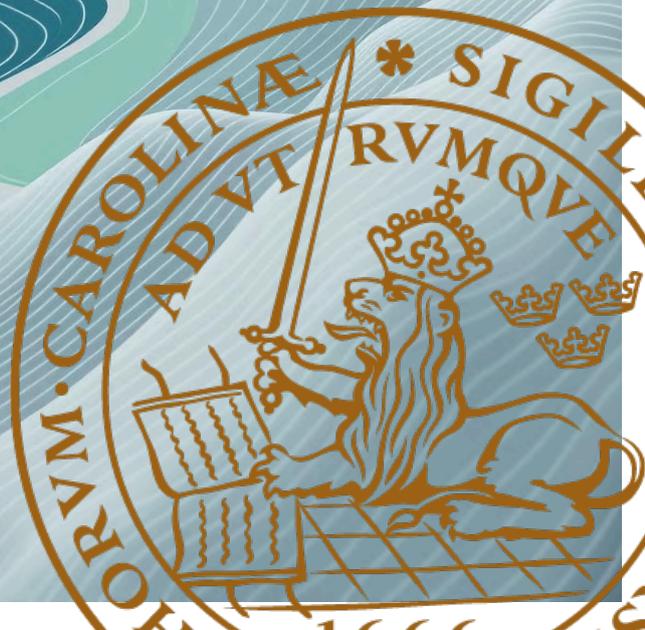

MASTER'S THESIS
Computer Science

LU-CS-EX: 2025-01

# GeoJEPA: Towards Eliminating Augmentation- and Sampling Bias in Multimodal Geospatial Learning

GeoJEPA: Att elimiera modifierings- och urvalsbias i multimodal geospatial inlärning

**Theodor Lundqvist, Ludvig Delvret**

# GeoJEPA: Towards Eliminating Augmentation- and Sampling Bias in Multimodal Geospatial Learning

## Master's Thesis

Theodor Lundqvist

theodor.lundqvist@gmail.com

Ludvig Delvret

ludvig.delvret@gmail.com

January 24, 2025

**Abstract**

Existing methods for self-supervised representation learning of geospatial regions and map entities rely extensively on the design of pretext tasks, often involving augmentations or heuristic sampling of positive and negative pairs based on spatial proximity. This reliance introduces biases and limits the representations' expressiveness and generalisability. Consequently, the literature has expressed a pressing need to explore different methods for modelling geospatial data.

To address the key difficulties of such methods, namely multimodality, heterogeneity, and the choice of pretext tasks, we present GeoJEPA, a versatile multimodal fusion model for geospatial data built on the self-supervised Joint-Embedding Predictive Architecture. With GeoJEPA, we aim to eliminate the widely accepted augmentation- and sampling biases found in self-supervised geospatial representation learning. GeoJEPA uses self-supervised pretraining on a large dataset of OpenStreetMap attributes, geometries and aerial images. The results are multimodal semantic representations of urban regions and map entities that we evaluate both quantitatively and qualitatively. Through this work, we uncover several key insights into JEPA's ability to handle multimodal data.

**Keywords:** Geospatial AI, Joint-Embedding Predictive Architecture, Map Entity Representation Learning, Self-Supervised Learning, Multimodal

# Acknowledgements

First, we would like to thank AFRY and *the Industry Partner*, for their generous support in the form of experienced supervisors, office space, and computing resources. A special thanks to Dan Svenonius at AFRY for always caring about our well-being and giving the best advice.

We want to thank our supervisor at LTH, Dr Jonas Skeppstedt, for his support on the thesis and help in bringing the project to completion, as well as Dr Marcus Klang for his support as a sparring partner on early design ideation.

We would also like to express our heartfelt gratitude to our respective partners, whose unwavering support, encouragement, and understanding have been invaluable for the completion of this thesis.

A special thanks to our close friend, who generously lent us his RTX 3080 graphics card for local development and model evaluation, saving us valuable time and effort. We would also like to give thanks to all our other friends at LTH, whom we have spent countless hours with, and have made these past years some of the best of our lives. We will always cherish these moments and hope to meet all of you in the future.

Finally, we offer our thanks to each other, for the collaboration over the past years, and for the remarkable effort put into successfully concluding this thesis.

*Theodor Lundqvist & Ludvig Delvret*

# Contents

# Chapter 1

# Introduction

*In this chapter, we introduce the problem area, briefly discuss preliminary knowledge, and specify our goals with this thesis.*

## 1.1 Background

*Geographic Information Systems (GIS)* play a vital role in managing and analysing global-scale data in applications ranging from location-based services to the monitoring and forecasting of traffic, air quality, tides, or weather. Despite their utility, these systems face a significant challenge in handling the heterogeneity, size, and ever-changing nature of geospatial data, which encompasses a variety of modalities such as spatial geometry, aerial and ground images, trajectories, check-ins, text, and intricate relationships over long and short distances. To make sense of these enormous volumes of data, the field of *Geospatial Intelligence*, or *GeoAI*, has seen rapid development [1]. While remote-sensing from aerial images can be an effective method for enriching digital map data and e.g. land-cover analysis [2], *Map Entity Representation Learning* (MERL) methods can be used to extract rich semantic information from digital entities such as points of interest (POIs), roads, buildings, and urban regions [1, 3]. As these entities are central to large parts of human activity, their semantics are useful for a wide variety of tasks such as traffic speed prediction, socio-economic analysis, crime rate prediction, or recommendation algorithms [4–8].

Based on the success of large foundation models in language and vision-language tasks, researchers envision a foundation model in the geospatial domain based on self-supervised learning [1, 9–11]. Naturally, the architecture of such a model must handle several challenges, of which unstructured, multimodal data and the reliance on heuristic augmentations and pretext tasks are major ones [1, 11].

## 1.2   Research Problem

The ability to identify map entities, such as places, roads, or areas, from incomplete or imprecise information offers significant potential as a tool for debugging and analysing large-scale geographic information systems. Given the complexity and scale of the data, constructing database field indices to support all possible queries becomes computationally prohibitive.

In the multimodal and heterogenous domain of geospatial features, even defining an appropriate similarity measure to rank potentially matching entities presents a significant challenge [6]. Map entities are commonly comprised of a variety of attributes, including spatial geometry, descriptive key-value fields, and explicit or implicit relationships with surrounding entities, each of which may contribute differently to the concept of similarity. Furthermore, the relevance of these attributes may vary depending on the specific context in which the comparison is applied.

A similar analogy can be observed in the context of words and sentences, where traditional lexicographical comparisons, such as *edit distance* [12], fail to capture semantic meaning. Modern similarity-based text-retrieval applications address these linguistic complexities by utilising embedding vectors from large language models (LLMs), which, as a byproduct of generative pretraining, encode the semantic meaning of words and sentences. As such, *Map Entity Representation Learning*, aims to replicate this for map entities, but often fails to incorporate important data or is constrained by the implicit bias of the chosen pretext task [1, 3, 10].

Recent breakthroughs in machine learning have demonstrated the transformative potential of foundation models across vision, language, and multimodal tasks which have spurred growing demand for foundational models tailored to the complexities of geospatial data, as highlighted in [1, 9–11]. Inspired by this demand and novel self-supervised learning objectives (SSL), this work seeks to address two key challenges in developing such models: (1) the multimodal nature of geospatial data, and (2) biases and generalisation constraints stemming from the choice of pretext tasks and augmentation methods.

## 1.3   Contribution

This thesis introduces a versatile multimodal model called **GeoJEPA**, for **Geo**spatial **J**oint-**E**mbedding **P**redictive **A**rchitecture, that eliminates the need for heuristic sampling and augmentations while providing a modular way to handle complex geospatial data.

With respect to the JEPA self-supervised training objective, we present performance evaluations for multimodal and geospatial data. In addition, we qualitatively analyse the semantics of token-level representations and examine JEPA's capacity to reason over embeddings generated by other models.

# 1.4 Disclaimer

This thesis was conducted in collaboration with a major global technology company that has chosen to remain anonymous. Therefore, specific details about the company, including its name, have been omitted to respect confidentiality. The company is referred to as *the Industry Partner*.

# 1.5 Preliminaries

This section provides a brief overview of OpenStreetMap and feature representations, both of which are essential for understanding the context of our aim and research questions.

## 1.5.1 OpenStreetMap

Founded in 2004, OpenStreetMap (OSM) is a free map built by contributions from volunteers. As a large dataset of unstructured map data with easy access, OSM is often used for academic research and personal projects. The community-driven approach also allows for rapid updates and detailed local insights, particularly in areas underserved by commercial map providers [13]. OSM contains data of three distinct modalities; geometry in the form of polylines, polygons, and multi-polygons, language in the form of key-value tags (such as `[building=yes]`), and graphs in the form of relations. We explore these more thoroughly in Chapter 3.

## 1.5.2 Embedding Vectors

In this section, we give an overview of *Embedding Vectors*, a concept that is key for understanding the subsequent chapters.

Embedding vectors, also known as *latent representations*, *feature vectors*, or *embeddings*, are numerical representations of data produced by a model during training or inference. These representations commonly capture semantic relationships or patterns in the input data. While *latent representations* or *latent features* typically refer to intermediate, non-observable states within a model, *embeddings* generally denote the final representation.

In machine learning, an embedding can be conceptually understood as a direction in some high-dimensional vector space. Numerically, embeddings are represented as lists of floating-point numbers. Given an appropriate encoding scheme, it is possible to represent the similarity of two words, images, videos, or otherwise, as some distance between their embedding vectors. As illustrated in Figure 1, abstract concepts such as sex, plurality, or even royalty, can be captured as specific directions in this space.

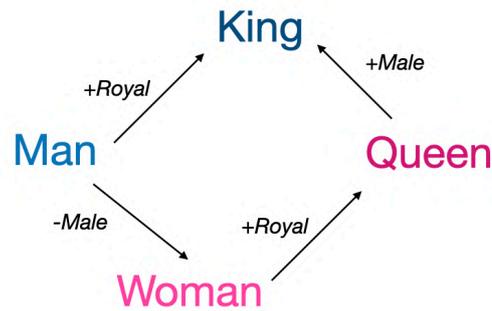

**Figure 1:** A classic example of the semantic relation between word embeddings originating from [14]. In practice, the embedding space occupies a much higher dimensionality than the two shown in the image, but less than the dimensionality of the input data.

These linear properties of embedding spaces are not imposed on models during training, instead, the *Linear Representation Hypothesis* (e.g. [14], [15], [16]) suggest that these properties emerge in performant models due to being computationally efficient.

Additionally, single neurons are capable of encoding multiple unrelated concepts simultaneously, a phenomenon referred to as *superposition* [16]. This occurs when concepts share directional components within the vector space rather than being orthogonal to one another. While this introduces interference, it enables the encoding of an exponentially larger number of concepts within the same representational space.

## 1.6 Aim

This thesis aims to explore and assess self-supervised methods for representation learning in the geospatial domain, specifically in the context of OpenStreetMap (OSM). Furthermore, we aim to explore how the Joint-Embedding Predictive Architecture (JEPA) and other self-supervised learning (SSL) objectives can be applied in this domain. We explicitly define the aim as follows:

*To explore SSL architectures for representation learning of unimodal and multimodal geospatial data.*

## 1.7 Research Questions

The purpose of this thesis is to explore how SSL architectures can be applied to geospatial data by addressing the following research questions:

1. *How can the quality of (multimodal) geospatial representations be evaluated?*
2. *How can SSL architectures be adapted to support unstructured and multi-modal map data?*

# Chapter 2
# Methodology

---

*This chapter aims to explain the methodology employed to address the research questions posed in this thesis. It provides an overview of the literature study, the implementation process, as well as the tools and computational resources used.*

## 2.1 Literature Study

Throughout the work on this thesis, extensive literature studies have been employed to further our understanding of the relevant topics. The studies were done through the use of keyword searches on Google Scholar and references in the selected papers, especially surveys like [1, 9, 17] were used for this purpose. We started with the keywords: *geospatial features*, *geospatial learning*, and *geospatial similarity* before moving on to reviewing different model architectures including *Transformers*, *Graph Neural Networks* and *Self-Supervised Learning (SSL)* as we found it necessary to understand the existing research. Additionally, we have read papers on *polygon representation learning*, *map entity representation learning*, *geospatial region embeddings*, *multi-modal models*, and all existing papers covering the *Joint-Embedding Predictive Architecture (JEPA)*.

## 2.2 Implementation Overview

In this section, we provide a short overview of our implementation process. First, we generate a geospatial dataset spanning approximately 24500 km$^2$ of the United States by implementing efficient algorithms for downloading and processing OpenStreetMap data and aerial imagery from the *National Aerial Imagery Program* (NAIP) [18]. Simultaneously, we develop a quantitative suite of tasks and baseline models. To handle the complex nature of geospatial data, we design a versatile multimodal model named **GeoJEPA**. Finally, we evaluate the models both quantitatively and qualitatively.

## 2.3 Tools

While we can't discuss all software frameworks and tools we used, some are of special importance for this thesis. The `C++` package *osmium* [19] is used to extract OSM data, and the open-source protocol *protobuf* [20] is used for serialization. For machine learning, *pytorch* [21], *pytorch lightning* [22] and *sk-learn* [23] are used. The report is written in *Typst* [24] together with the spell-checker *Grammarly*. The cover image is generated using the `flux-1.1-pro` generative image model [25].

## 2.4 Computing Resources

In total, we utilise four distinct computing setups. Notably, we utilised a desktop PC to rebuild all `C++` code and regenerate the datasets from scratch any time a change was introduced. This desktop was also used for model development and training of all models except GeoJEPA, which is trained on a single A100 GPU of a high-performance computing (HPC) cluster provided by *the Industry Partner*.

**Table 1:** Four different configurations for computation.

| GPU (1x) | RAM | CPU | Type | Task |
|---|---|---|---|---|
| - | 32GB | i9 16-vcpu | Laptop | OSM data processing |
| - | 64GB | Xeon 24-vcpu | Cloud | NAIP image processing |
| NVIDIA 3080 10GB | 64GB | Ryzen 3700x 16-vcpu | Desktop | OSM data processing, model development, and evaluation |
| NVIDIA A100 80GB | 135GB | Xeon 12-vcpu | Cloud | Large-scale model training |

## 2.5 Division of Labor

The work on this thesis has been divided into two main areas; data sourcing and processing led by L. Delvret, and the design and implementation of deep learning models, led by T. Lundqvist. The entire project has been highly cooperative, with frequent brainstorming and discussions in order to overcome obstacles in further development.

# Chapter 3

# Aerial Imagery and OpenStreetMap

*This chapter provides an introduction to the types and structure of geospatial data used in this thesis.*

## 3.1 Scope

Geospatial data can be defined as any data tied to the surface of the earth by coordinates, addresses, or other means [26]. Since "everything happens somewhere" [27], it is meaningful to define a relevant subset. For practical reasons, this thesis focuses on utilising map data from *OpenStreetMap* (OSM) [13] and aerial imagery from the *National Aerial Imagery Program* (NAIP) [18]. While other relevant data types, such as trajectories, ground images, and georeferenced text have been explored in previous research and may offer additional insights, they have been excluded from the scope of this study. A compilation of the modalities used in existing models is presented in Section 4.1.3, and further justification for the use of these datasets is provided in Section 5.1.

## 3.2 Aerial Imagery

There are two alternatives for top-down imagery, satellite and aerial. While satellite imagery contains substantial information, we use aerial imagery as it provides finer details and is freely available from e.g. NAIP [18]. The highest resolution open satellite imagery is from *Sentinel-2* [28], which, at 10m/pixel, does not compare favourably to the 0.6m/pixel imagery from NAIP, as seen in Figure 2.

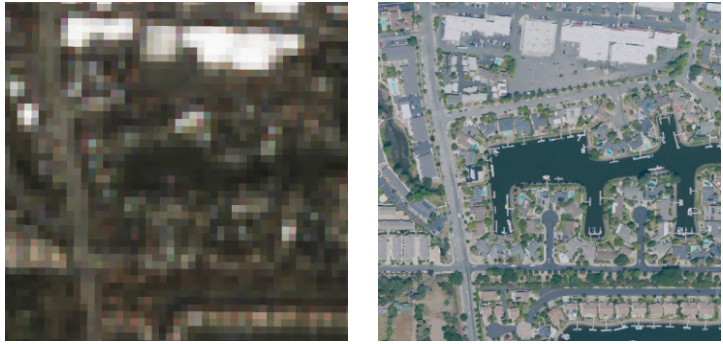

**Figure 2:** Satellite imagery from *Sentinel-2* (left) and aerial imagery from *NAIP* (right).

# 3.3 OpenStreetMap

In this section, we describe the structure of data available in OSM, namely nodes, ways, and relations. Additionally, each such feature may have descriptive text features referred to as tags.

## 3.3.1 Tags

Text-based information describing map entities is stored in key-value pairs called tags. As illustrated in Figure 3 and Figure 4, tags can contain many types of information, including road type and surface quality. Due to OpenStreetMap's "Any tags you like" policy [29], contributors can add their own tags, resulting in an unstructured and highly diverse dataset. A list of the most widely used tags can be found at the OpenStreetMap wiki [30].

## 3.3.2 Nodes

OSM nodes represent single points on the map. They can be part of larger structures, describe a singular point of interest such as a restaurant, or an item, such as a bench or a tree.

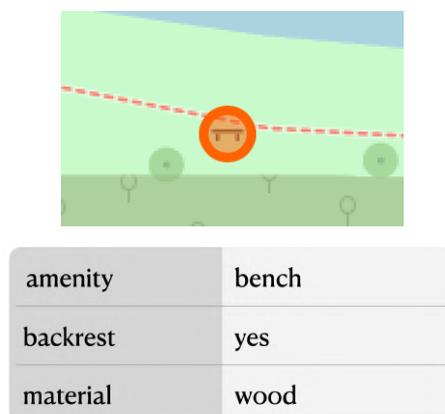

| amenity  | bench |
|----------|-------|
| backrest | yes   |
| material | wood  |

**Figure 3:** A bench-node with descriptive tags.

### 3.3.3   Ways

Everything in OSM that is not modelled as a single node is stored as a way. As shown in Figure 4, a way can be either closed, representing an area or a building, or open, representing something like a stretch of road or part of a larger feature.

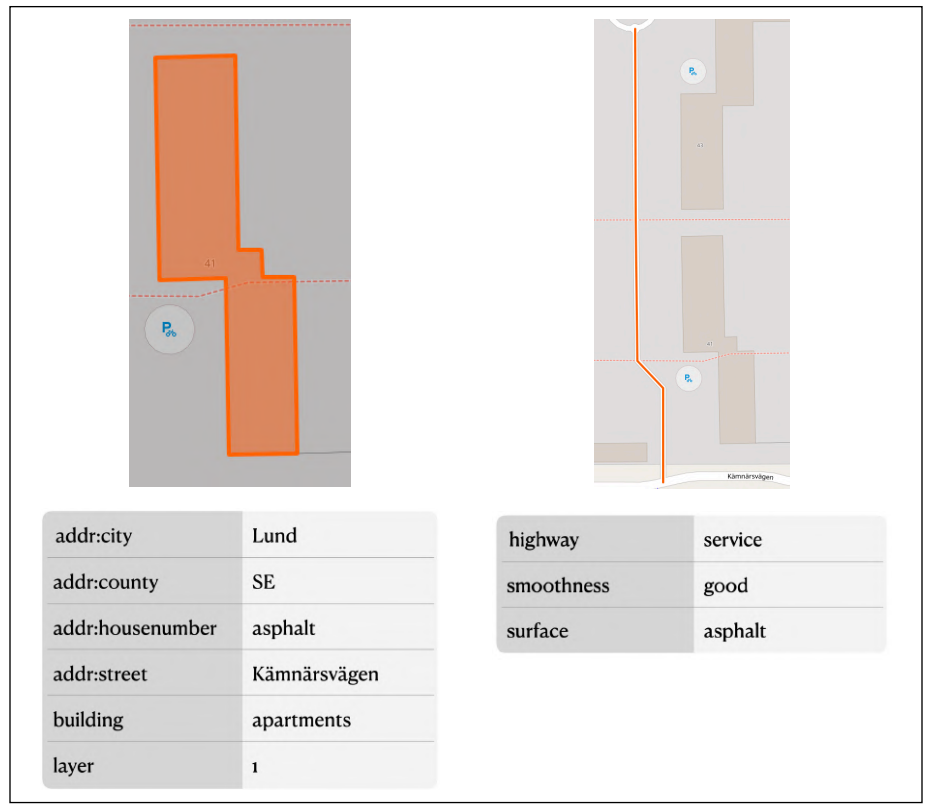

| | | | | |
|---|---|---|---|---|
| addr:city | Lund | | highway | service |
| addr:county | SE | | smoothness | good |
| addr:housenumber | asphalt | | surface | asphalt |
| addr:street | Kämnärsvägen | | | |
| building | apartments | | | |
| layer | 1 | | | |

**Figure 4:** A closed way, or polygon, describing the outlines of a building (left) and a way, or polyline, representing a stretch of road (right).

### 3.3.4   Relations

Sometimes, it is not possible to use a single way to accurately represent a geospatial feature. In those cases, multiple ways are combined into a relation. This allows for the modelling of hollow structures as seen to the left in Figure 5. It is also possible to generate larger features by combining more than one distinct feature. Notably, relations can represent massive networks, such as a bus line or entire city borders, as illustrated to the right in Figure 5.

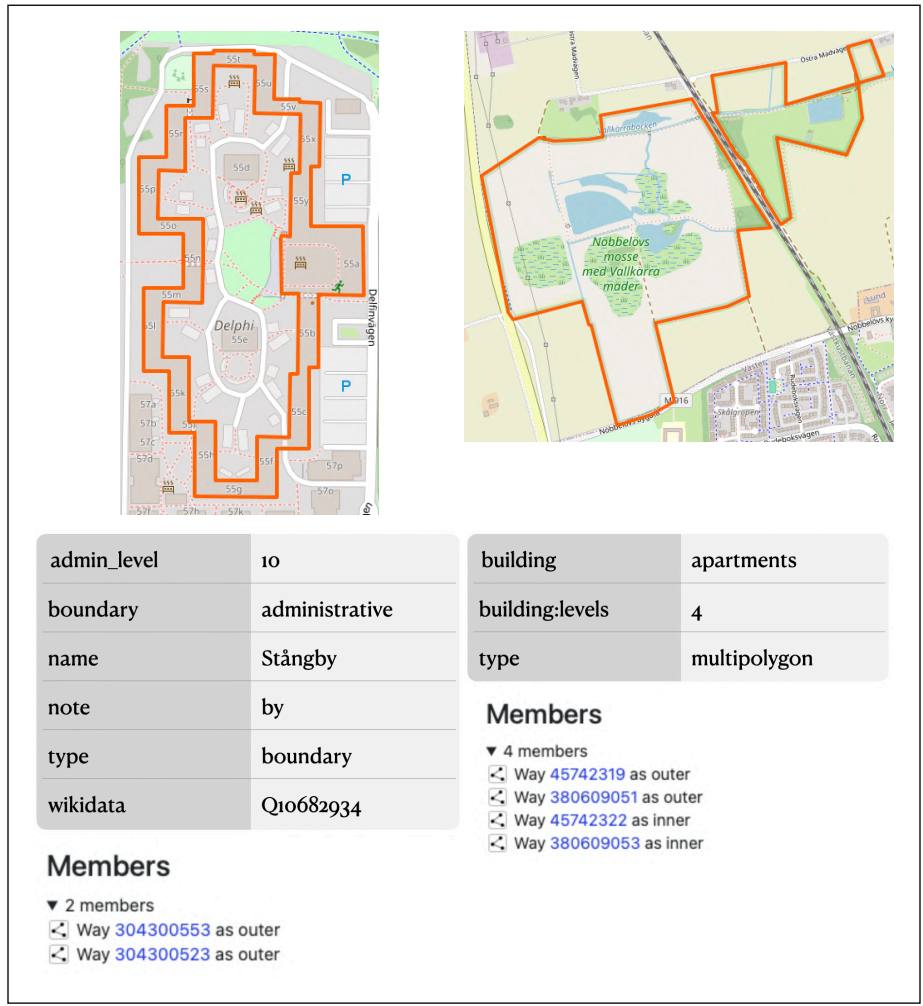

**Figure 5:** Four polylines/ways grouped together in a relation to form a building and its courtyard (left) and two distinct areas combined into a single relation, split into two parts by a railway track (right).

# Chapter 4
# Theory and Previous Work

*In this chapter we review the academic research of foundation models, multimodal architectures, self-supervision, and geospatial representation learning.*

## 4.1 Geospatial Intelligence

Geospatial intelligence, or geospatial AI, has been enabled by the unprecedented availability of geospatial data due to digitalisation and smartphone use. Geospatial intelligence leverages machine learning techniques to understand and predict patterns in geospatial data [1, 31].

### 4.1.1 Map Entity Representations

Map entities, in the form of points of interest (POIs), polylines and polygons, are fundamental components of geospatial data. They form the backbone of digital maps and location-based services and can provide valuable insight into human activity [1, 3]. Semantic representations of individual map entities can facilitate advanced methods of categorisation, recommendation and search. Additionally, they can be used as valuable input to algorithms that validate and enrich digital maps by inferring attributes such as lane number or the average speed as exemplified by [32–35]. Recently, a notable advancement in modelling the full spectrum of OpenStreetMap (OSM) data types, including the often-overlooked relation data type, was introduced with CityFM [10]. Leveraging BERT [36] for embedding OSM tags and ResNet [37] for visually capturing geometric features. CityFM combines these two modalities into a shared embedding space using a multi-layer perceptron. By using spatial proximity as a heuristic for entity similarity and carefully selecting three contrastive learning objectives (text, visual geometry, road-based) the model learns entity and region representations useful for predicting traffic speed, building functionality and population density.

## 4.1.2 Urban Region Representations

Urban regions provide an effective scale at which demographic and socio-economic trends may be analysed and predicted (e.g. [1, 4–8, 38–41]). Since the separation of cities into regions can be done arbitrarily and on multiple scales, regions themselves often do not hold any semantic information but are instead some sum of the entities in and around them. Therefore, incorporating a diverse set of data sources such as human mobility, POIs, road networks, geometries, images and relations, can improve the analytical opportunities.

Learned representations of urban regions have been used for a variety of downstream tasks such as predicting population density [8], housing prices [1, 5, 8], crime rates [4–6], delivery volumes [1], and air pollution [6, 7, 11, 34]. Beyond prediction tasks, region representations are very useful for identifying and clustering similar regions, enabling knowledge transfer in decision-making, pollution control, business site selection or recommendation algorithms [6–8].

## 4.1.3 Data Types in Existing Methods

Existing approaches for self-supervised Geospatial AI commonly employ custom pretext tasks and models that integrate with one or multiple data types to learn embeddings of a single type of data. Recently, models have emerged that learn embeddings of multiple types of geospatial data simultaneously, for instance, CityFM [10] and HOME-GCL [3]. Without going into detail on each relevant type of data, comprehensive tables of existing approaches are presented by Chen et al. [1] from which we have compiled a holistic overview of used data types and multimodal capabilities, seen in Table 2.

**Table 2:** Data type statistics compiled from [1], using similar terminology. Each column represents one embedding target data type and each row represents one input modality. For instance, six of the 26 urban region embedding models utilise aerial or satellite imagery.

|  | POI | Trajectory | Road Networks | Regions | Multi-type | Total | Ours |
|---|---|---|---|---|---|---|---|
| POI Attributes | 17 | 4 | - | 14 | 2 | 37 | x |
| POI Neighbours | 12 | - | - | 14 | 2 | 28 | x |
| POI Check-in Sequences | 6 | 4 | - | 1 | - | 11 | - |
| POI Co-query Context | 2 | - | - | - | - | 2 | - |
| Road Geometry/Graphs | - | 12 | 13 | - | 2 | 27 | x |
| Road Attributes | - | - | 5 | 1 | 2 | 8 | x |
| Trajectories | - | 28 | 5 | 16 | 1 | 50 | - |
| Knowledge Graph | 2 | - | - | 1 | - | 3 | - |
| Aerial or Satellite Imagery | - | - | - | 6 | - | 6 | x |
| Ground Imagery | - | - | 2 | 2 | - | 4 | - |
| Land Use | - | - | - | 2 | - | 2 | - |
| Knowledge Graph | - | - | - | 2 | - | 2 | - |
| Building Footprints | - | - | - | 2 | 1 | 4 | x |
| Land Parcels | - | - | - | - | 1 | 4 | - |
| Total | 17 | 28 | 13 | 26 | 2 | 85 | - |

### 4.1.4 Summary of Geospatial Representations

Existing frameworks for region and map entity representation learning employ contrastive or predictive learning and rely on carefully selected pretraining tasks and heuristics that shape the versatility of the resulting representations ( e. g. [7, 8, 40, 42–44]). Consequently, recent works utilise multiple predictive or contrastive pretraining objectives to create richer representations of multiple data types (e.g. CityFM [10], HOME-GCL [3]).

Given that the reliance on heuristics and the selection of pretext tasks or augmentation strategies significantly impact model performance and applicability to downstream tasks, Chen et al. [1] call for systematic research into the design of self-supervised pretraining objectives for modelling diverse and multimodal geospatial data.

## 4.2 Foundation Models

Foundation models are trained on large-scale data without any task-specific objective. The qualities of these models are *emergent* in the sense that their behaviour and abilities are not explicitly constructed. Instead, the large scale of the model and the training data enables generalisation to unseen tasks. [9, 45] As such, foundation models have reached massive success and enable modern generative AI in language and vision tasks. However, these models fall short in geospatial intelligence due to domain gap, the inherent multimodal nature of geospatial data, and the lack of large-scale training data [1, 2, 46, 47].

## 4.3 The Transformer

The transformer architecture serves as the foundation for a significant portion of the work presented in this thesis. As such, gaining a high-level understanding of its principles is essential.

The transformer model, famously introduced by Vaswani et al. [48] is an extraordinarily strong sequence modelling architecture. In contrast to previous architectures (e.g. RNN and LSTM) the transformer utilises *self-attention* to consider all tokens (e.g. words) simultaneously. The transformer encoder module outputs the same number of tokens that is fed to it, before which, each token "attends" to all other tokens in the sequence, updating its representation based on the contextual information provided by the surrounding tokens. Conceptually, this process can be illustrated through a common example. Consider the word "bank", which can have multiple meanings depending on its context. In the phrase "river bank," the token corresponding to "bank" will attend to the preceding token "river" to narrow its meaning. Conversely, in "financial bank," the token "bank" will attend to "financial", refining its interpretation in another way. While the architecture is built to mimic this, other processes that are more efficient, or beneficial in any way, may emerge during training (e.g. *register tokens* [49]).

In this manner, transformer-based architectures reach state-of-the-art performance for the modelling of language (e.g. BERT as shown in Figure 6) and images, videos, audio and time-series data.

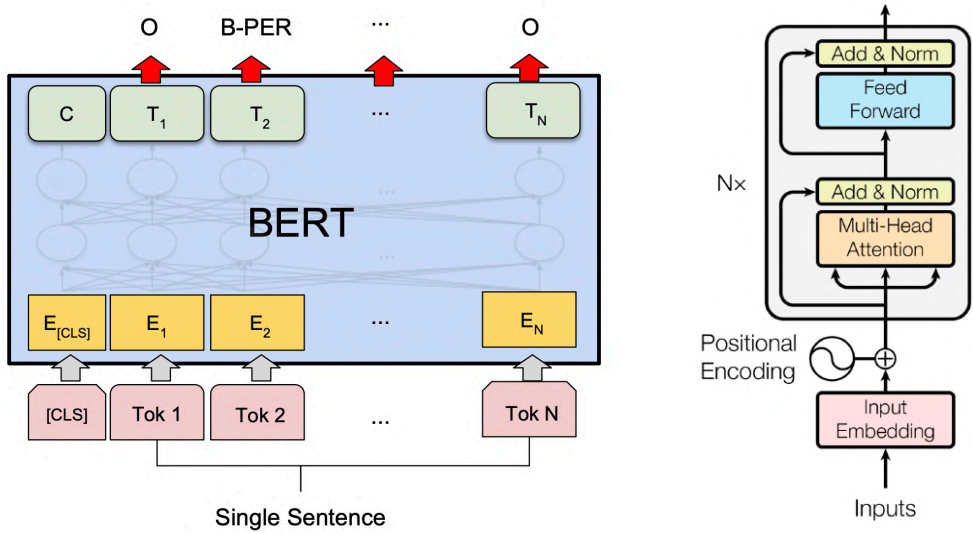

**Figure 6:** Named entity recognition with the BERT transformer model from [36] (left), and the original transformer encoder architecture adapted from [48] (right).

## 4.4 Multimodal Architectures

A modality refers to a specific type of data or sensory information, such as images, text, audio or video. Multimodal models are designed to process information from multiple data types simultaneously, enabling richer understanding and better flexibility. However, building multimodal models is challenging due to the inherent differences in data representation and structure across modalities. Text is sequential and discrete, images are spatial and continuous, and audio is temporal and waveform-based, making labeling, model architecture and positional alignment complex. The computational complexity of handling and interconnecting multiple data streams further complicates things. Major advancements have been made following the advent of the transformer, whose representational power allows for flexible modelling of distinct data modalities. A study of the literature suggests that there are two common approaches, the dual-stream approach; utilizing cross-attention between different modalities with separate weights per modality, and the single-stream approach; concatenating distinct modality tokens to the same transformer encoder stream with shared weights between modalities [17, 50–56].

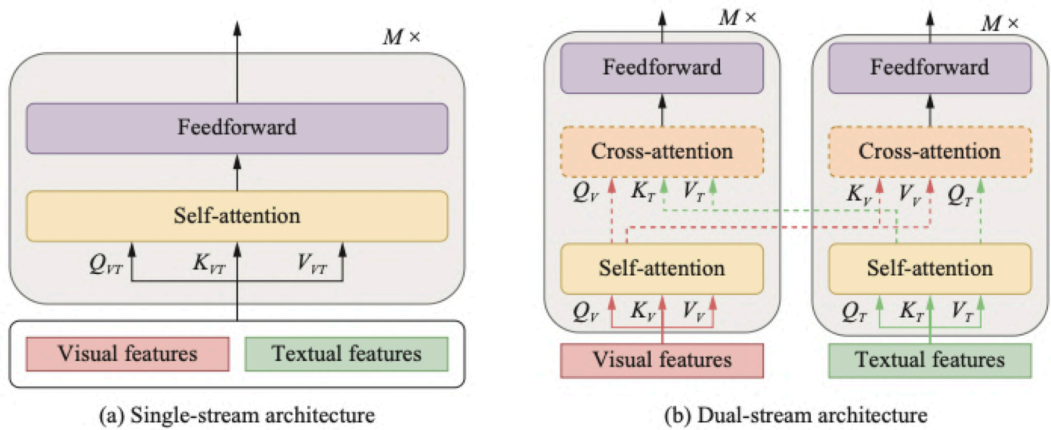

(a) Single-stream architecture  (b) Dual-stream architecture

**Figure 7:** Two separate categories of multimodal transformer architectures, illustration from Chen et al. [17].

While single-stream architectures are parameter efficient, dual- or multi-stream architectures have the benefit of parallel processing. Also, some authors suggest that dual-stream models can benefit from being able to process modalities separately as they may require very different handling. However, the literature do not show significant performance differences between the two architectures (e.g. [56]).

# 4.5 Self-Supervised Learning

Traditional machine learning depends on supervised training which requires expensive labeled datasets. In an attempt to mitigate this, unsupervised or self-supervised learning techniques (SSL) have emerged as powerful alternatives, enabling learning rich representations from large volumes of unlabeled data.

## 4.5.1 Generative Architecture

For both images and language, generative self-supervision has been shown to be a very strong learning objective. In short, generative methods aim to reconstruct pixel or word-level details from incomplete data. An example of this is the autoencoder [57] which learns how to compress input data into a latent representation and then reconstruct the original data from the compressed form. The learning signal is obtained by comparing the pixel values of the two images. Lately, the masked autoencoder has been a more flexible alternative that predicts the missing word in a sentence or the missing pixels of an image [58]. Non-trivially, the ability to predict missing words or pixels requires a deep understanding of the input data and therefore constitutes a powerful pretraining objective.

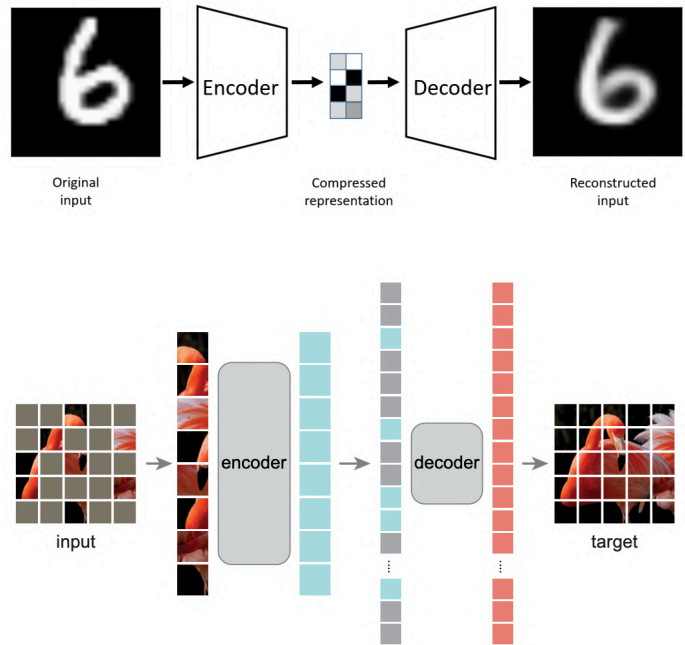

**Figure 8:** An example of an autoencoder (top), and a masked autoencoder (bottom). The illustrations are borrowed from [57] and [58] respectively.

## 4.5.2 Joint-Embedding Architecture

Joint-embedding architecture (JEA) refers to a class of self-supervised models that operate directly in latent space. These models optimise a loss function designed to minimise the latent distance between similar input samples. Similar samples are chosen by some predefined or heuristically guided criteria. In image models, similar samples are typically synthesised by adding noise or augmenting the original image in one or multiple ways [59], while for geospatial data, it is more common to use spatial proximity as a heuristic for similarity (e.g. [10]). Since no potentially complex reconstruction loss has to be defined, this architecture is perfect for aligning the semantics of different modalities, as shown in e.g. CLIP [60].

However, by bringing all similar samples closer together, the model can learn to predict a fixed output for all input samples, resulting in what is known as *representation collapse*, where all latent representations are close to each other but hold no meaningful information. This can be prevented with e.g. *Contrastive* learning, where dissimilar samples are pushed away from one another. The dissimilar samples can be sampled randomly or using a heuristic. Many successful models build on these principles, such as VICReg [61], SimCLR [62], DINO [59], and CLIP [60].

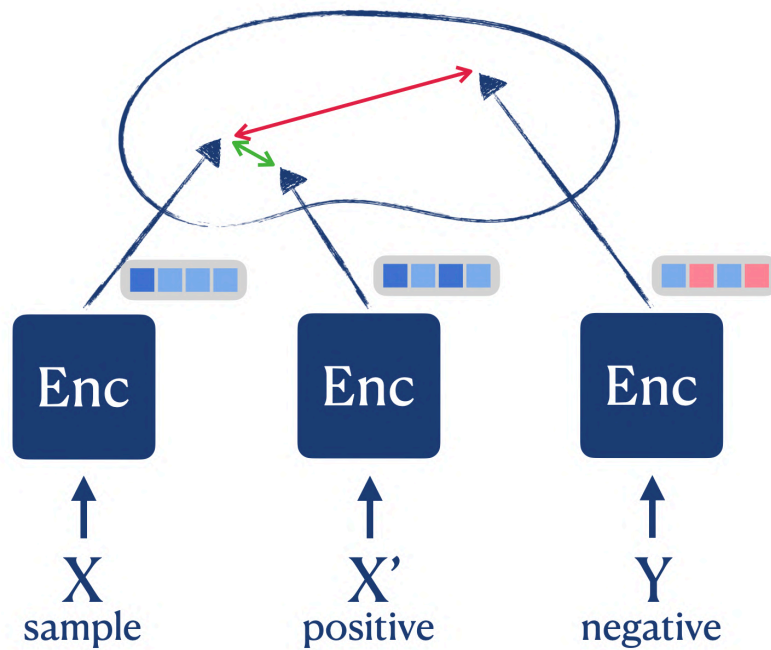

**Figure 9:** Illustration of a contrastive training objective. Augmentation of $X$ is used as a positive sample (attract) and a random sample $Y$ is used as a negative sample (repel).

## 4.5.3 Joint-Embedding Predictive Architecture

Theoretically introduced by Yann LeCun in 2022 [63], and practically, with the advent of the self-supervised image representation model I-JEPA by Meta in 2023 [64], the Joint-Embedding Predictive Architecture (JEPA) aims to address the limitations of generative and view-invariant models. LeCun argues that generative models spend unnecessary compute on reconstruction and focus too much on low-level features [63, 64]. Additionally, reconstruction imposes significant challenges in domains where decoding latent vectors into observable data and computing a loss is non-trivial. Furthermore, LeCun acknowledges contrastive and view-invariant models, noting their dependency on augmentations and vulnerability to representation collapse [63, 64]. Desai and Johnson further argue that contrastive learning presents an extremely sparse learning signal, demanding large batch sizes and extensive datasets to achieve effective training [65].

Instead, JEPA combines joint-embedding and generative approaches by masking and reconstructing latent space variables. The architecture relies on three core components: a context encoder, a target encoder, and a decoder, also commonly referred to as a predictor. In I-JEPA, the context encoder encodes a small fraction of the image. The decoder network predicts the latent representation of another part of the image from the encoded context. An illustration of this process would be to encode the image of a dog's head and use it to predict the latent representation of the dog's body. Providing a higher-level and potentially more efficient objective than predicting pixel-level details.

The target encoder is an exponential moving average (EMA) of the context encoder, which can be interpreted as a form of model ensembling that smooths out the optimisation trajectory [66], contributing to knowledge distillation and making the target encoder "smarter" than the context encoder [59].

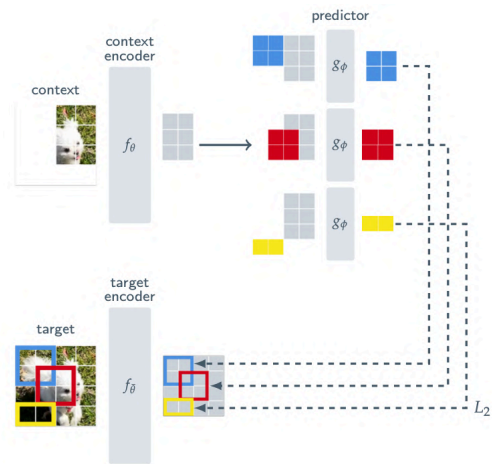

**Figure 10:** JEPA architecture overview from the original implementation by Assran et al. [63].

For the sake of completeness, BYLO [67] also performs prediction in latent space and avoids the reliance on contrastive learning, but still uses view-invariant augmentations, differentiating it from JEPA.

## 4.6 JEPA Research Review

As exploring how self-supervised learning objectives, and especially the Joint-Embedding Predictive Architecture, can be adapted to geospatial data is one of the main objectives of this thesis, we present a holistic overview of the work conducted on the subject.

With the very recent introduction of JEPA, the available research on the architecture is limited. For the purpose of this literature study, we made sure to scan Google Scholar thoroughly after any papers mentioning JEPA and compile the results in Figure 11 showing the development of the field.

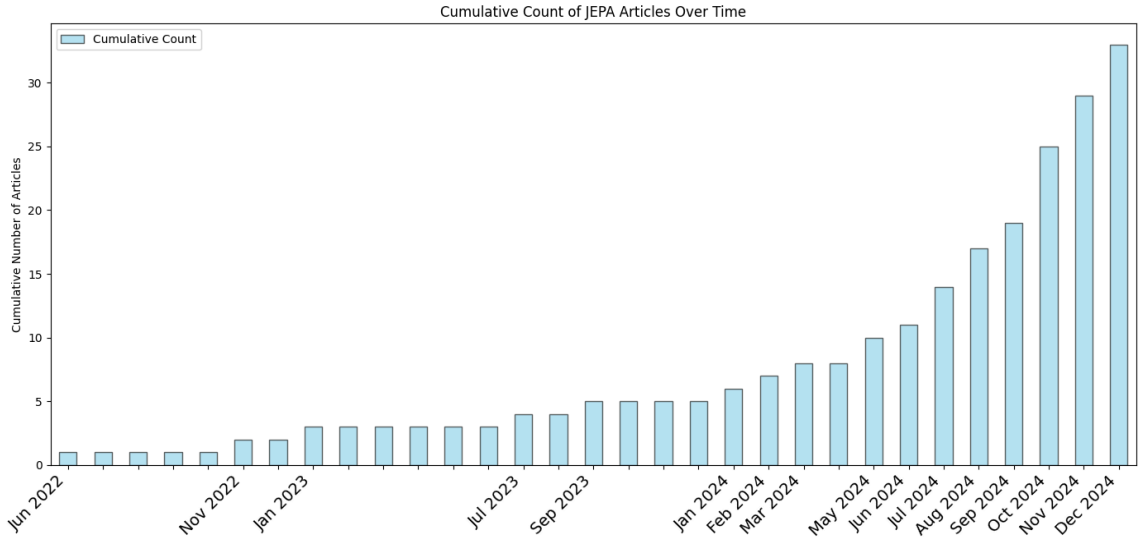

**Figure 11:** Total number of JEPA papers available on Google Scholar. For comparison, papers (with available code) mentioning contrastive learning is 2700+ [68], autoencoders 4700+ [69], and masked autoencoders 100+ [70].

Even with the limited available material, the current literature suggests that JEPA is a particularly fast and flexible learning objective, requiring minimal adaptation to modell new complex data types and proving competitive and compute efficient results in image encoding [64, 71–75], image generation [76, 77], video encoding [78–81], audio encoding [82–84], time-series [85, 86], point cloud [87, 88], tabular [89], and other domain specific adaptations [66, 90–97].

## 4.6.1 Masking in JEPA

With the first implementation using JEPA, Assran et al. [64] introduce the multi-block masking strategy and show that it is superior to random masking. Subsequent authors have presented ablation testing results on masking strategies, where the overall sentiment indicates that a high masking ratio of 75-90% is beneficial [64, 66, 72, 75, 79, 80, 88], depicted in Table 3.

**Table 3:** Performance ablation example on mask ratio, showing that masking a higher fraction of the data yields better results, values from [75]. The authors present multiple evaluation metrics with similar results. For simplicity, we only display the classification score using the k-nearest neighbour (kNN) algorithm.

| Mask Ratio | Evaluation Performance (kNN ↑) |
|:---:|:---:|
| 0.5 | 7.6 |
| 0.75 | 44.5 |
| 0.85 | 61.7 |
| 0.90 | **63.3** |
| 0.95 | 51.3 |

A. Kalapos and B. Gyires-Tóth [72] (CNN-JEPA) support the finding that multi-block masking performs better than random masking and additionally find that a mixed strategy with 25% random masking performs marginally better than multi-block masking as shown in Table 4.

**Table 4:** Masking strategy ablation example, showing that a mixed masking strategy perfoms better than any one of its parts, table from CNN-JEPA [72]. Again, we choose to only display the kNN score. Note that this paper and model architecture is different from the one presented in Table 3. However, the same dataset is used, meaning that scores are comparable.

| Mask Strategy | Evaluation Performance (kNN ↑) |
|---|---|
| Random | 36.64 |
| Multi-block | 55.62 |
| Mixed | **56.26** |

In order to create a prediction task in latent space that is neither trivial nor impossible, it is very important to choose an appropriate masking ratio. However, there is no guarantee that these ratios are transferable to other domains as they are dependent on the correlation between input tokens [75].

## 4.6.2 Training Regime

While JEPA is presented as an architecture that is not vulnerable to representation collapse [63, 64], others find that during training, JEPA will first reach a collapsed state before converging [75, 89]. As such, some authors find that e.g. variance-covariance regularisation (VICReg) [78, 81] or sample discrimination loss (InfoNCE) [75] brings benefits in escaping collapse and improves overall performance by trying to maximise the information in the representations.

An additional challenge of JEPA training is that the in-training latent space reconstruction loss does not reflect the quality of the model. Consequently, authors monitor metrics such as variance and token similarity to assess the performance during training. To better understand the training of JEPA models, Thimonier et al. depict the latent space during different key periods of training [89], which can be seen in Figure 12.

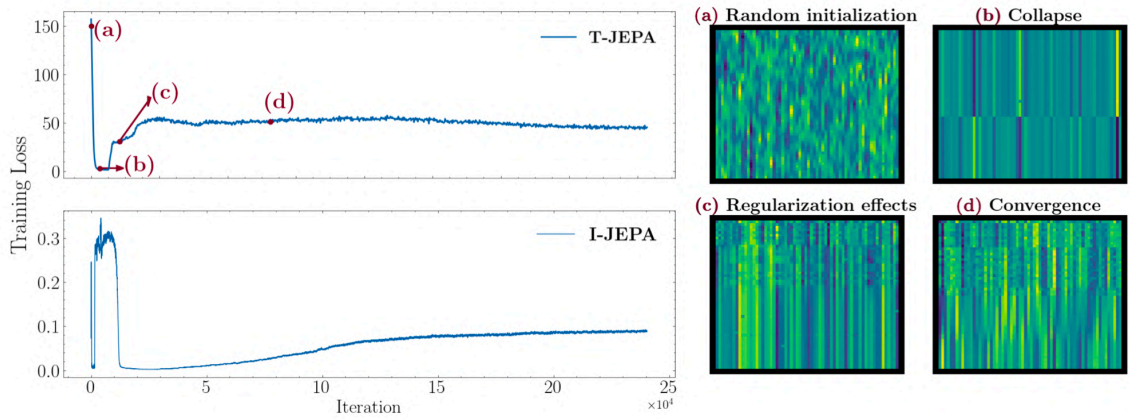

**Figure 12:** Illustration of JEPA loss during training from Thimonier et al. [89]. After initialisation (a), the model collapses (b), and bounces back (c), before converging (d). On the right, the latent space features are plotted. Each row corresponds to the representation of one token.

For performance results, Wei et al. show that the choice of loss function can have a large impact on model performance [75], as can be seen in Table 5.

**Table 5:** Performance ablation on loss function in image modelling, values from Wei et al. [75].

| Loss Function | Evaluation Performance (kNN ↑) |
|:---:|:---:|
| MAE ($L_1$) | 15.2 |
| MSE ($L_2$) | 18.8 |
| Huber (Smooth$L_1$) | 24.3 |
| PatchDisc (InfoNCE) | **30.0** |

To summarise, the choice of masking strategy and ratio, as well as the loss function and potentially other hyperparameters, greatly influence the performance of the JEPA pretraining objective.

# Chapter 5

# Data Sourcing and Pre-processing

*The training of machine learning models is reliant on two main resources: access to a sufficiently large, diverse, and high-quality dataset, and the availability of high-performance computing. In this chapter, we detail the methods employed for acquiring and processing data to meet the requirements of the study. We also present some key insights from our inspection of OpenStreetMap.*

## 5.1  Justification

The geospatial domain is inherently diverse, spanning various data sources, modalities, and spatial scales, creating significant challenges in selecting the optimal modelling strategy. Previous studies frequently employ custom datasets, ranging from tiled representations to graph structures, multi-scale configurations, temporal sequences, and various combinations of modalities to address specific research questions. While we argue that using unique approaches like these is justified in the absence of a universally accepted methodology for geospatial data modelling, it makes direct comparison of methods significantly complicated. Given these challenges, a comprehensive evaluation framework that can accommodate these heterogeneities is needed. However, this too falls outside the scope of this study.

Furthermore, datasets used in previous research are not generally made publicly available, and we are unable to identify one that aligns with our needs in terms of scale and modalities. Consequently, we justify the creation of yet another geospatial dataset.

Both OpenStreetMap (OSM) and the National Aerial Imagery Program (NAIP) offer large-scale, freely available datasets, with OSM itself providing multimodal data in the form of textual tags, geometries, and relational graphs. Combined with NAIP imagery, we can effectively construct a highly multimodal dataset. While this approach continues the trend of introducing new datasets and modelling strategies, it does allow us to explore without constraints.

In summary, we justify constructing a custom dataset, along with tailored tasks, reference models, and evaluation frameworks, to enable rigorous and controlled exploration of JEPA as a multimodal self-supervision framework for geospatial AI, even though it limits external validity.

## 5.2   Sourcing Map Entities

Initially, a data mining solution using the Python package `OSMnx` [98] to query the OSM Overpass API was explored. However, generating a dataset for a 45 km² area required more than four hours. Instead, we find that sourcing the entire OSM dataset from geofabrik.de in protobuf-format and processing it manually in `C++` enables the creation of significantly larger datasets.

## 5.3   Inspection of OpenStreetMap Entities

The purpose of this thesis is rooted in the research problem of identifying semantically similar map entities. Upon our inspection of the OSM dataset, we conclude that most map entities contain little information. These seemingly uninteresting entities can be small segments of roads, or rectangular houses. The houses have the tag `[building=yes]` and occasionally some extra information like the building's address or height. Since a single block can contain over 50 of these buildings, finding a similar house is not very useful. The same is true for road segments, where a single feature can be four meters of a roundabout with the tag `[surface=asphalt]`. These examples illustrate that most entities when isolated, provide little semantic meaning. With this inspection, we come to the conclusion that the semantics of an entity is highly dependent on its neighbours and other surrounding features, i.e. the *context*.

## 5.4   Tiling: A practical choice

To process the data in a way that allows for the efficient incorporation of contextual features, we consider saving each entity using the $X$ nearby entities as context, but ultimately reject it for a few reasons. Firstly, with a contextual area that varies in size and shape depending on the spatial layout and density of entities, including e.g. aerial imagery, as a contextual addition, becomes non-trivial. Secondly, the approach would entail consuming each entity as many times as the context window size.

To address these issues, we select a spatial tiling strategy. While suboptimal for entities near, or intersected by, the tile border, it is efficient and simple to implement. We choose a tiling size of 300 x 300 meters to balance the inclusion of sufficient contextual information with computational efficiency. The tiling approach ensures that each entity is processed only once, except for entities which span multiple tiles. Additionally, we get perfectly deterministic grouping, allowing for reproducible results. However, a drawback is that the number of entities in a tile may vary, necessitating high degrees of padding for batched processing, degrading training and inference efficiency. The complete tiling implementation is presented in Section 5.5.

## 5.5 Regional Tiling

To process OSM data in the protobuf format, the `C++` tool osmium is used. During the processing, features are sorted into tiles based on the tile grid defined by TomTom [99]. This gives consistent 300x300m tiles, avoiding the rectangular tiles caused by Mercator projection of lat-lon coordinates. As seen in Figure 13, it is necessary to perform clipping to only store relevant information. In addition to removing any nodes that are located outside the desired perimeter, it is important to make sure that any affected polygons are closed.

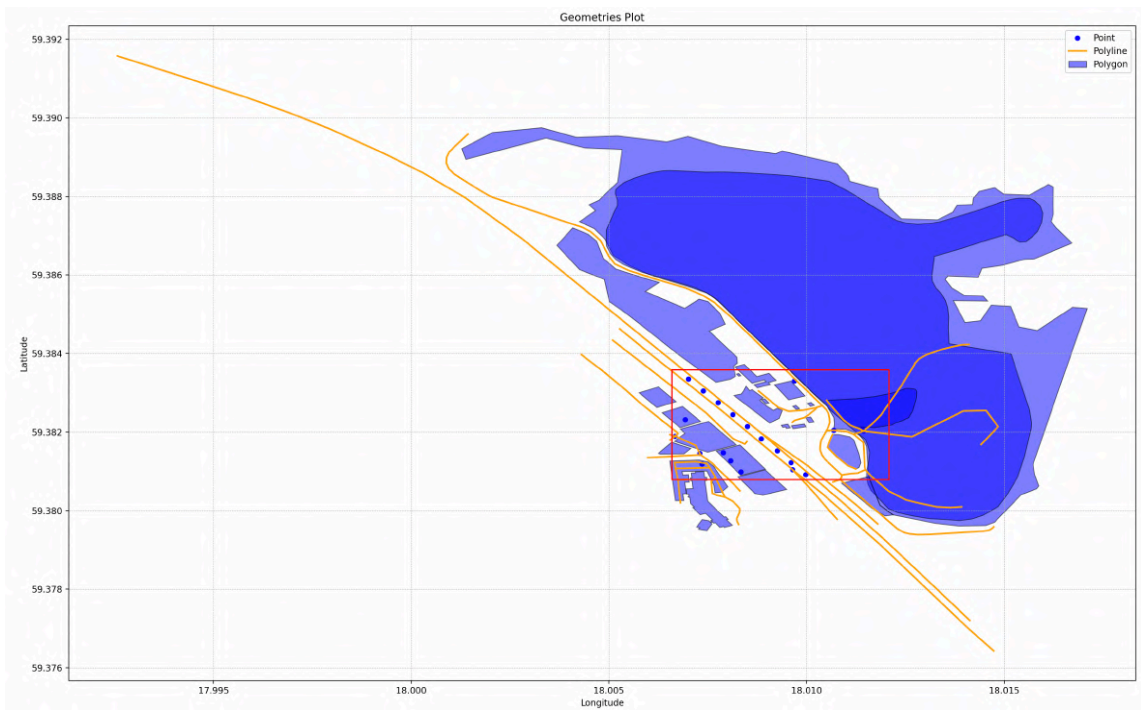

**Figure 13:** Preprocessing: A tile before clipping and processing, only information within the red perimeter is retained.

The approach does not generate any empty tiles, meaning that tiles with no overlapping OSM features are omitted. To avoid unnecessary overhead in read/writes, the 300m tiles are grouped together and stored in files of 16 tiles, lowering the number of files needed to store all information.

# 5.6   Simplification and Graph Creation

While the tiling divides the dataset into more manageable chunks and simplifies processing, the "unprocessed" tiles need to be processed for use in our models. To start, all geometries are simplified using the Douglas-Peucker algorithm [100], reducing the number of nodes without affecting the overall shape, saving size and complexity while preserving essential geometric information. All tiles are then projected into local geodetic coordinates [101] and normalised between 0.0 and 1.0, as seen in Figure 14. Geometries are converted to graph structures as detailed in Section 7.6.7. Additionally, each feature's minimal area bounding box is computed for use as positional encoding and masking as detailed in Section 7.6.4 and Section 7.6.8 respectively.

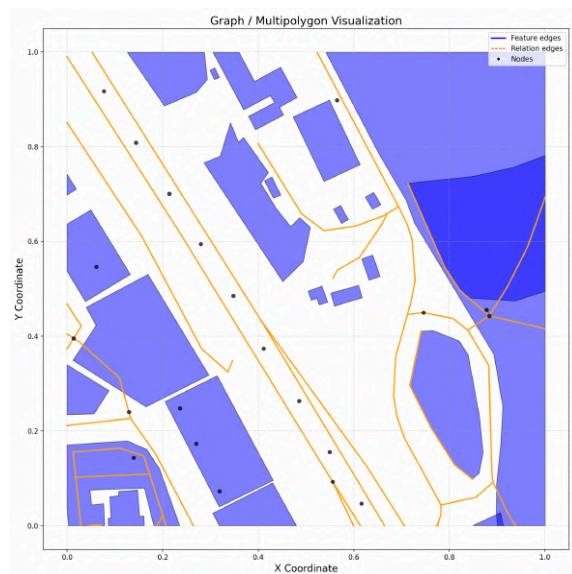

**Figure 14:** Completed tile processing.

# 5.7   Sourcing Aerial Imagery

Aerial images spanning large areas of the United States are downloaded from the *National Agriculture Imagery Program* [18] using the planetary-computer API [102]. We download images in rectangular bounding boxes of San Fransisco, Los Angeles, Sacramento, Charlotte, Washington, and Atlanta. Single images span roughly 6 km by 7.5 km with a resolution of either 0.6 or 1 meter per pixel and are tiled using the same coordinates and tiling system as the processed OSM tiles. In multiple cases, the border of the aerial images intersects our tiles resulting in incomplete image tiles. To address this, we merge image patches that end up at the same tiled coordinates by selecting the highest-intensity pixels. During tiling, images are resized to match the desired input of the image models, 224 by 224 pixels, resulting in a final scale of 1.33 meters per pixel.

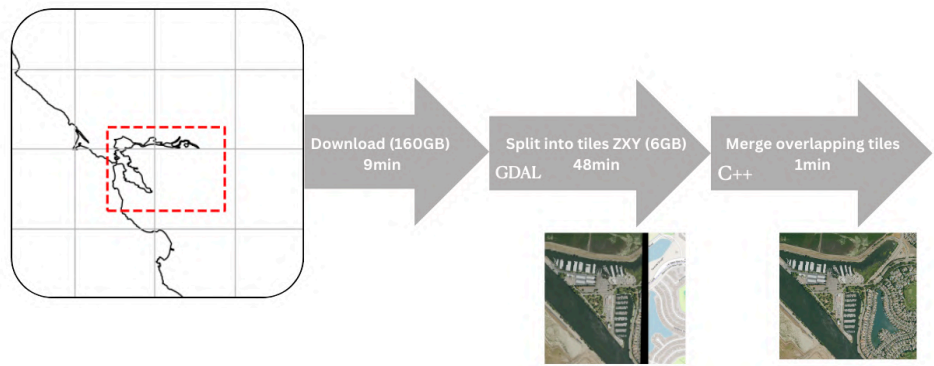

**Figure 15:** An overview of the pipeline for tiling aerial imagery of San Francisco.

## 5.8 Correlating Map- and Image Tiles

To make use of both the OSM data and the aerial imagery, it is imperative that the dataset only contains tiles where data from both is available. For this purpose, a simple Python script is used to cycle through the map tiles and consolidate all pbf files with corresponding images into a final dataset.

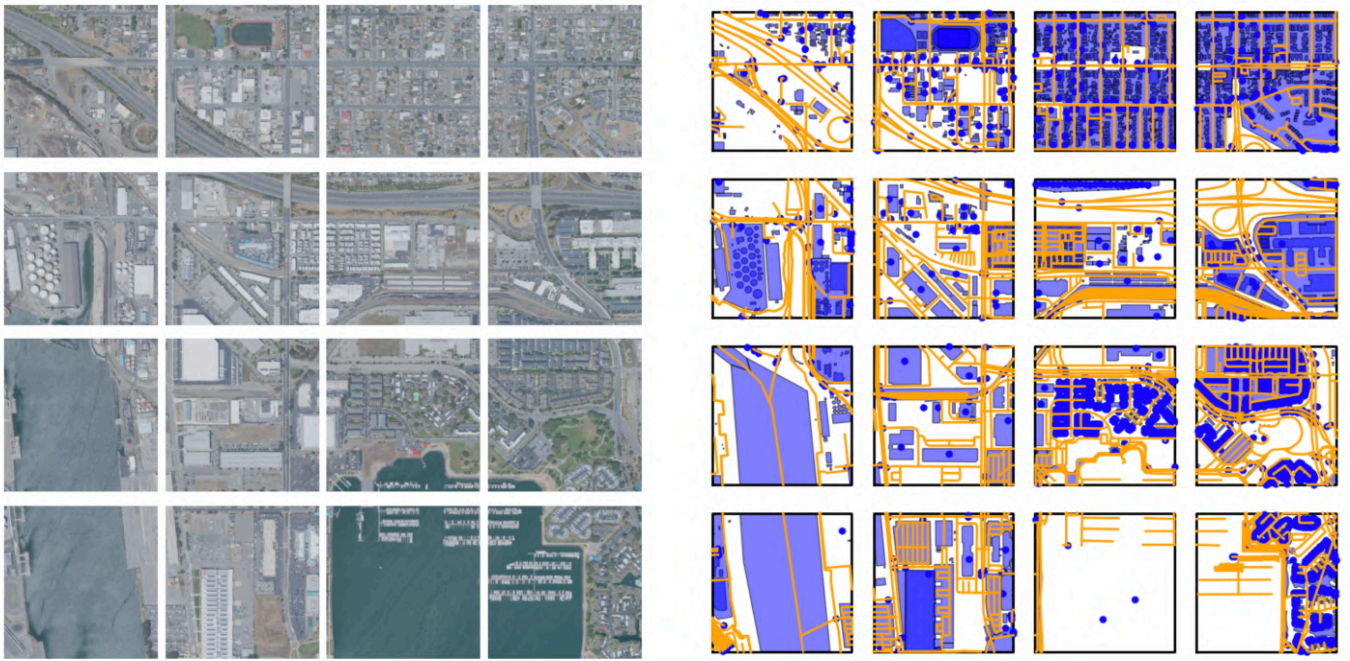

**Figure 16:** Tiled aerial imagery at 1.33m/pixel (left) and matching processed tiles from OSM (right).

# 5.9   Outlier Filtering

To reduce the memory footprint in intensive training workloads, tiles containing more than 1250 entities are filtered out. These high-content tiles amount only to 167 out of 290k+, and are as such deemed to have a negligible impact on the results.

Additionally, low-content tiles with fewer than five entities are filtered out. These tiles can be meaningful to train on but more often than high-content tiles lack sufficient coverage, rendering them suboptimal for training purposes. Approximately 30% of tiles belong to the low-content group. Removing them gives a higher-quality and more feature-rich dataset at the expense of potentially losing the ability to reason over a common tile type. We argue that excluding the low-content group represents a reasonable narrowing of the thesis scope, based on our belief that downstream tasks are generally more focused on higher-content areas, though this may not hold universally true.

# 5.10   Dataset Statistics

The dataset spans multiple US states, totalling 36.4GB after processing, including aerial imagery and OSM data, with the distribution shown in Table 6. Geographical coverage is shown in Figure 17, and Table 7 highlights the content of OSM tags.

**Table 6:** Dataset size for each stage. The final size where both image and OSM data are available is 36.4GB. The tiling and processing of OSM data takes 38 minutes on our 16-vcpu laptop.

| Dataset | Raw Protobuf | Tiled | Processed |
|---|---|---|---|
| California | 1.21GB | 5.45GB | 6.2GB |
| Georgia | 325MB | 1.1GB | 1.8GB |
| Maryland | 192MB | 871MB | 1.2GB |
| North Carolina | 387MB | 1.85GB | 2.2GB |
| South Carolina | 149MB | 757MB | 878MB |
| Virginia | 392MB | 1.73GB | 2.1GB |
| Total | 2.6GB | 9.9GB | 14.4GB |
| Filtered + Images | - | - | 6.4GB + 30GB img |

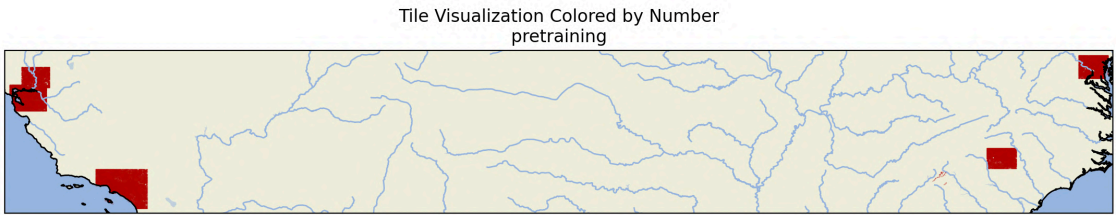

**Figure 17:** Map of the US, dataset coverage marked in red.

**Table 7:** The number of map entities in some common categories.

| Tag | Small | Huge |
| --- | --- | --- |
| Amenity | 3 500 | 403 300 |
| Bridge | 578 | 59 784 |
| Building | 10 701 | 11 283 910 |
| Crossing | 4 214 | 408 807 |
| Footway | 5 046 | 494 182 |
| Highway | 64 925 | 6 177 150 |
| Maxspeed | 4 484 | 478 289 |
| Railway | 2 007 | 124 957 |
| Shop | 799 | 19 491 |
| Traffic Signal | 264 | 27 604 |

# 5.11 High-Performance Data-Loading

Since the dataset does not fit in memory, and loading tiles is identified as a bottleneck in model training, we do as much processing as possible in the `C++` processing binary. Additionally, we store all processed OSM data and images in HDF5 file databases for high bandwidth reads [103]. This increases raw tile throughput by a factor of ten, minimising data-loading overhead and improving overall training efficiency and scalability.

# Chapter 6

# Evaluation and Baselines

*To assess the performance of geospatial embeddings, we design a comprehensive test suite spanning five geospatial and remote sensing regression tasks which are synthesised from the tiled dataset. As a baseline, a diverse set of models are trained and evaluated on these tasks.*

## 6.1  Task Selection and Generation

We create synthetic tasks from the tags of the OSM dataset. To efficiently generate specific datasets for every downstream task, we create a pipeline where a simple task configuration file is manually created, after which a command is used to generate the file structure where the files are split into predetermined *train*, *val* and *test* sets before being processed as the configuration file specifies. The *train*, *val*, *test* split is done on the file level, and as each file contains 4 by 4 tiles, spatial proximity between *train* and *test* tiles is somewhat reduced.

The configuration files specify how different tags should contribute to the tile label (see Appendix C for exact configurations). The first task we present is *Traffic Signals*, where the number of traffic signals is counted. This task presents some uncertainties since crossings are sometimes marked with a single traffic signal entity, and sometimes marked on each read leading up to the crossing.

The second task is *Buildings*, where each building is counted once. We hypothesise that image models are well suited for this task, seeing that buildings are visible in the images, unlike *Traffic Signals*. Additionally, `[building=yes]` is the most common tag in OSM, suggesting higher-quality data.

Furthermore, we introduce two closely related variations: *Bridge* and *Car Bridge*. Both are binary classification tasks, with the goal of determining whether a tile contains a bridge or an overpass. We further developed *Car Bridge* to ignore any pedestrian bridges, as these seem unfairly difficult to identify from aerial imagery.

The final task is *Max Speed*. This task is different from the rest in that we only care about a single value out of many available in the tile, namely the maximum speed limit of any road in the tile. However, most roads in OSM do not contain speed information. Our solution is to remove all tiles where roads are present, but no max speed tags can be found. Furthermore, we assign a negative value of $-100$ to tiles where no roads are present to differentiate between low speeds and no roads.

For all datasets, the labels are analysed and filtered to create a more suitable distribution. For example, in the *Buildings* task, 90% of all tiles containing no buildings are removed, leading to about 12% of tiles where the correct answer is 0, compared to around 60% unfiltered. The reasoning for doing this is to limit the success that can be achieved by simply predicting zero on every sample and to aid learning algorithms to escape that local minimum.

For simplicity, we model all tasks as regression problems, even binary tasks like *Bridge*. In this case, 0.0 is *no bridge* and 1.0 is *bridge*. Some tasks like *Buildings* and *Traffic Signals* benefit from this approach while some, like *Bridge*, do not. During testing, we clamp predictions to the range specified by the task, such as [0.0-1.0] for *Bridge*. During training, we optimise the Mean Square Error (MSE) also commonly referred to as the $L_2$ loss.

$$L_2 = \frac{1}{n} \sum_{i=1}^{n} (y_i - \hat{y}_i)^2 \tag{1}$$

**Figure 18:** $L_2$ loss, $n$ is the number of samples, $y$ is the true value, and $\hat{y}$ is the predicted value.

Since it is hard to gauge practical performance from the $L_2$ loss, all results are displayed with Mean Absolute Error (MAE) or $L_1$ loss.

$$L_1 = \frac{1}{n} \sum_{i=1}^{n} |y_i - \hat{y}_i| \tag{2}$$

**Figure 19:** $L_1$ loss, $n$ is the number of samples, $y$ is the true value, and $\hat{y}$ is the predicted value.

## 6.1.1 Masked Datasets

Some models are trained end-to-end on OSM data and have unfair access to the labels that are used to create the task. As such, their ability to generalise to cases where the correct tag is not used can be severely impaired.

With this in mind, we choose to create an additional dataset for each task where we remove tags that are strongly connected to the task we want to solve. As can be seen in Table 36, for the task *Bridge*, this naturally encompasses the `[bridge=yes]` tag, but also any `[layer=*]` tag since any entity that has layering is likely to either be a bridge or be located under a bridge. Another example of masking is shown in Figure 20.

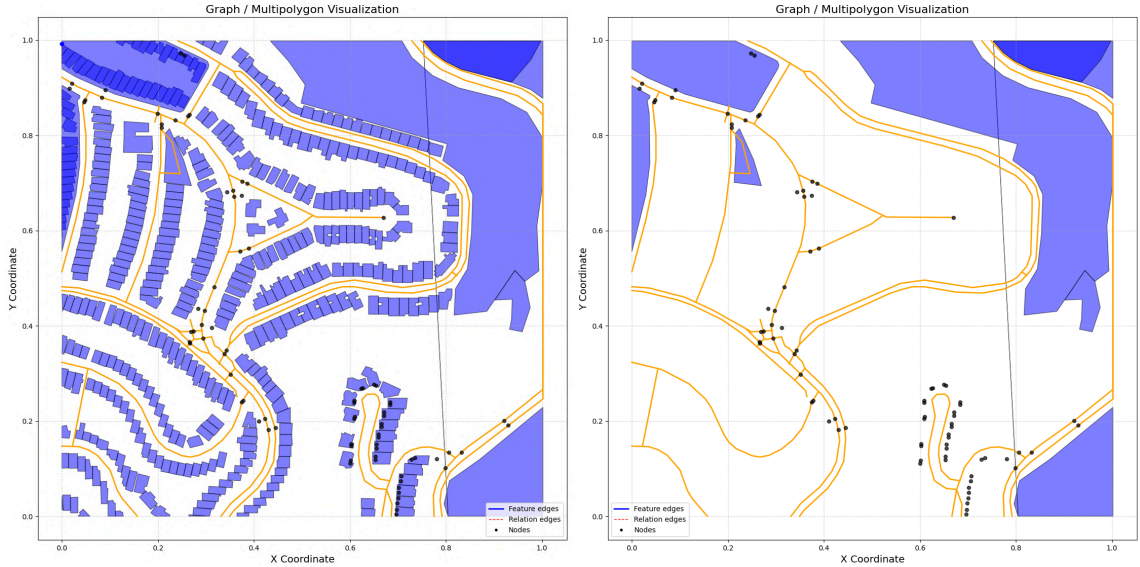

**Figure 20:** Unmasked tile (left) and masked tile (right) from the *Buildings* task.

The masking of different tasks leaves us with a dilemma; we want to limit the unfair advantage of end-to-end training on OSM labels, but while highly related tags should be removed, all co-occurring tags can not be removed since that would make predictions impossible. To help us make these decisions we calculate and analyse the co-occurrence of tags, both inter-entity and intra-entity. All masked tags and entities can be found in Appendix C. By implementing these masking strategies, we aim to provide harder tasks that are better at assessing a supervised model's ability to reason over geospatial data.

## 6.2 Model Summary

In short, the test suite aims to provide baseline references on numerous remote-sensing and geospatial intelligence tasks. We evaluate two LLM type models, four image models utilising transfer learning, three fine-tuned image models and five custom models trained on OSM tags.

**Table 8:** All supervised baseline models with parameter count and modality type.

| Model | Frozen | Trained | Modalities | Training |
|-------|--------|---------|------------|----------|
| *Large Language Models* | | | | |
| ChatGPT 4o | ? | - | Text/Images | Zero-Shot |
| ChatGPT 4o-mini | ? | - | Text/Images | Zero-Shot |
| *Frozen Backbone Image Models* | | | | |
| EfficientNet | 4M | 526k | Images | Frozen |
| ResNet50 | 20M | 329k | Images | Frozen |
| ViT-B/16 | 80M | 198k | Images | Frozen |
| ScaleMAE | 300M | 264k | Satellite Imagery | Frozen |
| *Fine-tuned images models* | | | | |
| AG-4M | - | 4M | Images | Fine-tuned |
| AG-97M | - | 97M | Images | Fine-tuned |
| AG-197M | - | 197M | Images | Fine-tuned |
| *Custom OSM tag models* | | | | |
| TagCountMLP | - | 3.5M | OSM Tags | Full |
| Tagformer | - | 2.6M | OSM Tags | Full |
| Tagformer-P | - | 2.7M | OSM Tags | Full |
| Tagformer-S | 12.8M | 1.6M | OSM Tags | Frozen |
| Tagformer-SP | 12.8M | 1.7M | OSM Tags | Frozen |

## 6.3 Multimodal LLMs

With the rapid performance improvement and multimodal capabilities of Large Language Models (LLMs), they earn a definite place in the evaluation suite.

The tests of LLMs are made using the ChatGPT API, where the ChatGPT 4o and ChatGPT 4o-mini models are requested. For each task, we first create a specific query by manually testing on a few images. We iterate on each query until the results are reasonable or we can't improve the results. The final queries can be seen in Appendix B.

In every request, 10 aerial images are sent to the API. Compared to sending them one by one, it significantly improves execution time and performance. This is not something we experiment heavily with, but the API imposes certain message length limits. This approach is neither *zero-shot* nor *multi-shot* evaluation since no labeled examples are passed to the model, but multiple unlabeled ones. We theorise that having access to multiple unlabeled samples lets the model compare the samples before labeling them. In favour of ChatGPT, we refer to the approach as *zero-shot* evaluation. Due to the costs associated with querying the ChatGPT API, a smaller dataset of 400 random but representative tiles from the *Huge* test dataset is created for each task.

# 6.4 Transfer Learning

Transfer learning refers to the act of applying a pretrained model on a new domain or task by leveraging its knowledge in a related domain [104]. In this section, we study two such implementations.

## 6.4.1 Frozen Backbone

A simple approach is to pass a pretrained model's latent features through a classification or regression head without training the parameters of the pretrained model. This is a memory- and compute-efficient approach as only a small fraction of the parameters are optimised. Conceptually, the pretrained model extracts key features of the image while the head learns to transform those into the correct answer.

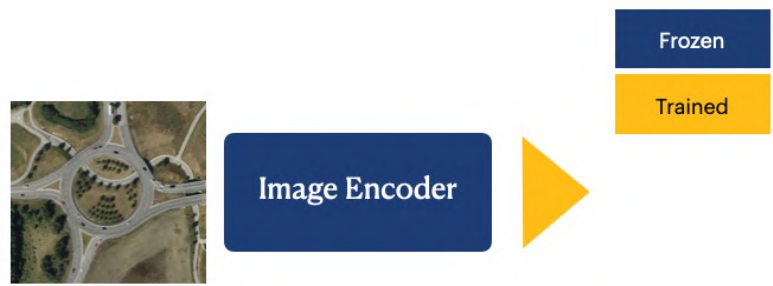

**Figure 21:** Transfer learning using frozen backbone.

There are many different kinds of heads that can be applied to a model like this, but for simplicity and robustness, a linear feed-forward network is typically used.

$$\hat{y} = Wx + b \tag{3}$$

**Figure 22:** *Linear feed-forward network:* The predicted value ($\hat{y}$) is given by multiplying the input vector ($x$) with the weight matrix ($W$) and adding the bias vector ($b$).

However, in some cases, non-linearity is needed to map the backbone latent space to the output space, such as when the task has a non-linear nature. In this case, a Multi-Layer Perceptron can be used instead, essentially a stacked fully-connected network with an activation function on the hidden layer.

$$\hat{y} = W_2 f(W_1 x + b_1) + b_2 \tag{4}$$

**Figure 23:** *Two-layer perceptron:* Two applications of a linear net intercepted by an activation function ($f$) to introduce non-linearity.

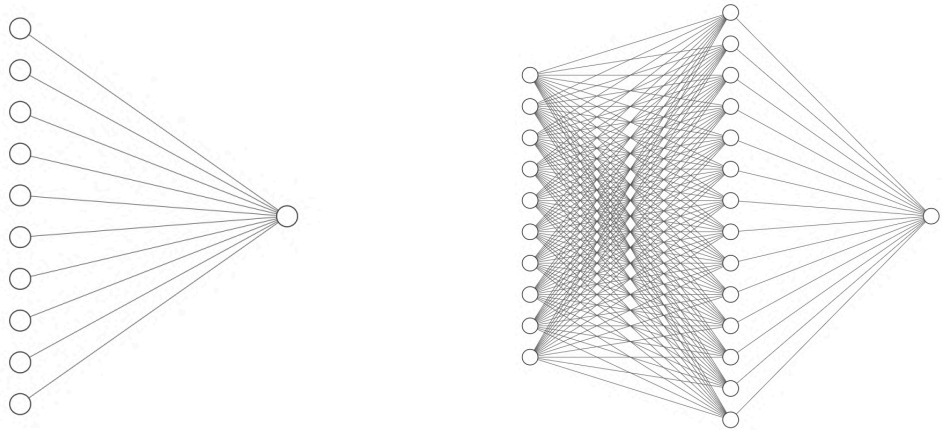

**Figure 24:** A linear network regressor (left) and a multi-layer perceptron regressor (right). These are also referred to as linear/multi-layer **heads**.

We evaluate four different image backbones, listed in Table 9. Notably, the ScaleMAE model is trained on varied-scale satellite imagery and excels at building segmentation. For all model-task combinations, we train both a linear regressor (LIN) and a multi-layer regressor (MLP) during 200 epochs using a *reduce-on-platue* learning rate scheduler.

**Table 9:** The four pre-trained image backbones used for our frozen backbone evaluation.

| Model | Frozen parameters | Trained parameters (LIN) | Trained parameters (MLP) | Training task | Architecture |
|---|---|---|---|---|---|
| EfficientNet-B0 [105] | 4 M | 1281 | 329 k | Image classification | CNN |
| ResNet-50 [37] | 20 M | 2049 | 526 k | Image classification | CNN |
| ViT-B/16 [106] | 80 M | 769 | 198 k | Image classification | Transformer |
| ScaleMAE [107] | 300 M | 1025 | 264 k | Building segmentation | Transformer |

The purpose of including this type of model (*Frozen Backbone*), is to assess some upper limit of generic embedding evaluation from image-only data. An important aspect to note is that since the image backbone is not trained, the image embedding could be saved to a database and used for various downstream tasks like classification or semantic tile lookup. In fact, we precompute the embedding vectors for each backbone once and save them to disk. The embeddings for ViT-B/16 take up 0.5% of the total image library size, so instead of running the ViT-B/16 model 200 times per image and task and head, the embeddings can be kept in working memory for fast training.

## 6.4.2 Autogluon Fine-Tuning

On a similar note, it is common to unfreeze the backbone and train it at a lower learning rate than the head, as illustrated in Figure 25. This allows the backbone to encode important information for the specific task and can improve performance. However, we lose any guarantee that the backbone encodes features that benefit generalisability to other downstream tasks. For those reasons, it is not a good fit for a general-purpose application. Instead, these models are added in an attempt to assess the maximum task-specific performance using image-only data. For that reason, we do not use the same frozen backbones as in Section 6.4.1, but instead utilize the Auto-ML framework AutoGluon [108, 109]. This also removes the reliance on our specific training regiment. AutoGluon will automatically figure out how to normalise the data and select a backbone to fine-tune. The problem is still modelled as a regression problem.

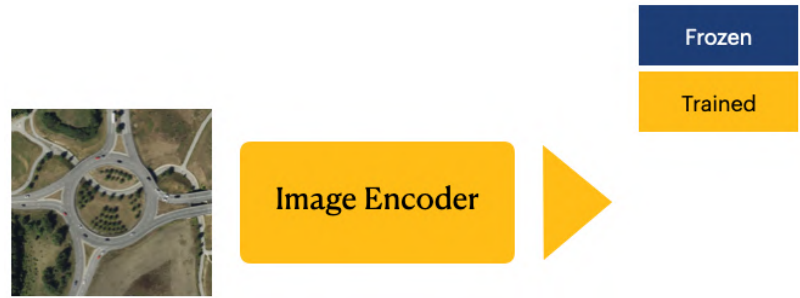

**Figure 25:** Transfer learning using fine-tuning.

We train AutoGluon image models with the *medium_quality* and *high_quality* presets for 600 and 900 seconds. We also train one model with the *highest_quality* preset for 3600s. To simplify comparison they are referred to by their parameter count, listed in Table 10.

**Table 10:** Autogluon fine-tuned image models.

| Model | Quality | Parameters | Training time |
|---|---|---|---|
| AG-4M | *medium* | 4 M | 600s/900s |
| AG-97M | *high* | 97 M | 600s/900s |
| AG-197M | *highest* | 197 M | 3600s |

Additionally, AutoGluon can train on a combination of text, image, and tabular data with a multimodal fusion model, illustrated in Figure 26.

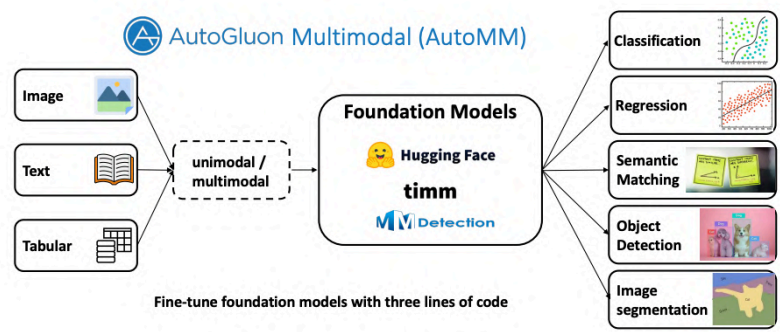

**Figure 26:** Autogluon multimodal overview from [109].

We design a tabular dataset containing the count of the 12500 most common OSM tags but immediately exceed the memory limits on a 64GB RAM machine. We prune all tags with less than ten occurrences in our dataset and end up with approximately 2500 columns. Training takes 4-5 hours per epoch and does not converge. Our tabular + image dataset requires even more resources and does not converge either.

## 6.5 TagCountMLP Model

As an alternative to Autogluon Tabular we design an efficient custom model that works in a similar fashion. We download a list of the same 12500 most common OSM entities [30] and multi-hot encode tag counts on a per tile basis.

**Table 11:** Example of tag counts per tile.

| tile_name | building=yes | highway=primary | highway=secondary | ... |
|---|---|---|---|---|
| 16_18052_25957 | 24 | 0 | 3 | |
| 16_17147_25253 | 250 | 34 | 56 | |
| ... | | | | |

The encoded vector is then normalised and projected down from 12500 dimensions to 256, after which a multi-layer perceptron is engaged. The model directly learns which columns are important for doing task-specific predictions, with no prior knowledge about column semantics or significance. On unmasked tasks, the label is directly available in the input data. We test the model on both masked and unmasked datasets.

## 6.6 Tagformer Model

To further improve on tag-based performance, we implement a transformer-based model referred to as the *Tagformer*. This model has entity-to-entity attention, semantic priors and positional encoding.

We multi-hot encode tag existence on a per-entity basis, modelling each OSM entity as one token. Each token is projected down to 128 dimensions and passed to stacked transformer encoder layers. For every tile, a class token, `[CLS]`, is added. After attending to the entire tile, the `[CLS]` token is passed to a linear regression head.

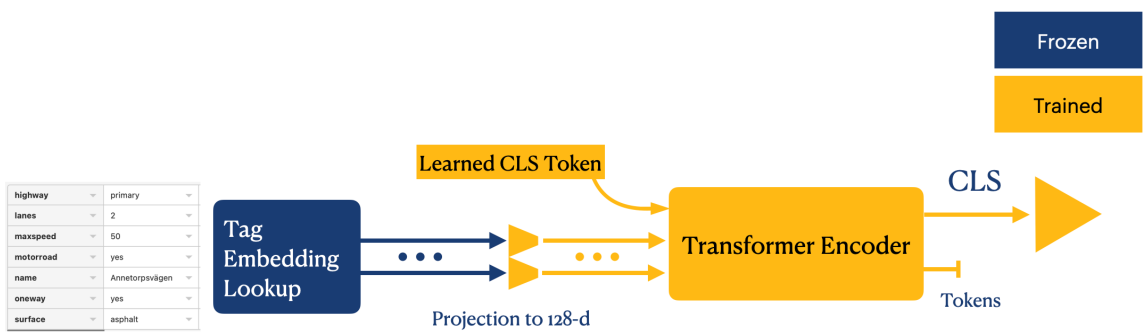

**Figure 27:** Tagformer model architecture.

## 6.6.1   One-hot Encoding of Tag Occurance

Following the straight-forward approach of tag counting used in the previous model, the *Tag Embedding Lookup* of the base *Tagformer* produces the 12500-dimensional tag count vector but for every entity. These are then projected down to 128 dimensions and used as tokens.

## 6.6.2   Semantic Priors

A limitation of the one-hot encoding method of tags is that the model can not generalise to out of training tags. In OSM, different countries and regions use slightly different tagging guidelines [110], and some are just misused by editors or ambiguous, such as `[bicycle=yes]` vs `[bicycle=designated]` vs `[bicycle_road=yes]` [111]. As such, it could be beneficial to include some semantic prior or description of how these tags are similar or different.

In classification tasks, a model is usually trained to output the probabilities on a fixed set of labels. In instrument classification from multi-track sound recordings, the model proposed by Riou et al. [84] generalises to instruments not seen during training by predicting the word embedding of the instrument. In the same spirit, we believe that using semantic embeddings for tags may allow the model to generalise to unseen OSM tags and even allow for free text input. Unseen tags and free text input is not evaluated for the *Tagformer*-architecture, which is task-specific, but the ideas are important in GeoJEPA. We therefore evaluate feasability and potential information loss of semantic priors by implementing such a model, referred to as *Tagformer-S*.

To create the semantic embeddings of OSM tags, we enrich each key-value pair with the corresponding description from the OSM wiki [112] and embedded them with voyage-3-large [113] into a 1024-d embedding space.

**Table 12:** Some ambiguous OSM tags with their corresponding WIKI description, in total there are 116 distinct tags containing `bicycle` and 85 distinct tags containing `cycleway`.

| OSM Tag | WIKI Description |
|---|---|
| `[bicycle=designated]` | Roads and other objects designated/signed to use for cyclists |
| `[bicycle=yes]` | Roads and other objects accessible to cyclists, but not explicitly designated/signed for their use |
| `[railway=crossing_box]` | A railway building where the crossing attendant stays. |

During training or inference, we compute the unweighted average of each entity's tag embedding vectors. A process which may result in information loss and inadequate differentiation between important and less important tags. Other alternatives, such as weighted average, attention mechanisms, or TF-IDF variants are ignored for simplicity.

### 6.6.3 Positional Encoding

We hypothesise that the relative position, area and rotation of OSM entities can be useful for certain classification and regression tasks and add a positional encoding based on the entities' minimum area bounding boxes. We implement such a model and refer to it as *Tagformer-P*. The approach is further detailed in Section 7.6.4 as the technique is also used in GeoJEPA. Additionally, we implement a model with both positional encoding and semantic priors, referred to as *Tagformer-SP*.

### 6.6.4 Training Efficiency

Irrespective of the small model size compared to some of the larger image models, the memory usage is high due to the context length variability in different tiles. While some tiles contain one or more OSM features, there are some outliers in the 1000-7000 feature range. The architecture utilises *flash-attention* [114] which significantly improves the performance of attention mechanisms. Even so, full attention is still fundamentally a quadratic operation. By removing the 167 highest-content outliers we move the longest sequence from 7100 to 1250, meaning large improvements to worst-case memory usage. Still, there are efficiency limits during batched training and inference due to the high padding ratio. There can be a 250x difference in sequence length within the same batch, wasting resources on discarded computations. Therefore, we implement a sequence length-aware data loader. We load multiple batches into memory, sort all tiles by length and once again bin them into batches, resulting in lower GPU utilization but a 2x speed up during training. We accumulate the gradients over the same number of batches to prevent training instability, as depicted in Figure 28.

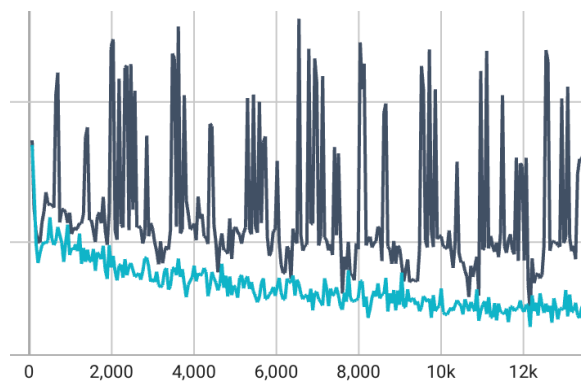

**Figure 28:** The loss scales with the number of tokens, so when it is calculated over short sequence batches separately from long sequence batches instability follows, such as when gradients are not accumulated over the same number of batches as we sort (black line).

### 6.6.5  Training Configuration

The *Tagformer* is trained from scratch on each separate task during 25 epochs in 3-12 minutes depending on the task. For each configuration, we employ a hyperparameter search over the base learning rate, final learning rate, and weight decay using 4 random samples and 6 samples using Bayesian optimisation.

   As the non-semantic configuration trains 60% more parameters, it is approximately 20% slower to train and utilises 8-9GB of VRAM compared to 3GB for the semantic configuration.

# Chapter 7
# General Purpose Models

*The baseline models in Chapter 6 either lack the ability to generalise to unseen tasks or are limited to aerial image data only. In this chapter, we present our general-purpose architectures for geospatial raster, tag and geometry data.*

*Additionally, previously unmentioned model-specific literature is intertwined with our implementation details to explain our choices.*

## 7.1 Requirements

For our general-purpose models, we base the architectural designs on the following requirements, in order of importance.

- *OSM map entities*: The model has to be able to encode geospatial information in the form of OpenStreetMap entities, with respect to their tags and/or geometric shape.
- *Region representations*: Based on our method of evaluation (tile-wise), a model has to create representations of entire tiles (regions). However, it would be of great practical benefit if entity-wise representations are created as well.
- *Multi-purpose embeddings*: The model should produce embeddings that are useful in a variety of analytical or practical downstream tasks.
- *Contextual information*: A single OSM entity is not very interesting on its own. It may have tags such as [maxspeed=50_mph] and [surface=asphalt], and a geometry of 3 points covering 4 meters. Therefore, it is of great interest to base the semantic meaning of an entity partly on its surrounding features.
- *Can reason over aerial imagery*: Aerial imagery provides a source of ground truths and therefore constitutes an important addition to lower the dependency on noisy OSM data.

- *Works with missing modalities*: Multimodal models that accept any combination of supported modalities naturally address a larger space of downstream problems, such as scenarios where only images are available and no map data.

## 7.2 Model Summary

In short, we introduce three self-supervised OSM tag models as baselines and present details regarding our novel GeoJEPA model. After training, the encoder models are used to produce embeddings which are evaluated using multi-layer and linear regressors as described in Section 6.4.1.

**Table 13:** Self-supervised models with parameter count and supported modalities. Modalities are **T** - OSM tags, **G** - OSM geometries, **I** - aerial images. In most models, only a fraction of the parameters are active during inference but the parameter count increases due to large frozen tokenisation modules and lookup tables.

| Model | Trainable | Inference | Modalities |
|---|---|---|---|
| *Baselines* | | | |
| TagPool | - | 12.8 M | T |
| TagAE | 17.6 M | 9.7 M | T |
| TagformerLMAE | 17.6 M | 22.5 M | T |
| *GeoJEPA Configurations* | | | |
| GeoJEPA-T | 26.5 M | 34.1 M | T |
| GeoJEPA-GT | 27.2 M | 43.5 M | G + T |
| GeoJEPA-TI | 26.8 M | 115.1 M | T + I |
| GeoJEPA-GTI | 27.2 M | 123.7 M | G + T + I |

## 7.3 TagPool Model

The *TagPool* model offers a straightforward method for generating representations of regions and map entities without the need for training. Similar to the *Tagformer-S*, it relies on the pre-computed word embeddings of tag and description pairs, which are averaged to produce a map entity representation. Notably, these representations lack contextual information. Regional representations are calculated by performing max pooling and average pooling operations which are subsequently concatenated, as illustrated in Figure 29. This process results in map entity representations of 1024 dimensions, and regional embeddings of 2048 dimensions.

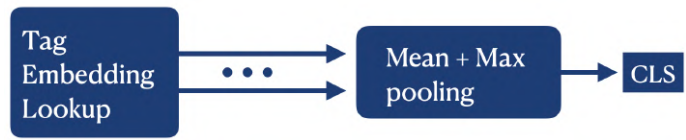

**Figure 29:** The TagPool model aggregates LLM embeddings of OSM entity tags.

## 7.4 TagAutoencoder Model

In an attempt to achieve generalisation across unseen tasks, we adopt a generative pre-training objective (Section 4.5.1) to our existing *TagCountMLP* model from Section 6.5. As the reconstruction loss is non-trivial to define in the multimodal scenario, this is an easy, yet incomplete way to build a general-purpose embedding model for OSM data. In this naive architecture, we use the *TagCountMLP* to encode a 256-d embedding vector from the normalised multi-hot encoded tile-wise tag count. An MLP-based decoder then reconstructs the 12500-d tag count vector. The same idea has been explored by Leśniara and Szymański [40], in which 88 highway-specific tags were used to perform micro-region clustering.

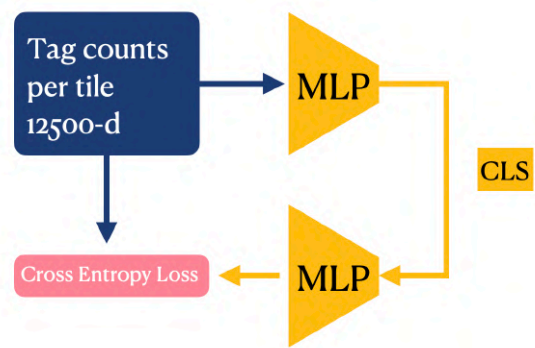

**Figure 30:** The TagAE model employs autoencoding to learn distilled representations of regional tag counts. Again, blue modules are frozen and yellow modules are trained.

## 7.5 TagformerLMAE Model

Seeing that the *Tagformer* has higher capabilities than the *TagCountMLP*, we implement a self-supervised model based on the *Tagformer-SP* by utilising latent masked autoencoding. Semantic LLM embeddings are used for initial map entity representations. By using a high masking ratio and CLS pass-through, we hope to increase the representational power of the CLS token. Token-wise information is kept by reconstructing the entity tag embeddings. The result is 256-dimensional representations of map entities and regions.

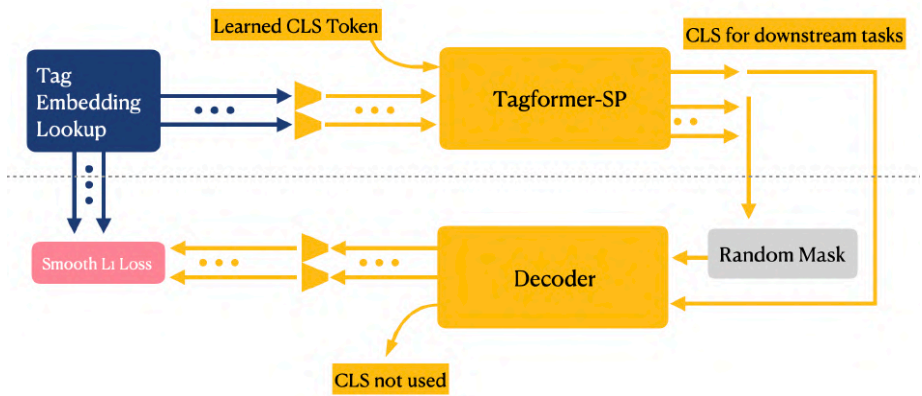

**Figure 31:** The TagformerLMAE model employs masked autoencoding in latent space. Tagformer output tokens are replaced with mask tokens containing min-box positional encoding.

## 7.6 GeoJEPA Model

To address the need for a general-purpose embedding model in geospatial data, we present GeoJEPA, one of the first multimodal adaptations of JEPA (after JEP-KD [115] from August 2024 and D-JEPA-T2I [76] from November 2024). Additionally, GeoJEPA is the first model for geospatial data based on JEPA.

GeoJEPA leverages pretrained unimodal models in the tokenisation step to produce rich semantic representations of aerial images, tags, and geometries. These representations are passed as tokens to a single-stream transformer encoder trained with JEPA. As such, GeoJEPA can be easily adapted to any relevant input data without changing the training objective or loss function.

### 7.6.1 GeoJEPA Representations

GeoJEPA generates semantic representations of regions, in this study 300x300 meter tiles, as well as token-level representations corresponding to individual map entities and image patches. During the transformer encoder layers, each token updates its representation by attending to the surrounding context. While this process suggests that the output token embedding captures the semantics of a specific map entity and its context, such an assumption may not hold. For instance, Darcet et al. [49] demonstrate that redundant tokens in transformers can function as registers, storing global rather than local information. While raising concerns, the effectiveness of transformer models for both sequence-level and token-level tasks has been demonstrated in other domains, such as in masked language modelling by Devlin et al [36]. Notably, the application of token-level embeddings from JEPA-pretrained models remains unexplored, making it a high-reward topic to evaluate.

## 7.6.2 Architectural Overview

GeoJEPA tokenises three distinct modalities: OSM tags, OSM geometries, and aerial images, which are all linearly projected to a joint vector space. For every map entity, the corresponding tag and geometry tokens are fused using a multi-layer perceptron, reducing the sequence length and enabling the formation of joint map entity representations.

As illustrated in Figure 32, a single concatenated token sequence is formed, encompassing tag, geometry, and image information. To make GeoJEPA able to differentiate between different tokens, we employ a positional encoding in the form of tile-local normalised coordinates and token modality.

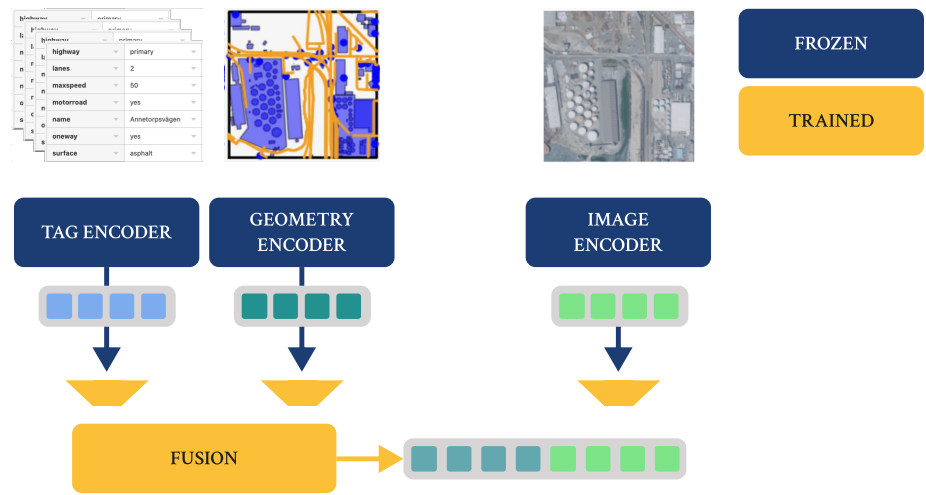

**Figure 32:** GeoJEPA tokeniser architectural overview. Frozen encoders produce semantic representations of tags, geometries, and images. Tags and geometries are fused to create map entity tokens. Map entity tokens and image patch tokens are then concatenated to create the region token sequence. Simultaneously, positions and modalities are extracted for later use.

The region token sequence is then separated into *context* and *targets*, as detailed in Section 7.6.8, which are passed to the GeoJEPA training architecture as illustrated in Figure 33.

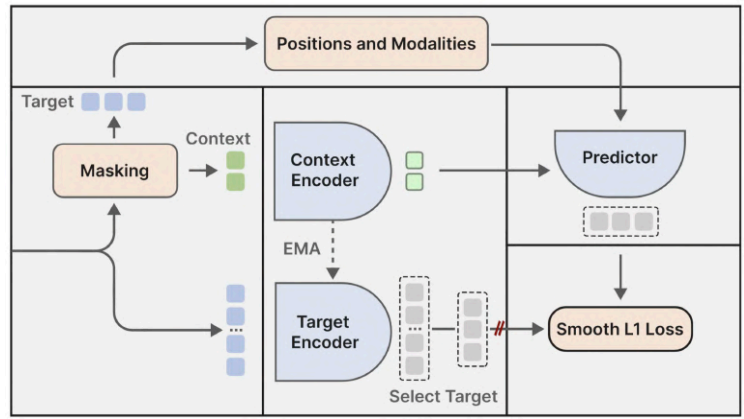

**Figure 33:** GeoJEPA training objective, diagram adapted from [75, p. 5]. The target encoder consumes the entire region token sequence and produces an informed latent representation of the target sequence, which the predictor is tasked to predict from the encoded context and the positions of the targets.

### 7.6.3 Modularity

The tokenisation step enables encoding any relevant information as latent tokens for the encoder, providing a modular foundation for integrating various modalities. Theoretically, training the model with multiple modalities can enable it to learn from the complementary information and enhance its ability even during unimodal tasks. A significant advantage of the single-stream architecture in this aspect lies in its inherent flexibility in handling scenarios where one or more modalities are missing, something that is more challenging to achieve with dual- or multi-stream architectures.

To support missing modalities during inference, it is highly important to precondition the model for such scenarios. Consequently, we employ modality-wise dropout in the fusion of tag and geometry tokens with a 30% chance of dropping either one. Additionally, we present a novel multimodal masking strategy (see Section 7.6.8).

### 7.6.4 Positional Encoding

In previous work, both traditional sinusoidal [64, 79] and learned [88] positional encoding have been utilised. While Vaswani et al. argue that a sinusoidal encoding may allow the original transformer model to extrapolate to longer sequences than the ones encountered in training [48], we choose a learned positional encoding for versatility. We theorise that the positional encoding should include information such as position, size and rotation. As a solution, we base the encoding on the feature's bounding box. Since the axis-aligned bounding box would be misleading for diagonal features, we compute an approximative minimal bounding box in the `C++` pre-processing step by rotating the convex envelope of the geometry 10 degrees over 18 iterations and saving the minimal area box. A minimum side length of 1.5 meters is applied and bounding boxes for points are randomly rotated.

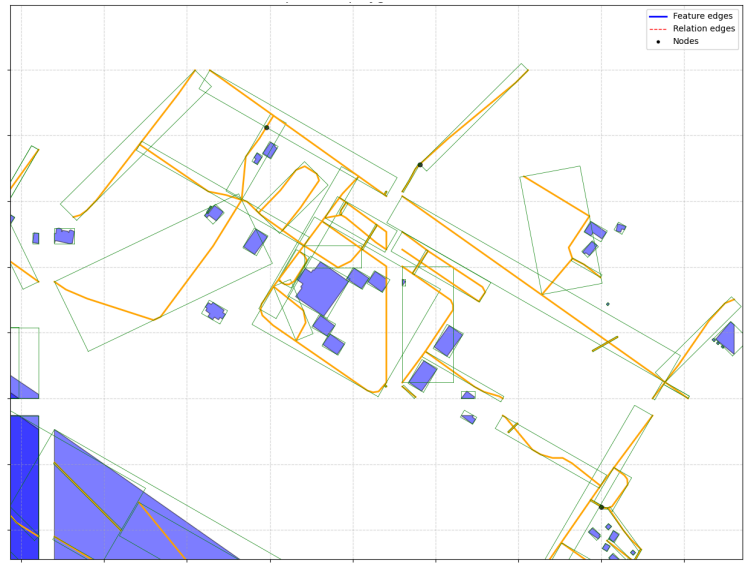

**Figure 34:** Minimum area bounding-box visualisation.

During training, the 4 coordinates of the bounding box are passed as an 8-dimensional vector through an MLP to create the positional embedding.

For `[IMAGE]` tokens, we compute axis-aligned bounding boxes corresponding to the area covered by the image patch. We theorise that the positional embedding allows GeoJEPA to cross-reference semantic information from tokens in different modalities that cover the same land area.

Modality-wise, we simply one-hot encode the token modality and embed it with a multi-layer perceptron. The modality encoding and the bounding-box encoding are fused using an MLP to maximise information content.

### 7.6.5   Tag Encoder

Going with the learnings from the Tagformer, we keep the same semantic tag embedding lookup as in Section 6.6. The tag encoder outputs one `[TAG]` token for every map entity in the encoded tile, ranging from 5 to 1250.

### 7.6.6   Image Encoder

In the quantitative evaluation (Section 8.4.1), we can observe that the domain-specific model, ScaleMAE [107], performs on par with the ViT-B/16 model [106]. Therefore we choose the pre-trained ViT-B/16 checkpoint from Pytorch as our image encoder due to its smaller size. The ViT-B/16 model outputs 196 image patch tokens and one class token, all with 768 dimensions. An image patch corresponds to a tile in a 14x14 grid of the image while the class token is an informational aggregation corresponding to the semantics of the image as a whole. While we, in ViT-B/16 transfer learning, only use the class token, we now keep all tokens. The reasoning behind this is to enable more detailed cross-modal learning. The image encoder, based on ViT-B/16 pretrained weights, outputs 197 `[IMG]` tokens.

### 7.6.7 Geometry Encoder

For the geometry encoder, a model that can encode polylines, polygons and multi-polygons is desired. To the best of our knowledge, only two such models exist. Veer et al. [116] show that CNNs and RNNs can learn directly from polygonal data, but don't verify their approach on multi-polygons or multi-lines. Siampou et al. [117] show that 2D Fourier transform and multi-layer perceptrons can be used to accurately encode points, polylines, and polygons, but don't attempt multi-polygonal input. The first polygon and multi-polygon encoder is the NUFTspec model by Mai et al. [31], which shows promising results on an adaptation of the popular MNIST dataset, where the handwritten digits have been converted from images to multi-polygons.

Yu et al. improve on this result using Graph Neural Networks (GNNs) [118]. Their model, PolygonGNN shows promising results on different polygon and multi-polygon tasks. In line with the paper, we train PolygonGNN on MNIST-P-2 classification (recognition of numbers between 10 and 99 represented as multi-polygons) and evaluate the embeddings of different OSM features with UMAP.

Since the model is not compatible with CUDA graph compilation using `torch.compile`, and ends up consuming a significant portion of the tokenisation step's resources, we first reduce the model size from 33M to 8.4M parameters by changing the hidden dimension from the default of 512 to 256 dimensions. We achieve an accuracy of 0.883, which is in line with the results presented by Yu et al as depicted in Figure 35.

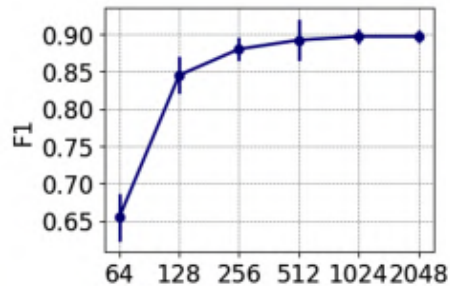

**Figure 35:** Expected PolygonGNN performance with varying hidden dimension size, from Yu et al. [118].

Additionally, we evaluate the embedding space using UMAP dimensionality reduction and conclude that the PolygonGNN model produces similarity-preserving embeddings of OSM geometries, illustrated in Figure 36 and Figure 37.

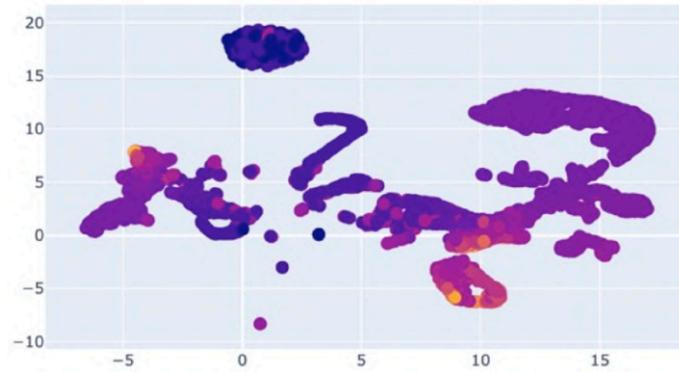

**Figure 36:** UMAP visualisation of PolygonGNN embedding space, the colour interpolates from purple to orange as the number of points in the geometry increases. In total, 10,000 OSM geometries are displayed.

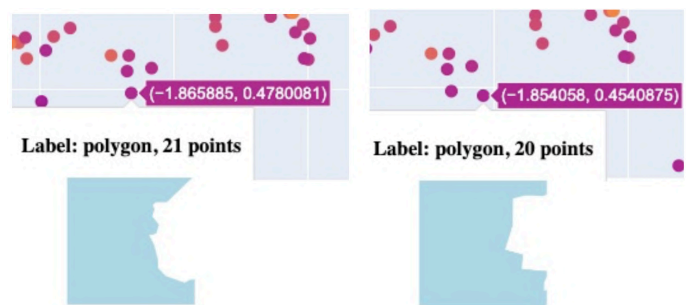

**Figure 37:** Close-up of two neighbouring geometries in embedding space. We see similar results for polylines.

Graph Neural Networks understand nodes and edges, not geometries. To address this, Yu et al. model multi-polygons by adding visibility edges, which are computed in a Python pre-processing step. We evaluate the provided code and conclude that it would be computationally tractable, but not practical, to compute visibility edges for all multi-polygons in OSM. Instead, we re-implement and optimize the algorithm in C++ and show a performance improvement exceeding 1500x. Additionally, we parallelise the computation on the file level (each file being ≤ 16 tiles). Pre-processing takes 14 seconds per task on the *Huge* dataset on our 16-vcpu machine.

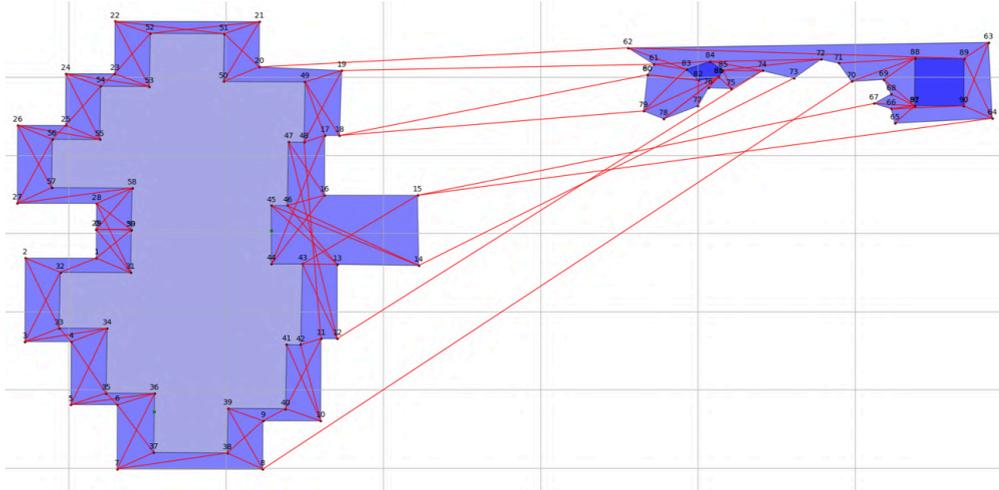

**Figure 38:** Visibility edges calculated using the `C++` implementation. The left geometry is an OSM building consisting of four separate tag-less polyline features, which are all part of a relation with the `[building=yes]` tag. The dark blue part is the building while the lighter part is the courtyard (left). A park with holes added to the relation for experimentation (right).

The trained PolygonGNN weights are loaded as part of GeoJEPA model instantiation. The geometry encoder outputs one `[GEOMETRY]` token per map entity.

## 7.6.8 Target Selection

Multiple authors find that context and target selection are of utmost importance in JEPA pretraining (e.g. [64, 72, 75, 88]). Therefore, in this section, we present novel masking strategies for unstructured geospatial data.

In traditional masked image modelling, a random subset of patches are removed. I-JEPA [64] introduces multi-block masking of images, for example predicting the upper right part from the upper left part. V-JEPA [79] performs spatial+temporal masking of videos. In both cases, JEPA is found to benefit from masking a high fraction of input data (75% − 90%). Intuitively, filling in a single missing pixel requires less high-level understanding than predicting the entire right side of the image, leading to a less sophisticated model. For videos, temporally consistent masking helps for the same reason. In both cases, simple interpolation of pixels would be viable for predicting the low-level details. In our case, three factors further complicate potential masking strategies.

1. *Varied sequence length*

Previously explored problems in JEPA literature have mostly considered structured, fixed-size data, such as images, videos, audio, or tabular. In our case, each sample (one 300x300m tile) has a fixed number of image tokens, but a varied number of OSM features. As such, a batch usually contains a high fraction of padding values, effectively preventing batch-wide selection strategies from being useful. Instead, each sample has to be masked individually and padded appropriately.

2. *Multimodality*

The multimodal nature of our problem give rise to a series of problems, but at its core, the model should be able to reason over any single modality or the combination of them, potentially requiring novel masking strategies to pre-condition the model for such scenarios.

3. *Non-trivial separation*

Selecting a contiguous block of image patches and OSM feature tokens is not trivial. Image patches have fixed size and rotation and discrete positioning while OSM features have freely varying size, rotation and positioning. Practically, an OSM feature can overlap any fraction of available image patches.

To address these challenges, we design and implement three viable masking strategies. All algorithms are fully vectorised over batches and present no substantial performance impact. By "vectorised", we mean that all samples in a batch are processed in parallel on the GPU, rather than processed sequentially. The algorithms are also padding-aware, and after application, tokens are reorganised to minimise necessary padding.

We begin by implementing `random_mask`, a function that randomly selects $X\%$ of the non-padded tokens in each sample. This process can be repeated multiple times, producing multiple *targets*, and leaving the remaining tokens as *context*.

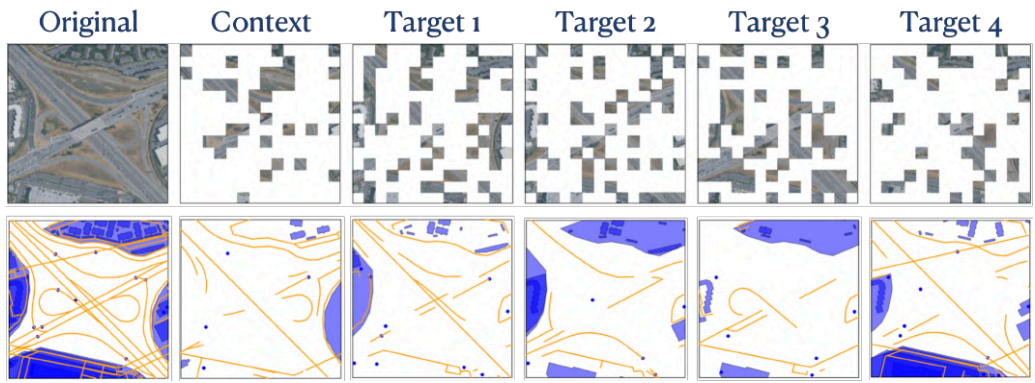

**Figure 39:** `random_mask` algorithm visualization for `[IMAGE]` tokens (upper) and `[OSM]` tokens (bottom).

It may seem like a sufficiently challenging prediction task, but previous research empirically shows that multi-block sampling yields superior results due to increased difficulty [64, 79, 88]. To imitate the concept of multi-block masking in unstructured data, we design a position-based masking algorithm that samples contiguous blocks of entities and image patches called `area_mask`. The algorithm samples a varied aspect-ratio bounding box and selects all tokens where the min-box centre coordinate is inside the bounding box.

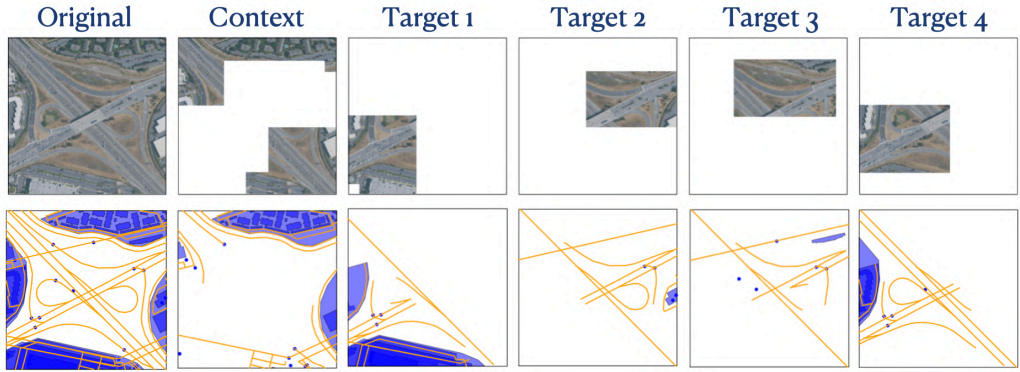

**Figure 40:** `area_mask` algorithm visualization.

Finally, we introduce the `modality_mask`. By forcing the model to learn cross-modal dependencies and to handle unimodal input we theorise that it achieves better cross-modal performance. The masking algorithm selects a single modality as context and creates targets from the remaining modalities, in this case only one.

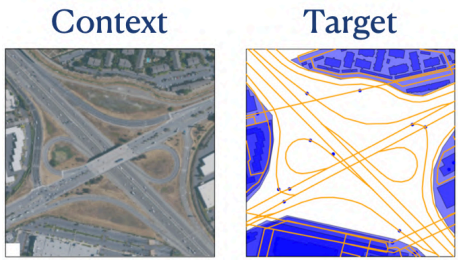

**Figure 41:** `modality_mask` visualisation IMAGE→OSM

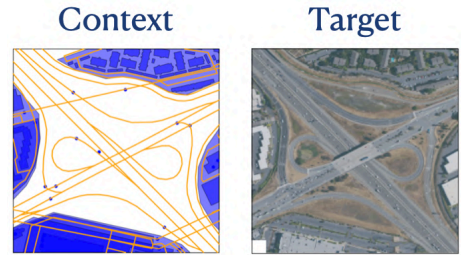

**Figure 42:** `modality_mask` visualisation OSM→IMAGE

We employ a `minimum_context` check that adds tokens randomly to reach a minimum context size.

During GeoJEPA pretraining we employ a weighted mask selection strategy. As suggested in the literature [72, section 5], we believe that increased training diversity will improve performance.

**Table 14:** Mixed strategy configuration.

| Strategy | $P(S)$ | ratio/target | min_ctx | num_targets | resulting ctx size |
|---|---|---|---|---|---|
| `random_mask` | 20% | 45% | 10% | 4 | 11% (avg) |
| `area_mask` | 60% | 40% | 15% | 4 | |
| `modality_mask` | 20% | - | 15% | - | 2.5%-86% |

## 7.6.9  GeoJEPA Loss Function

As a batch in GeoJEPA training may have a high number of padded values due to sequence length variance, as discussed in Section 7.6.8, the loss function must be modified in order to avoid biases and training instability.

For the latent reconstruction loss, the literature suggests (see Section 4.6.2) that *Huber loss* (Smooth $L_1$) is superior to $L_1$ or $L_2$ loss. For performance and stability reasons, a vectorised, padding-aware adaptation is implemented.

To address regularisation, we implemented an iterative, padding-aware InfoNCE loss [75]. However, the iterative approach, or our implementation in general, proved computationally heavy. As we fail to vectorise the loss function without compromising its correctness, we adopt the variance-covariance loss as outlined in VICReg [61]. It is easily vectorised and made padding-aware by computing variance and covariance over the concatenation of all valid tokens in the batch.

Both the reconstruction loss and the regularisation loss are scaled by the number of valid tokens to avoid biases.

## 7.6.10  GeoJEPA Configurations

To assess the efficiency at which GeoJEPA can benefit from the combination of different modalities, we train GeoJEPA in three configurations which are summarised in Table 15.

**Table 15:** Configuration comparison in terms of supported modalities.

| Model | Modalities |
|---|---|
| GeoJEPA-T | OSM Tags |
| GeoJEPA-TI | OSM Tags + Aerial Imagery |
| GeoJEPA-GT | OSM Geometries + OSM Tags |
| GeoJEPA-GTI | OSM Geometries + OSM Tags + Aerial Imagery |

## 7.6.11  GeoJEPA Training Regime

JEPA is not a straightforward architecture to train, as outlined in Section 4.6.2. In an effort to address the limited interpretability of training this architecture, this section provides visualisations of different phenomena, some of which remain unexplained.

As an example of complexity, the reconstruction loss is optimised to decrease, but the model learns when the loss is increasing. To address this, we adopt the methods proposed by Thimonier et al [89]. During training, we continuously monitor the model by saving visualisations of latent representations at various stages of the pipeline once every fifth epoch. Additionally, we visualise the representation space by applying PacMAP dimensionality reduction the an entire batch of target tokens. These techniques serve as invaluable tools for debugging the architecture and developing a deeper intuition of the training process.

In Figure 43, we depict the latent representations of the target encoder output, which are particularly relevant as they are used in downstream tasks.

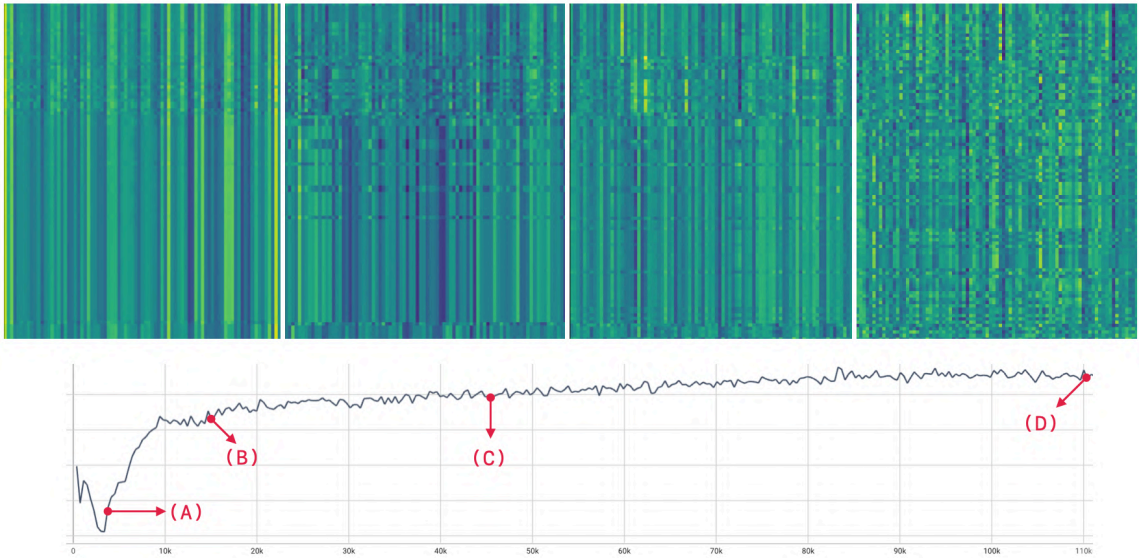

**Figure 43:** A visualisation of GeoJEPA-T latent representation of a certain tile region (top) and reconstruction loss (bottom) during training. Initially, the model collapses (A), resulting in a near-zero loss. It subsequently escapes the collapsed state, but the representations in intermediate stages (B and C) still exhibit distinct patterns, indicating that the model has not yet learned to differentiate between tokens. Upon completion of training, the representations appear more informative and less characterised by such patterns (D). For visibility, the representation visualisations are zoomed in on 75 out of 384 hidden dimensions and 100 out of 350 entity representations. Each row in the representation corresponds to one map entity. Note that some rows are similar even in D, which is reasonable as these tokens, in this case, correspond to close to identical map entities (e.g. all houses).

In the dimensionality reduction of target tokens (Figure 44), we can observe how different modalities start to merge. As aerial images of a roundabout should be similar to the corresponding tags and geometry, we believe that this effect is wanted, but we have not excluded other possible explanations.

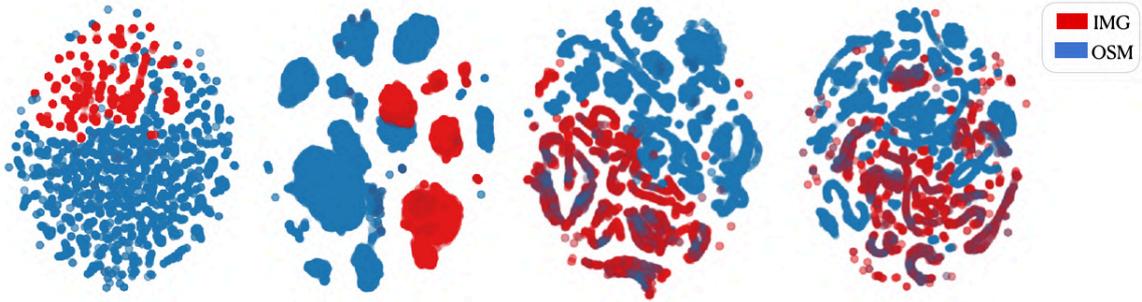

**Figure 44:** A PacMAP [119] dimensionality reduction of all map entity and image patch tokens in one specific batch (96 tiles). From left to right, the images correspond to random initialisation (1), collapse (2), escape (3), and convergence (4).

Furthermore, after 80-90% of the training period, we start to see some artefacts in the output representations of the target encoder. These only appear when training for longer periods. We hypothesise that they might be akin to e.g. *register tokens* [49], or stem from some fault in our training process, but do not inspect them any further. GeoJEPA-GTI does not have these artefacts.

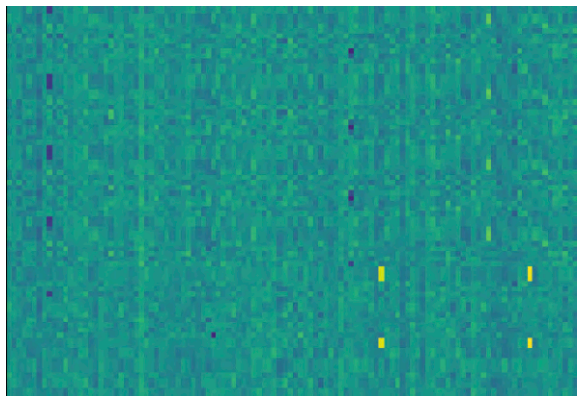

**Figure 45:** GeoJEPA "artefacts" in target encoder output (the yellow and black dots).

We can also observe that while the variance loss goes to zero, the covariance loss does not. This is undesirable as it indicates that there are hidden features that are linearly dependent, meaning that they encode the same information.

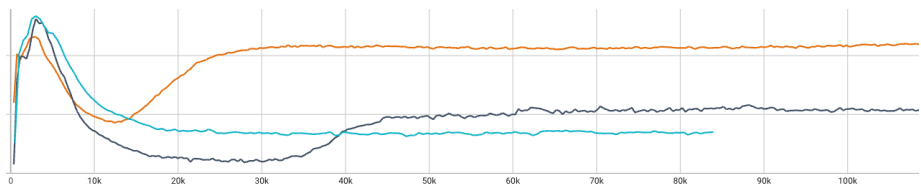

**Figure 46:** Covariance loss during pretraining of GeoJEPA-T (orange), GeoJEPA-TI (black), and GeoJEPA-GTI (blue).

# Chapter 8

# Quantitative Analysis

*In this chapter we present the quantitative results of our test suite, together with a light analysis. For ease of interpretability, the results are displayed as $L_1$-loss (mean absolute error), as introduced in Section 6.1. In all tables, a lower score is better.*

## 8.1  Significance

The primary goal of this quantitative evaluation is to assess the versatility and performance of urban region representations across various downstream tasks and establish supervised and self-supervised baselines against which GeoJEPA can be compared. Although all tasks are synthesised from OSM data, and some supervised models directly incorporate this data in their input, they provide an estimate of the upper bound for achievable performance.

Despite using out-of-training data to synthesise the tasks, GeoJEPA's performance on these tasks cannot guarantee any cross-domain applicability. Instead, the evaluation serves to quantitatively determine whether the JEPA pretraining objective successfully encodes information relevant to solving diverse tasks within this domain.

## 8.2  Dummy Reference

To create a context in which the scores on these tasks can be understood, we calculate the best fixed MAE by predicting the median of the *Huge* training set on the test sets. As can be observed in Table 16, the small dataset has a different distribution from the *Huge* dataset, which will explain the large difference in scores between the two in the following sections. Since the *small* dataset only has 400 samples per task, the results are not as representative as on the *Huge* dataset.

**Table 16:** Best fixed guess on different distributions, subsequently referred to as the *Dummy* model.

| Model | Buildings | Max Speed | Traffic Signals | Bridge | Car Bridge |
|---|---|---|---|---|---|
| Huge (Train) | 31.27 | 43.06 | 3.08 | 0.32 | 0.38 |
| Huge (Test) | 31.87 | 46.83 | 3.10 | 0.32 | 0.38 |
| Small (Test) | 45.14 | 41.69 | 4.78 | 0.66 | 0.90 |

## 8.3 Harmonic Mean

To ensure a consistent evaluation of model performance on tasks with varying baseline errors, it is essential to calculate a single score per model. Therefore, we employ the harmonic mean as it emphasises low values more than the arithmetic mean. For each task, we calculate the performance ratio between the model's MAE score and the best performance on the task. We then compute the harmonic mean over these ratios. If a model is the best performer on all tasks, it receives a score of 1.0 and all subsequent models receive lower scores.

$$H(x_1, x_2, ..., x_n) = \frac{n}{\frac{1}{x_1} + \frac{1}{x_2} + ... + \frac{1}{x_n}} \tag{5}$$

**Figure 47:** *Harmonic Mean:* $x_1$ to $x_n$ correspond to individual samples, where $n$ is the total number of samples.

## 8.4 Image Models

In this section, we present the results from the *frozen backbone* image models and the *fine-tuned* image models.

### 8.4.1 Frozen Backbone

Results from the *frozen backbone* models show that the largest, domain-specific model, *ScaleMAE* [107], with a multi-layer perceptron probe, performs on tie with the smaller ViT-B/16 model. Additionally, we observe that the linear probe falls behind more than usual on the *max_speed* task which is modelled in a non-linear way.

**Table 17:** Comparison of *frozen backbone* image models sorted by the harmonic mean of performance relative to the best performance per task. Listing relative $L_1$ compared to ViT-B/16 ($\downarrow$) and sorted by harmonic mean ($\uparrow$).

| Model | Buildings | Max Speed | Traffic Signals | Bridge | Car Bridge | Harmonic Mean |
|---|---|---|---|---|---|---|
| ScaleMAE (MLP) | <u>−8.7%</u> | +15.7% | <u>−4.2%</u> | −5.1% | +3.3% | 0.958 |
| ViT-B/16 (MLP) | 17.756 | <u>18.044</u> | 2.393 | 0.266 | 0.235 | 0.957 |
| ViT-B/16 (LIN) | +16.7% | +50.9% | −1.8% | <u>−6.7%</u> | <u>−1.2%</u> | 0.860 |
| ScaleMAE (LIN) | +12.1% | +76.2% | +2.7% | −4.8% | +5.4% | 0.813 |
| EfficientNet (MLP) | +81.4% | +113.2% | +31.0% | +24.7% | +65.3% | 0.589 |
| ResNet (MLP) | +81.4% | +113.2% | +31.0% | +24.7% | +65.3% | 0.589 |
| Dummy | +79.5% | +159.5% | +28.7% | +20.4% | +61.4% | 0.566 |
| EfficientNet (LIN) | +81.4% | +256.6% | +31.0% | +24.7% | +65.3% | 0.504 |
| ResNet (LIN) | +81.4% | +256.7% | +31.0% | +24.7% | +65.3% | 0.504 |

Since the multi-layer regressors consistently outperform the linear regressors across all models, we will exclude the linear regressors (LIN) from subsequent comparisons and refer to the multi-layer regressors (MLP) without the postfix.

## 8.4.2 Autogluon Fine-Tuning

This section presents the results of automatic model selection and fine-tuning using autogluon. In Table 18, the results of the three AutoGluon-trained image models, with 197M, 97M, and 4M parameters, can be observed in comparison to the frozen backbone models ViT-B/16 and ScaleMAE.

Notably, there are gaps in the table caused by the multimodal model *AG-TI-1800s* failing to complete the first half epoch within the time limit. Additionally, we observe that only the two largest AutoGluon models surpass the frozen ViT-B/16 model. Given that AutoGluon's fine-tuning process and data loading are significantly slower than training the frozen backbone regressors, it is likely that further extending the time limit could improve performance.

**Table 18:** Autogluon fine-tuned models compared to frozen ViT-B/16 and ScaleMAE. Listing absolute $L_1$ (↓) and sorted by harmonic mean (↑).

| Model | Buildings | Max Speed | Traffic Signals | Bridge | Car Bridge | Harmonic Mean |
|---|---|---|---|---|---|---|
| AG-197M-3600s | 20.09 | 15.44 | **2.24** | 0.25 | 0.21 | 0.9 |
| AG-97M-900s | 25.52 | **14.26** | 2.43 | **0.23** | **0.18** | 0.88 |
| ViT-B/16 | 17.76 | 18.04 | 2.39 | 0.27 | 0.24 | 0.85 |
| ScaleMAE | **16.21** | 20.88 | 2.29 | 0.25 | 0.24 | 0.84 |
| AG-4M-900s | 35.45 | 18.41 | | 0.27 | 0.26 | 0.66 |
| AG-4M-900s | 19.7 | 30.84 | 2.81 | 0.36 | 0.34 | 0.62 |
| AG-4M-600s | 19.88 | 31.54 | 2.91 | 0.39 | 0.35 | 0.6 |
| Dummy | 31.87 | 46.83 | 3.08 | 0.32 | 0.38 | 0.49 |
| AG-TI-3600s | 63.23 | 93.04 | 2.95 | 0.34 | 0.34 | 0.33 |
| AG-TI-1800s | | 90.43 | 3.02 | | | 0.26 |

## 8.5 Large Language Models

The table below presents the results of ChatGPT 4o and ChatGPT 4o-mini on our test suite, using a smaller dataset as detailed in Section 6.3. Additionally, the table is complemented with the image models ViT-B/16 and ScaleMAE for reference. In all tasks, ChatGPT 4o outperforms 4o-mini, likely due to model size. Furthermore, the language models are mostly outperformed by the image-only models but do put up respectable results on all but the *max_speed* task, where the LLMs experience great difficulties which are further analysed in Section 9.5. The regression heads of the image models are trained on the *Huge* training dataset, and then evaluated on the *small* test set, no overlap exists.

**Table 19:** Results from ChatGPT 4o and ChatGPT 4o-mini compared to frozen ViT-B/16 and ScaleMAE. Listing absolute $L_1$ (↓) and sorted by harmonic mean (↑).

| Model | Buildings | Max Speed | Traffic Signals | Bridge | Car Bridge | Harmonic Mean |
|---|---|---|---|---|---|---|
| ViT-B/16 | 28.34 | 19.15 | 3.12 | 0.23 | 0.22 | 0.96 |
| ScaleMAE | 24.44 | 21.85 | 2.94 | 0.24 | 0.26 | 0.93 |
| GPT-4o | 29.92 | 39.33 | 3.0 | 0.24 | 0.23 | 0.79 |
| GPT-4o-mini | 36.94 | 70.97 | 3.25 | 0.35 | 0.37 | 0.53 |
| Dummy | 45.14 | 41.69 | 4.78 | 0.66 | 0.9 | 0.4 |

# 8.6 Tag-based Models

Here we present the results of the supervised tag-only models, meaning the Tagformer models and the TagCountMLP model. From Table 20, a few conclusions can be made. First, as expected, the semantic Tagformer does not retain all information. While this is suboptimal for some of the tasks presented, it may still be advantageous due to its potential for generalisation to tags not present in the training data. The limitation is most notable on the *max_speed* task, where we hypothesise that the text embedding distance between different speeds is quite low (i.e. 30km/h $\simeq$ 50km/h), or not included in the lookup table. However, the semantic capability seems to directly benefit the *traffic_signals* task. Furthermore, note that the results are affected by the large variability between training runs. Even with Bayesian optimisation, cosine learning rate decay and moderately long training sessions, it is common for the Tagformer to diverge or not learn optimally.

**Table 20:** Results from tag-based models on masked and unmasked tasks. Listing relative $L_1$ compared to Tagformer ($\downarrow$) and sorted by harmonic mean ($\uparrow$).

| Model | Buildings | Max Speed | Traffic Signals | Bridge | Car Bridge | Harmonic Mean |
|---|---|---|---|---|---|---|
| *Models trained on unmasked data* | | | | | | |
| Tagformer-P | −25.1% | −5.8% | +15.8% | −31.6% | −5.0% | 0.74 |
| Tagformer | 3.02 | 0.74 | 0.12 | 0.05 | 0.01 | 0.64 |
| Tagformer-S | −61.5% | +288.7% | −37.1% | −8.8% | +36.4% | 0.56 |
| Tagformer-SP | −36.1% | +293.0% | −25.1% | −2.1% | +80.6% | 0.48 |
| TagCountMLP | +185.4% | +251.4% | +657.3% | +2.6% | +505.4% | 0.16 |
| *Models trained on masked data* | | | | | | |
| (M) Tagformer-P | +0.5% | +1.0% | +2.3% | −14.7% | −5.7% | 0.99 |
| (M) Tagformer | 13.45 | 8.03 | 1.84 | 0.16 | 0.14 | 0.95 |
| (M) Tagformer-SP | +12.5% | +13.3% | −2.4% | −12.4% | +20.0% | 0.9 |
| (M) Tagformer-S | +11.8% | +20.8% | −1.4% | +11.8% | +28.0% | 0.83 |
| (M) TagCountMLP | +4.6% | +11.8% | −2.2% | +35.8% | +40.4% | 0.8 |

The TagCountMLP performance is noticeably inferior to the Tagformer. However, as can be observed in Table 21, it outperforms ViT-B/16 by a significant margin, even with masked datasets. As the masked datasets are relatively hard (all buildings and highly related entities removed), this indicates that it might be easier to predict missing map entities from OSM data than from aerial imagery.

**Table 21:** TagCountMLP compared to VIT-B/16. Listing absolute $L_1$ ($\downarrow$) and sorted by harmonic mean ($\uparrow$)

| Model | Buildings | Max Speed | Traffic Signals | Bridge | Car Bridge | Harmonic Mean |
|---|---|---|---|---|---|---|
| TagCountMLP | **8.63** | **2.6** | **0.92** | **0.05** | **0.05** | 1.0 |
| (M) TagCountMLP | <u>14.06</u> | <u>8.98</u> | <u>1.8</u> | <u>0.21</u> | <u>0.19</u> | 0.32 |
| ViT-B/16-MLP | 17.76 | 18.04 | 2.39 | 0.27 | 0.24 | 0.23 |
| Dummy | 31.87 | 46.83 | 3.08 | 0.32 | 0.38 | 0.13 |

## 8.7 General Purpose Models

This section compares the performance of the general-purpose models. First, it is meaningful to compare the models in terms of their input data, compiled in Table 22.

**Table 22:** Model comparison in terms of supported modalities.

| Model | Modalities |
|---|---|
| TagAE | OSM Tags |
| TagPool | OSM Tags |
| ViT-B/16 | Aerial Imagery |
| GeoJEPA-T | OSM Tags |
| GeoJEPA-GT | OSM Geometries + OSM Tags |
| GeoJEPA-TI | OSM Tags + Aerial Imagery |
| GeoJEPA-GTI | OSM Geometries + OSM Tags + Aerial Imagery |

From the quantitative results of the general-purpose models, displayed in Table 23, we observe that the GeoJEPA variations are not the best at encoding fine-grained details, with TagPool achieving the best overall result. Furthermore, contrary to our initial hypothesis, the addition of image and geometry data does not improve the results. Especially the addition of image data impairs the model's performance significantly, which is discussed further in Section 10.3.

**Table 23:** General purpose models with ViT-B/16 for reference. All models use multi-layer heads. Listing absolute $L_1$ ($\downarrow$) and sorted by harmonic mean ($\uparrow$).

| Model | Buildings | Max Speed | Traffic Signals | Bridge | Car Bridge | Harmonic Mean |
|---|---|---|---|---|---|---|
| TagPool | 7.15 | <u>12.38</u> | **1.01** | **0.1** | **0.1** | 0.84 |
| **GeoJEPA-T** | 7.13 | **11.5** | <u>1.68</u> | 0.17 | <u>0.14</u> | 0.65 |
| TagformerLMAE | 5.43 | 16.05 | 1.72 | <u>0.17</u> | 0.16 | 0.64 |
| **GeoJEPA-GT** | 4.19 | 13.34 | 1.83 | 0.21 | 0.2 | 0.61 |
| TagAE | **3.86** | 20.63 | 2.02 | 0.23 | 0.25 | 0.52 |
| **GeoJEPA-TI** | <u>4.02</u> | 20.73 | 2.07 | 0.24 | 0.24 | 0.51 |
| **GeoJEPA-GTI** | 4.4 | 20.76 | 2.07 | 0.25 | 0.25 | 0.5 |
| ViT-B/16 | 17.76 | 18.04 | 2.39 | 0.27 | 0.24 | 0.37 |
| Dummy | 31.87 | 46.83 | 3.08 | 0.32 | 0.38 | 0.22 |

The relative difference between GeoJEPA-T and TagPool decreases on the *Masked* datasets which are supposed to be more difficult to solve. This can be observed in Table 24.

**Table 24:** Top performers in comparison to supervised TagCountMLP model on masked and unmasked datasets. TagformerLMAE fails to produce embeddings for the masked tasks. Listing absolute $L_1$ ($\downarrow$) and sorted by harmonic mean ($\uparrow$).

| Model | Buildings | Max Speed | Traffic Signals | Bridge | Car Bridge | Harmonic Mean |
|---|---|---|---|---|---|---|
| TagCountMLP | 8.63 | 14.43 | **0.92** | **0.05** | **0.05** | 0.77 |
| TagPool | 7.15 | 12.38 | <u>1.01</u> | <u>0.1</u> | <u>0.1</u> | 0.61 |
| GeoJEPA-T | 7.13 | **11.5** | 1.68 | 0.17 | 0.14 | 0.44 |
| TagformerLMAE | 5.43 | 16.05 | 1.72 | 0.17 | 0.16 | 0.43 |
| | | | | | | |
| *Masked datasets* | | | | | | |
| (M) TagCountMLP | 14.06 | 20.56 | 1.8 | 0.21 | 0.19 | 0.31 |
| (M) TagPool | 26.17 | 13.06 | 1.91 | 0.18 | 0.18 | 0.29 |
| (M) GeoJEPA-T | 27.79 | <u>11.79</u> | 1.96 | 0.19 | 0.17 | 0.28 |
| (M) TagformerLMAE | - | - | - | - | - | - |

# Chapter 9

# Qualitative Analysis

*In this chapter we use qualitative techniques to understand and assess the embedding performance of our self-supervised models.*

## 9.1 Significance

As outlined in Section 1.2, identifying semantically similar regions and map entities presents substantial opportunities for debugging and analysing global-scale geographic information systems. To enable such applications, the representations must demonstrate equivariance with respect to the input data, ensuring that the embeddings of semantically similar inputs are positioned closely in the high-dimensional embedding space. The linear representation hypothesis suggests a correlation between this property and high performance on quantitative tasks, however, it can be discussed whether it is a requirement. For instance, while the ability to recall fine-grained details may benefit quantitative performance, it does not necessarily align with high-level relational and contextual awareness. To investigate this further, we qualitatively explore the learned representations by examining latent space neighbours with the k-nearest neighbour (kNN) algorithm.

## 9.2 Region Representations

From the complete dataset, 25,000 tiles are embedded with TagPool, ViT-B/16, ScaleMAE, TagformerLMAE, and GeoJEPA models. From those, 300 tiles are randomly, but deterministically, selected as queries and a kNN search is performed in embedding space with $k = 8$. The neighbours are then manually evaluated on their similarity to the query tile based on their corresponding aerial imagery, geometry, and tags. In this section, we present evaluations corresponding to five distinct query regions. These examples are deliberately selected based on the presence of prominent features which can serve as assessment criteria. All models perform highly varied results, and we do our best to present them without bias. For detailed query results and evaluation criteria, see Appendix E. Notably, the *Motorway-fields* task is evaluated on three criterias; the presence of motorways, fields, and off-ramps, which are then averaged.

Table 25 presents the results of the qualitative evaluation, highlightning variations in model performance across the five tasks. Notably, the TagPool and TagformerLMAE models, which demonstrate strong performance in the quantitative evaluation, struggle to identify similar regions and rank among the lowest-performing models in this evaluation.

We hypothesise that this discrepancy can be explained by the models' focus on fine-grained details, benefitting performance in entity-wise reconstruction and synthetic tasks, where accurately recalling input data properties is crucial. However, this does not necessarily align with effectiveness in region and entity similarity as those supposedly require contextual awareness and higher-level semantic understanding.

**Table 25:** Compiled qualitative results. Each model-task combination receives a score based on the number of the eight nearest neighbours that are similar to the query region, ordered by mean score.

| Model | Airfield | Airport | Canal | Motorway-Forest | Motorway-Fields | Mean Score |
|---|---|---|---|---|---|---|
| GeoJEPA-T | 7 | **7** | 2 | 1 | **6.6** | **4.72** |
| GeoJEPA-GT | **8** | 2 | 5 | 2 | 6.0 | 4.60 |
| GeoJEPA-GTI | 0 | 1 | 6 | **6** | 2.6 | 3.12 |
| ViT-B/16 | 0 | 2 | 4 | 3 | 2.6 | 2.32 |
| TagPool | 0 | 1 | 3 | 0 | 6.0 | 2.00 |
| ScaleMAE | 0 | 2 | 1 | 2 | 3.0 | 1.60 |
| GeoJEPA-TI | 0 | 1 | 4 | 0 | 2.6 | 1.52 |
| TagformerLMAE | 0 | 1 | 0 | 3 | 1.6 | 1.12 |

Notably, some tasks seem to benefit from image-based information, for instance, the *Motorway-forest* task. Overall, GeoJEPA demonstrates limited effectiveness in integrating visual information, with GeoJEPA-TI performing at the bottom and GeoJEPA-GTI underperforming compared to the -T and -GT variants. When incorporating visual information, GeoJEPA's performance on the *Airfield* task drops to zero. This airfield is hard to distinguish using the image alone, yet GeoJEPA appears to disproportionately rely on it. In Table 26, we display the airfield region and the best matches produced by some different models.

**Table 26:** The *Airfield* region and its immediate surroundings. While GeoJEPA-T finds another airfield, GeoJEPA-GTI finds fields and a road, and TagPool finds a motorway running through a city area.

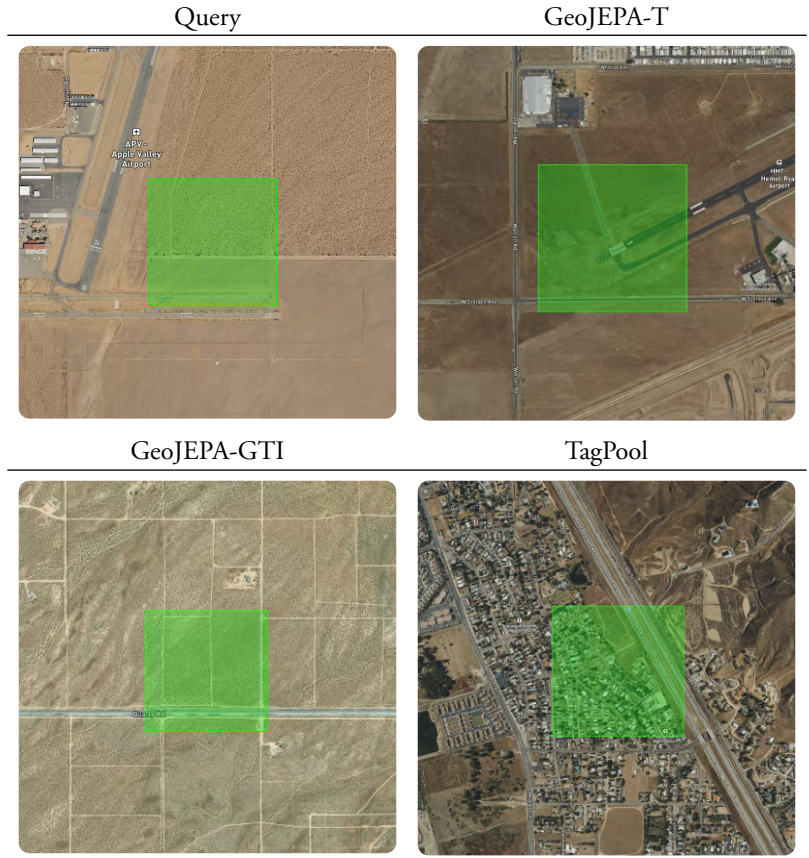

Furthermore, analysing the GeoJEPA results unveils a large variance throughout training, displayed in Table 27. After 100 epochs, GeoJEPA-T performs well on the *Canal* task but poorly on the *Airport* task. However, after all 300 epochs, the trend is reversed, the model has now learned the importance of the airport but has forgotten about the canal. The same pattern can be seen for GeoJEPA-GTI. We are not able to explain this phenomenon.

**Table 27:** Comparison of qualitative GeoJEPA results from intermediate checkpoints (100 and 170 epochs) against completed training (300 epochs).

| Model | Airfield | Airport | Canal | Motorway-Forest | Motorway-Fields | Mean Score |
|---|---|---|---|---|---|---|
| GeoJEPA-T 300e | 7 | **7** | 2 | 1 | **6.6** | **4.72** |
| GeoJEPA-T 100e | **8** | 1 | **8** | 0 | 6.3 | 4.66 |
| GeoJEPA-GTI 170e | 0 | 1 | 6 | 6 | 2.6 | 3.12 |
| GeoJEPA-GTI 300e | 0 | 2 | 1 | 5 | 2.6 | 2.12 |

# 9.3 Map Entity Representations

As outlined in Section 7.6.1, JEPA token-level representations have not been explored in previous research. To evaluate their suitability as map entity representations, we present selected examples of both good and poor representations. We examine GeoJEPA-T, GeoJEPA-GTI, and TagPool map entity embeddings. In Table 28 and Table 29 examples of GeoJEPA producing good map entity representations can be observed.

**Table 28:** kNN-search using a car bridge over a motorway as the query. The retrieved nearest neighbours are ranked from left to right. The TagPool model identifies a footway bridge (neighbour 1) and bridges over water (neighbours 2 and 3). GeoJEPA-T retrieves a guardrail on a bridge over a motorway (neighbour 4).

Query

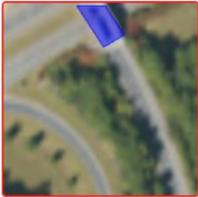

TagPool

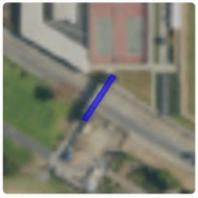 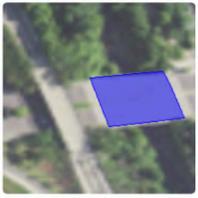 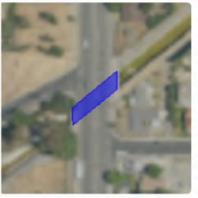 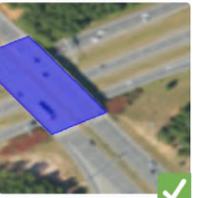

GeoJEPA-T

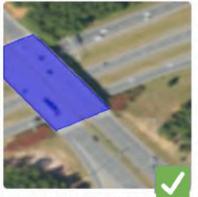 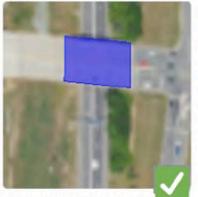 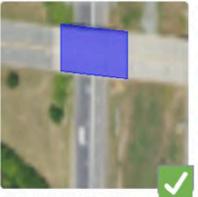 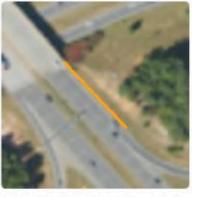

GeoJEPA-GTI

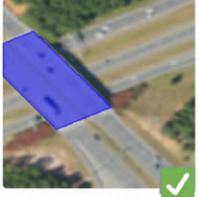 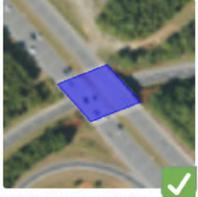 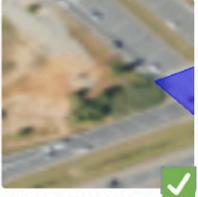 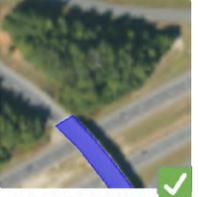

The TagPool model, relying exclusively on OSM tags without incorporating spatial context, demonstrates limitations in this task. In contrast, GeoJEPA, which integrates contextual information, offers both advantages and challenges. For instance, while the guardrail shares a similar context to the query entity, more similar candidates were retrieved by GeoJEPA-GTI. In Table 29, a similar result can be observed. The surroundings are important even in the case of understanding a single map entity.

**Table 29:** kNN-search of a house in the outskirts of a forest.

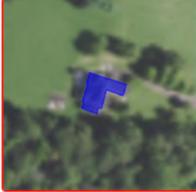

In Figure 48 and Figure 49, an instance can be observed where GeoJEPA-T exhibits notably poor performance due to an overemphasis on the surrounding context.

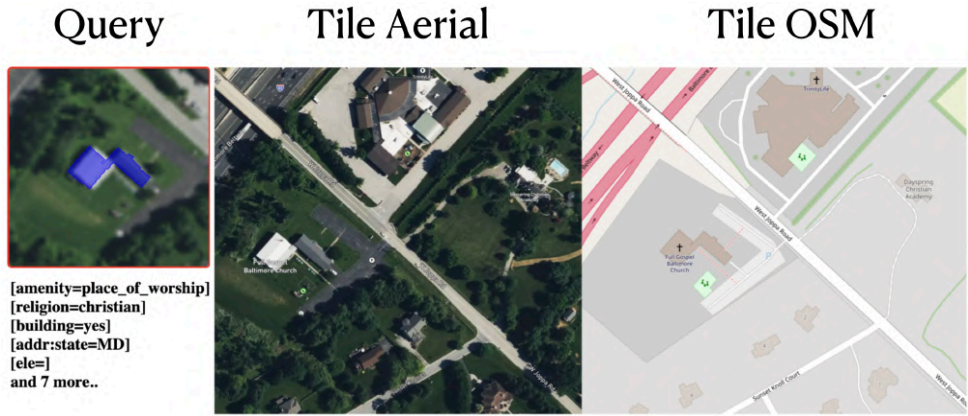

**Figure 48:** A query of a church (left) and the tile containing it (right). Notably, there is a church just across the road which should be semantically similar.

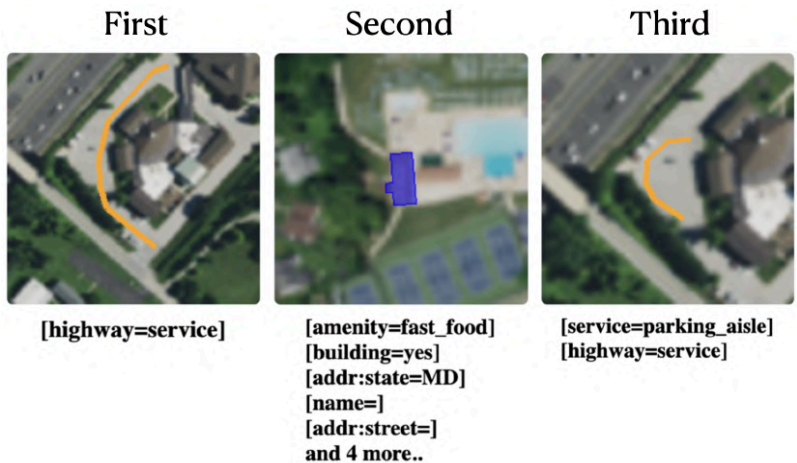

**Figure 49:** The result of the church query. Best matches are; a service road, a fast food restaurant, and a parking aisle.

To conclude this section, JEPA token-level representations encode semantic information related to both the specific map entity and its surroundings. However, in some cases, contextual information dominates entity-specific information, compromising retrieval accuracy as observed in Figure 49.

## 9.4  GeoJEPA Ablation

Additionally, we conduct multiple experiments to evaluate different hyperparameter settings for GeoJEPA, with the directly comparable results presented in this section. Specifically, we evaluate the impact of the variance-invariance regularisation (VICREG), and of using a larger effective batch size of 960 instead of 576 tiles. Each model is trained during 300 epochs and finishes in 42-44 hours depending on configuration. The qualitative and quantitative results are presented in Table 30 and Table 31 respectively.

**Table 30:** Qualitative comparison of different GeoJEPA-T experiments.

| Model | Airfield | Airport | Canal | Motorway-Forest | Motorway-Fields | Mean Score |
|---|---|---|---|---|---|---|
| Standard | 6 | 7 | **3** | **1** | 7 | **4.8** |
| w/o VICREG | 4 | 3 | **3** | 0 | 6 | 3.2 |
| Larger Batches | 7 | 2 | 1 | 0 | 5.33 | 3.07 |

**Table 31:** Quantitative comparison of different GeoJEPA-T experiments.

| Model | Airfield | Airport | Canal | Motorway-Forest | Motorway-Fields | Mean Score |
|---|---|---|---|---|---|---|
| Standard | 6.64 | **10.92** | **1.75** | <u>0.19</u> | **0.17** | 0.97 |
| w/o VICREG | <u>6.4</u> | 14.4 | <u>1.77</u> | **0.18** | <u>0.18</u> | 0.91 |
| Larger Batches | **6.04** | <u>13.25</u> | 1.81 | 0.21 | 0.18 | 0.91 |

As presented, variance-invariance regularisation boosts performance in both tasks. However, larger batch size impairs performance, suggesting that JEPA needs a certain number of steps to converge.

# 9.5 Large Language Models

As seen in Section 8.5, the greatest discrepancy between ChatGPT and a traditional image model is found for *max_speed*. Looking at Table 32 and Table 33, it is clear that both models perform reasonably well when the road is clearly visible. If not, the smaller 4o-mini tries to rely on road markings, while the larger model is more proficient at using context clues. Finally, both models struggle with roads in forests where the road is partially covered by leaves and shadows.

**Table 32:** Handpicked images illustrating successful predictions by ChatGPT on the *max_speed* task. The reasoning is abbreviated.

| Image | Correct | 4o | 4o reasoning | 4o-mini | 4o-mini reasoning |
|---|---|---|---|---|---|
| 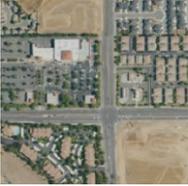 | 64 | **60** | Commercial area, wide roads and intersection | **50** | Intersection and surrounding roads, urban with multiple lanes |
| 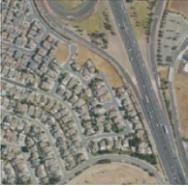 | 104 | **120** | Highway, multiple lane, interstate or major road | **100** | Highway, multiple lane |

**Table 33:** Deliberately selected examples of suboptimal ChatGPT performance on the *max_speed* task, illustrating the difference between 4o and 4o-mini. The reasoning is abbreviated. *The model is instructed to output a score of −100 if no road is visible.*

| Image | Correct | 4o | 4o reasoning | 4o-mini | 4o-mini reasoning |
|-------|---------|-----|--------------|---------|-------------------|
| 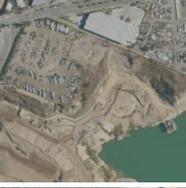 | 72 | 30 | Construction, dirt roads | −100 | Construction, no visible roads or road markings |
| 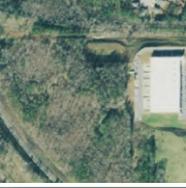 | 80 | −100 | Forest, a large building, no visible roads | −100 | Wooden area, no visible roads |
| 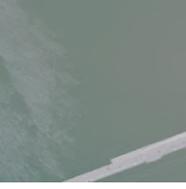 | 104 | 100 | Bridge over water, based on typical architecture | −100 | No visible road, only water and a bridge |

# Chapter 10

# Limitations and Future Work

*In this chapter, we discuss the limitations of the GeoJEPA model, and propose some areas of development that are interesting to explore further.*

## 10.1   External Validity

The geospatial domain spans diverse data modalities and formats, resulting in significant variability across datasets. As no existing dataset met our requirements in terms of modalities and size, we chose to construct our own dataset, tasks, reference models, and evaluation frameworks. This limits external validity and is a situation that, unfortunately, is not uncommon in geospatial AI. While this limits external validity, it provides a controlled environment suitable for exploration.

To support reproducibility and validity, we release our datasets, tasks, model weights, and code. Moving forward, we plan to evaluate GeoJEPA on real-world, non-synthetic tasks and benchmark it against existing geospatial representation models.

## 10.2   GeoJEPA Training Regime

JEPA hyperparameter tuning is particularly challenging to solve as the dynamics are only partially understood. For instance, GeoJEPA-T appears to perform slightly better at the quantitative evaluation after 100-200 epochs than after all 300 epochs. We are yet to identify the reason for this. Additionally, GeoJEPA exhibits high sensitivity to hyperparameter variations, necessitating extensive experimentation. As such, future research can improve training efficiency and hyperparameter-tuning by exploring JEPA training dynamics and how performance develops over time. Developing methods to assess and project performance during early training based on relevant metrics remains an important direction for future work. Likewise, extensive ablation studies are essential to gain a deeper understanding of the JEPA optimisation landscape.

## 10.3 Tokenisation Problems

As simple linear or multi-layer heads can solve complex tasks when combined with a frozen backbone, we initially assumed that using pretrained models in the tokenisation step would facilitate access to rich semantic information and accelerate convergence. However, we can now observe that this assumption is ill-suited for the JEPA pretraining objective.

> **Major Flaw 1:** JEPA relies on high variability between context and target tokens to create challenging training objectives. In contrast, the latent image patch representations generated by ViT-B/16 contain global information, reducing reconstruction complexity and undermining training dynamics.

Conversely, the tag and geometry encoders employed during the tokenisation step do not encode global information. As such, frozen backbones are still relevant for JEPA tokenisation. Nevertheless, introducing global information tokens should be approached cautiously in future research to ensure alignment with the JEPA objective.

## 10.4 Multimodal Performance

Although we hypothesised that incorporating additional modalities would improve performance by giving access to more information, our results indicate a decline in both quantitative and qualitative performance when more modalities are included. This can be due to the argument in Section 10.3 but might also arise from suboptimal training procedures. The inherent differences in variance and covariance across modalities could potentially disrupt optimal masking strategies and hyperparameters. For instance, image data might be more redundant than map entity data, necessitating higher masking ratios to reach optimal performance. A key assumption in the design of GeoJEPA is that JEPA would inherently learn the significance of each modality and token by learning their contribution to the reconstruction of masked tokens. Therefore, it is crucial to avoid introducing any additional biases that could interfere with the learning process.

> **Major Flaw 2:** While we impose no explicit control over the relative importance of modalities, our masking strategies, tokenisation steps, and intrinsic modality properties may significantly influence each modality's contribution to the latent reconstruction loss.

For instance, map entity tokens have an average sequence length of 84, and range between five and 1250, while the aerial image consistently contributes 198 tokens. Consequently, the reconstruction objective may inherently prioritise the accurate reconstruction of image tokens, especially in cases where the number of map tokens is small. This highlights the need for further research on how loss functions can be designed to promote unbiased multimodal learning.

## 10.5  Sequence Length Variability

Sequence length variability is a major limitation of model performance, even with flash-attention optimised kernels. Using one token for every map entity and one for every image token adds up quickly, especially for larger tiles. After removing the worst high-content regions, the maximum sequence length is 1250 map entity tokens and 197 image tokens. Expanding this architecture to larger context windows and additional modalities could lead to an explosive growth in sequence length. However, large-scale language models and sparse attention have shown that context windows of 128k-1M tokens are feasible, but perhaps not commonly used during training. In the name of training efficiency, a more pronounced problem may be the issue of sequence length variability.

## 10.6  Out-of-Region Context

Map entities that are close to, or intersected by an edge are naturally hard to distinguish in low-resolution aerial imagery. As such, using images to solve the quantitative tasks may be inherently harder than when using OSM data.

Additionally, the representations of these entities, and entities in tile corners, have less potential for encoding relevant context as most of it is hidden. Hexagonal tiles, as explored by Woźniak and Szymański [38], may be better suited for this purpose.

In our approach, the representation of a region can depend only on the geospatial information that is directly covered by it, ignoring long-range relationships. Conversely, as regions are relatively small, their semantic meaning should potentially be greatly affected by the content of their neighbouring regions, such as if they contain industrial buildings or beaches and leisure areas. A potential way to mitigate this is to apply a Gaussian blur of desired kernel size over the computed representations, linearly interpolating neighbouring information into the local representations. However, this does not account for the fact that the influence range and weight of different semantic content may vary. In future research, methods for handling regions and relationships of different scales would be of great benefit to the field.

## 10.7  Map Entity Representations

Our evaluation of token-level representations has notable limitations, specifically the lack of a quantitative assessment, which restricts our conclusions to subjective interpretations and small samples. Nevertheless, from our evaluation, GeoJEPA token-level representations are not optimally informative about the specific map entity but rather focus more on the regional context. While we do not want to manually control the distribution of contextual and entity-specific information in map entity representations, we currently have no way of knowing or assessing the ratio. It may be beneficial to be able to include only short-distance neighbours or weigh the influence of the context. Future work could explore how to observe, or impose more control over, what is included in the representation. For future research, we specifically want to encourage the exploration of token-level representations of JEPA pretrained models, in all domains.

## 10.8   Other Applications of Interest

As discussed previously, the semantic *Tagformer-S*, and in the extension *TagPool*, *TagformerLMAE*, and *GeoJEPA* theoretically allow free-text queries. Using such techniques, it would be interesting to construct a query engine where no existing sample is needed, e.g. by drawing roads or buildings and inputting free-text attributes to find matching regions or entities.

A model capable of capturing the correlations between aerial imagery, map data, and other available modalities, can be the base of a powerful tool for error detection, correction and map entity generation. Advancing research in this area has the potential to significantly improve the accuracy and comprehensiveness of digital maps.

# Chapter 11

# Conclusion

*In this chapter we present answers to our research questions and summarise our key findings with some concluding remarks.*

## 11.1 Research Questions

We now present our findings structured around the two central research questions that have guided this thesis, the first of which is stated as follows.

*A. How can SSL architectures be adapted to support unstructured and multi-modal map data?*

We find that adapting self-supervised architectures to this type of data presents significant challenges. Traditional approaches, such as input-space reconstruction and contrastive sampling, present notable shortcomings. In response, we present a novel adaptation of the Joint-Embedding Predictive Architecture (JEPA) to geospatial data by optimising a single-stream multimodal encoder utilising frozen unimodal models in the tokenisation step.

*B. How can the quality of (multimodal) geospatial representations be evaluated?*

We evaluate the quality in two distinct ways, quantitatively and qualitatively. Quantitatively, we assess the informational content within the embeddings by examining their effectiveness in solving synthetic downstream tasks derived from OSM. In the qualitative evaluation, we employ the k-nearest neighbours (kNN) algorithm to explore the representation space. By analysing regions and map entities that are close together in latent space, we can assess the models' performance on similarity-based applications (e.g. recommendation algorithms or clustering). We find this multi-faceted evaluation strategy to bring conflicting views on performance. For instance, while the TagformerLMAE and TagPool models achieve some of the best results in quantitative tasks, they demonstrate significant shortcomings in region similarity. This finding highlights the importance of employing multiple evaluation criteria, as relying on a single metric is likely to result in incomplete conclusions.

## 11.2 Concluding Remarks

This thesis focuses on self-supervised learning for multimodal geospatial data, specifically for urban regions and map entities. Through GeoJEPA, we demonstrate that JEPA is a versatile architecture and a promising direction for future research in geospatial data modelling. Additionally, the architecture eliminates all dependencies on complex augmentations and bias-inducing positive or negative sampling which currently dominate the literature.

While GeoJEPA performs well in the qualitative evaluation, our work reveals multiple limitations that necessitate further research. For instance, we observe that performance drops during the last half of training, indicating an incomplete understanding of the optimisation landscape. Additionally, while multimodal data is easily integrated, GeoJEPA does not significantly benefit from the extra information. We find that key factors behind this challenge include the use of pretrained tokenisation models that introduce global information, reducing the difficulty of latent reconstruction, and significant variance in sequence lengths across modalities, which complicates balancing the loss. Furthermore, we identify difficulties in evaluating and aligning contextual influence in constructing map entity representations, where GeoJEPA sometimes places a larger emphasis on the context than the entity itself.

Overall, we find that GeoJEPA demonstrates significant promise in geospatial learning and hope that this work will inspire further exploration of JEPA's potential for robust and versatile learning from heterogeneous multimodal data.

# Appendices

# Appendix A
# Training and Hyper-parameters

**Table 34:** GeoJEPA pretraining hyper-parameters. These are the best-performing configurations. However, not many trials have been run, instead, the parameters are based on the consensus in the literature with some random sampling. Except for the explicit experiments in Table 31, we do not have much measurable insight into parameter importance.

| Hyper-parameter | GeoJEPA-T | GeoJEPA-TI | GeoJEPA-GTI | GeoJEPA-GT |
|---|---|---|---|---|
| *model* | | | | |
| token_dim | 384 | 384 | 384 | 384 |
| encoder_depth | 12 | 12 | 12 | 12 |
| predictor_depth | 4 | 4 | 4 | 4 |
| predictor_dim | 256 | 256 | 256 | 192 |
| *optimization* | | | | |
| batch_size | 96 | 96 | 48 | 96 |
| group_size | 8 | 6 | 9 | 4 |
| effective batch size | 768 | 576 | 432 | 384 |
| epochs | 250 | 300 | 300 | 300 |
| momentum_init | 0.997 | 0.99 | 0.99 | 0.997 |
| momentum_end | 1.0 | 0.99999 | 0.99999 | 0.99999 |
| lr_warmup_frac | 0.1 | 0.1 | 0.1 | 0.1 |
| lr_base | 1e-3 | 8e-4 | 1e-3 | 1e-3 |
| lr_end | 1e-6 | 1e-6 | 1e-6 | 1e-6 |
| weight_decay_init | 0.04 | 0.05 | 0.05 | 0.04 |
| weight_decay_end | 0.4 | 0.1 | 0.1 | 0.4 |
| adam_beta1 | 0.9 | 0.9 | 0.9 | 0.9 |
| adam_beta2 | 0.95 | 0.95 | 0.95 | 0.95 |
| vicreg_beta | 0.05 | 0.01 | 0.01 | 0.02 |
| smooth_l1_beta | 2.0 | 1.0 | 1.0 | 2.0 |

# Appendix B
# LLM Queries

**Table 35:** The prompts sent to ChatGPT on each task.

| Task | Query |
|------|-------|
| Traffic Signals | For each provided satellite image, estimate the number of traffic signal crossings. Consider area density, road types, and visible pedestrian activity. Analyze each image individually. Images for analysis: |
| Bridge | For each satellite image, provide a confidence score between 0 and 1 indicating the presence of bridges (including overpasses and footbridges). Use 1 for definite presence and 0 for absence. Images for analysis: |
| Car Bridge | For each satellite image, provide a confidence score between 0 and 1 indicating the presence of bridges (including overpasses, excluding footbridges). Use 1 for definite presence and 0 for absence. Images for analysis: |
| Building Count | For each 224x224 pixel satellite image (representing 300x300 meters), count the number of visible buildings. Focus on rooftops, structures, walls, enclosures, and distinct outlines. Ignore roads, trees, and water bodies. Provide an accurate count based solely on visible buildings. Analyze each image individually. Images for analysis: |
| Max Speed | For each satellite image, identify the highest speed limit (km/h) on visible roads. If multiple roads have different limits, report the highest one. If no roads are visible, respond with –100. Base your assessment on road type, lane markings, signage, and infrastructure. Include a brief explanation for each determination. You may estimate any value; it doesn't need to be a multiple of 10. Pay attention to image edges. Images for analysis: |

# Appendix C
# Task Construction

**Table 36:** The designated tasks for evaluating performance. For masking, all counted tags are removed by default, and if all tags are removed, the entire feature is pruned. Max Speed is calculated by using the maximal value of any matching tag, while all tiles containing highways while not containing a max speed value are pruned from the dataset to avoid unclear data.

| Task | Tags Counted | Additional Masking Action |
|---|---|---|
| Traffic Signals | `traffic_signals=*`<br>`*=traffic_signals`<br>`crossing:signals=*` | Remove all features consisting of a single geometric point, only denoting the presence of a traffic signal. |
| Bridge | `bridge=*`<br>`*=bridge` | Tags containing `layer=*` are removed, due to the strong coupling it has to bridges. |
| Car Bridge | `(bridge=* or *=bridge)`<br>`and highway=*` | Tags containing `layer=*` are removed, due to the strong coupling it has to bridges. |
| Building Count | `building=*` | All features where `building=*` is present are removed. |
| Max Speed | `maxspeed=*` | |

# Appendix D
# All Quantitative Results

**Table 37:** All models, ViT-B/16 as reference.

| Model | Buildings | Max Speed | Traffic Signals | Bridge | Car Bridge | Harmonic Mean |
|---|---|---|---|---|---|---|
| *Task specific models* | | | | | | |
| Tagformer-P | −87.2% | <u>−96.1%</u> | −94.1% | <u>−88.1%</u> | <u>−96.8%</u> | 0.74 |
| Tagformer | −83.0% | −95.9% | −94.9% | −82.6% | −96.7% | 0.64 |
| Tagformer-S | <u>−93.4%</u> | −84.1% | <u>−96.8%</u> | −84.1% | −95.5% | 0.56 |
| Tagformer-SP | −89.1% | −83.9% | −96.2% | −83.0% | −94.0% | 0.48 |
| TagCountMLP | −51.4% | −85.6% | −61.7% | −82.1% | −79.9% | 0.16 |
| (M) Tagformer-P | −23.9% | −55.0% | −21.3% | −49.8% | −44.5% | 0.07 |
| (M) Tagformer | −24.3% | −55.5% | −23.1% | −41.1% | −41.2% | 0.07 |
| (M) Tagformer-SP | −14.8% | −49.6% | −24.9% | −48.4% | −29.4% | 0.07 |
| (M) Tagformer-S | −15.3% | −46.2% | −24.1% | −34.1% | −24.7% | 0.06 |
| (M) TagCountMLP | −20.8% | −50.2% | −24.8% | −20.0% | −17.4% | 0.06 |
| AG-97M-600s | +5.0% | N/A | N/A | +16.8% | +25.1% | 0.05 |
| AG-197M-3600s | +13.2% | +42.5% | +8.4% | +26.0% | +30.7% | 0.04 |
| AG-97M-900s | +7.3% | N/A | N/A | N/A | +25.9% | 0.04 |
| AG-4M-900s | +11.0% | +70.9% | +17.4% | +35.2% | +44.7% | 0.03 |
| AG-4M-600s | +12.0% | +74.8% | +21.8% | +47.7% | +46.7% | 0.03 |
| *General-purpose models* | | | | | | |
| TagPool | −59.7% | −31.4% | −57.7% | −62.6% | −58.4% | 0.09 |
| GeoJEPA-T | −74.2% | −36.3% | −29.6% | −34.8% | −38.8% | 0.07 |
| (M) TagPool | +50.4% | −24.8% | N/A | −31.4% | N/A | 0.06 |
| GeoJEPA-TI | −83.5% | +14.9% | −15.1% | −10.7% | +1.1% | 0.05 |
| TagAE | −78.3% | +14.3% | −15.7% | −12.6% | +4.4% | 0.05 |
| GeoJEPA-GTI | −66.5% | +13.4% | −11.5% | −8.7% | +5.3% | 0.05 |
| ViT-B/16 | 17.76 | 18.04 | 2.39 | 0.27 | 0.24 | 0.04 |
| ScaleMAE | −8.7% | +15.7% | −4.2% | −5.1% | +3.3% | 0.04 |
| EfficientNet | +81.4% | +113.2% | +31.0% | +24.7% | +65.3% | 0.03 |
| ResNet | +81.4% | +113.2% | +31.0% | +24.7% | +65.3% | 0.03 |

# Appendix E
# Qualitative Examples

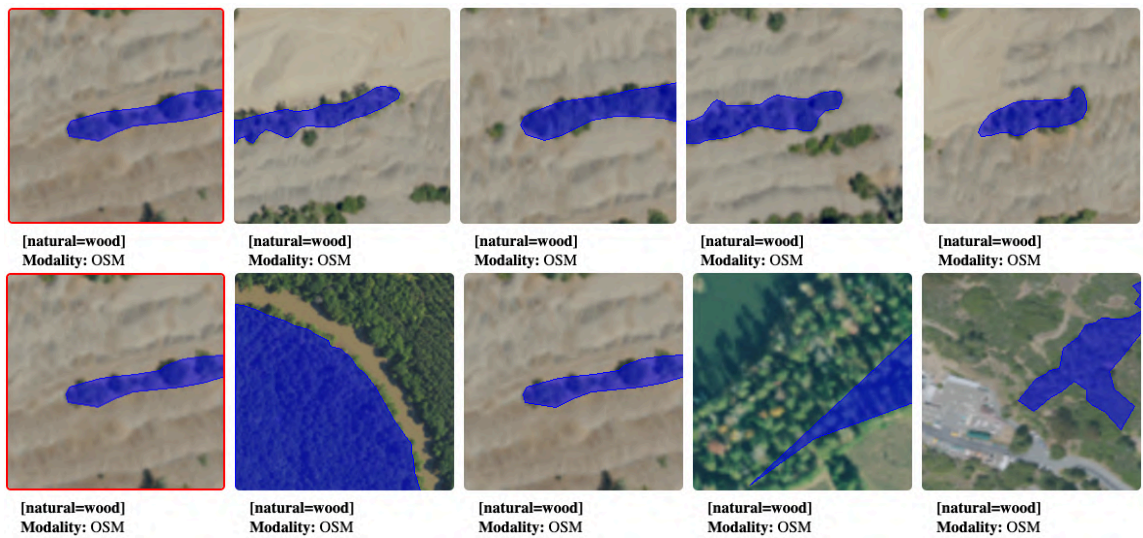

**Figure 50:** GeoJEPA-T (top) and TagPool (bottom). TagPool finds map entities with the same tags, but not similar size and context. The first four images on the top row are from the same tile, i.e. the same context, the last image on the top row is a neighbouring tile.

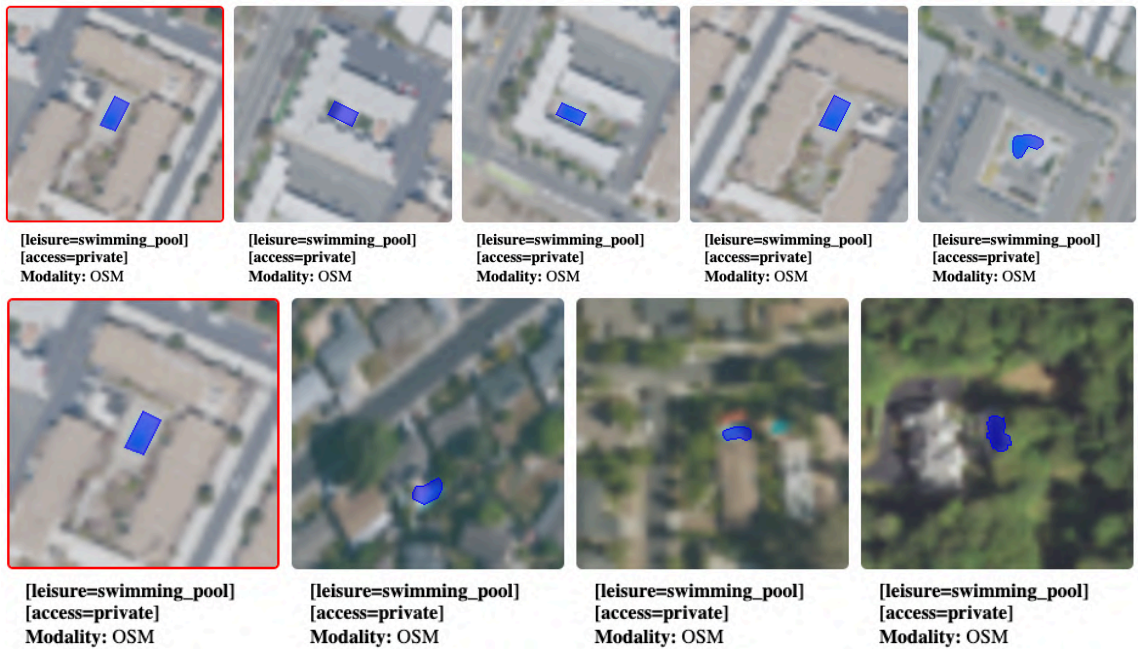

**Figure 51:** GeoJEPA-T (top) and TagPool (bottom). TagPool finds private swimming pools, but not in similar contexts. GeoJEPA again finds 3 out of 4 similar map entities in the same tile.

**Table 38:** kNN-search with query region featuring a taxiway on an airfield, and a field. All regions with anything related to airfields have a green checkmark.

| | | | |
|---|---|---|---|
| Query | 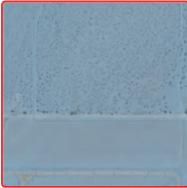 | 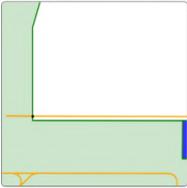 | 3x [PAD=PAD]<br>3x [aeroway=taxiway]<br>2x [name=]<br>2x [barrier=fence]<br>1x [barrier=gate]<br>and 16 more.. |
| TagPool | 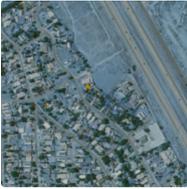 | 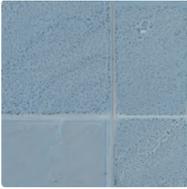 | 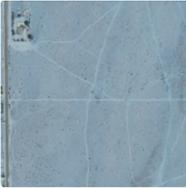 |
| ViT-B/16 | 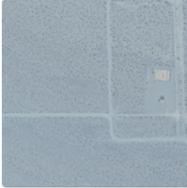 | 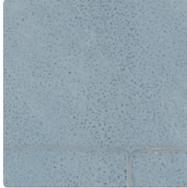 | 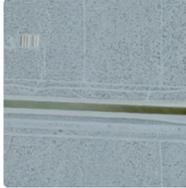 |
| GeoJEPA-T | 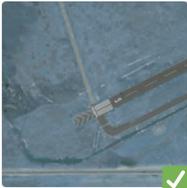 | 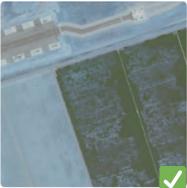 | 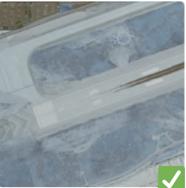 |
| GeoJEPA-TI | 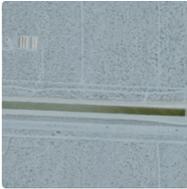 | 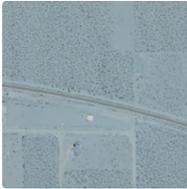 | 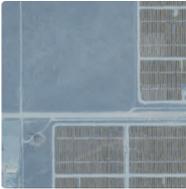 |
| GeoJEPA-GTI | 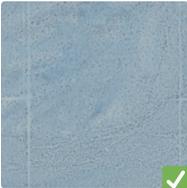 | 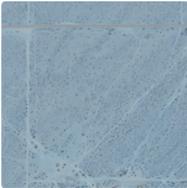 | 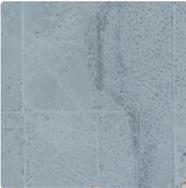 |

**Table 39:** kNN-search with query region featuring an airport area, listing the top-3 similar samples. All samples containing an airport area have a green checkbox, only GeoJEPA-T performs consistently over all eight neighbours.

Query 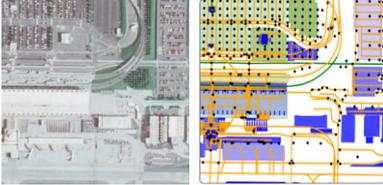

**201x [highway=street_lamp]**
**154x [level=]**
**141x [highway=footway]**
**123x [indoor=yes]**
**105x [highway=service]**
**and 144 more..**

TagPool 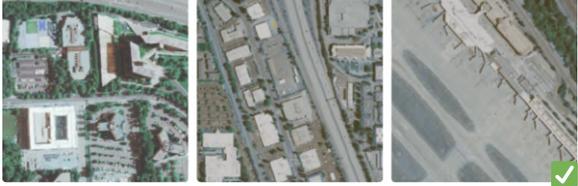

ViT-B/16 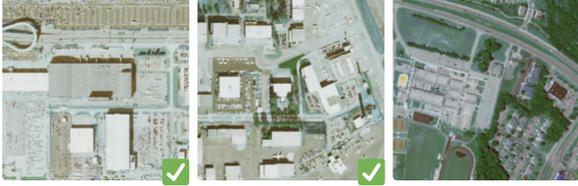

GeoJEPA-T 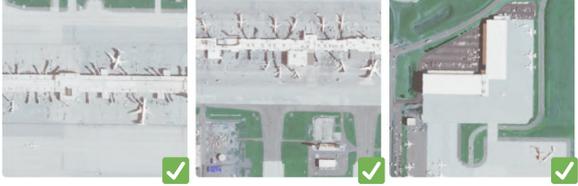

GeoJEPA-TI 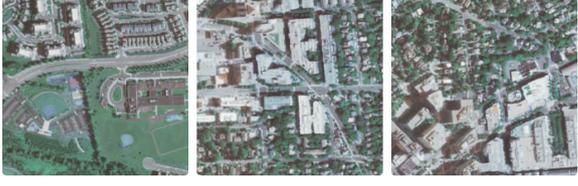

GeoJEPA-GTI 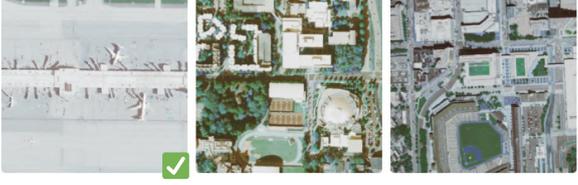

**Table 40:** kNN-search with the guery region feature a canal and fields. All tiles that feature a canal or large river have a green checkmark. GeoJEPA-GTI performs the best on this one but starts to fail after training until completion. Interestingly, GeoJEPA-T select regions featuring large roads.

| | | | | |
|---|---|---|---|---|
| Query | 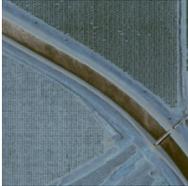 | 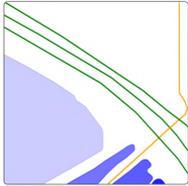 | 4x [nhd:com_id=]<br>4x [intermittent=yes]<br>4x [nhd:fdate=]<br>4x [nhd:reach_code=]<br>4x [landuse=orchard]<br>and 15 more.. | |
| TagPool | 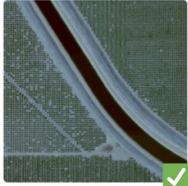 | 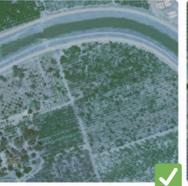 | 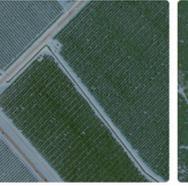 | 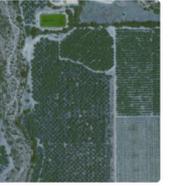 |
| ViT-B/16 | 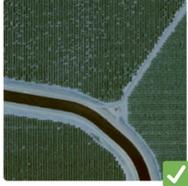 | 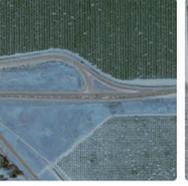 | 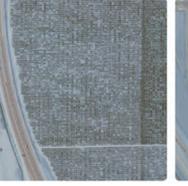 | 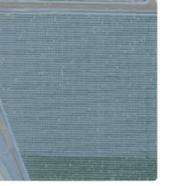 |
| GeoJEPA-T | 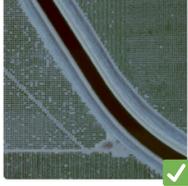 | 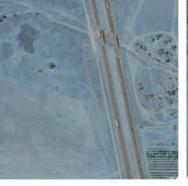 | 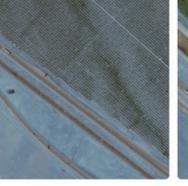 | 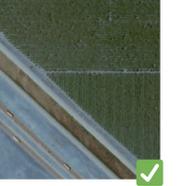 |
| GeoJEPA-TI | 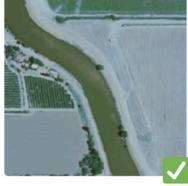 | 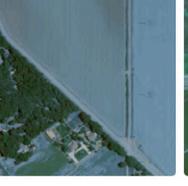 | 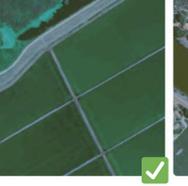 | 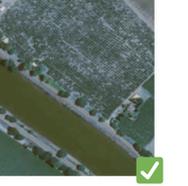 |
| GeoJEPA-GTI-170e | 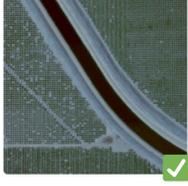 | 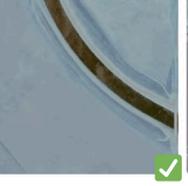 | 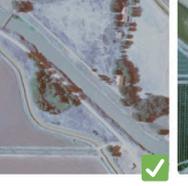 | 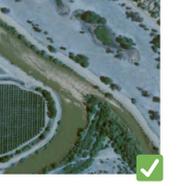 |
| TagformerLMAE | 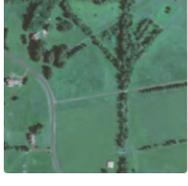 | 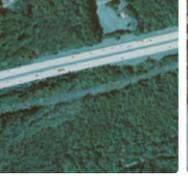 | 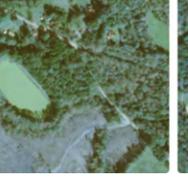 | 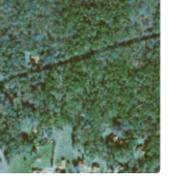 |
| GeoJEPA-GTI-300e | 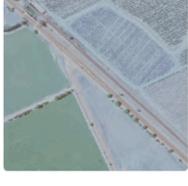 | 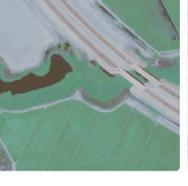 | 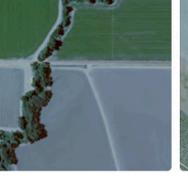 | 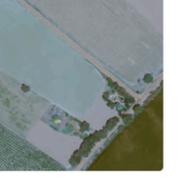 |

**Table 41:** kNN-search with query region featuring a motorway in a forest. All regions containing a motorway have a green checkmark. GeoJEPA-GTI performs the best.

| | | | |
|---|---|---|---|
| Query | 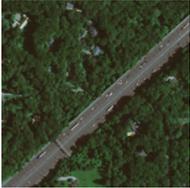 | 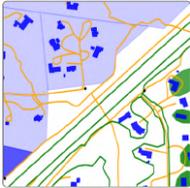 | **22x [addr:city=]**<br>**21x [highway=service]**<br>**21x [service=driveway]**<br>**21x [addr:postcode=]**<br>**21x [addr:housenumber=]**<br>**and 44 more..** |
| TagPool | 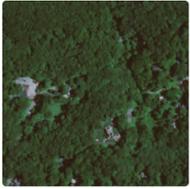 | 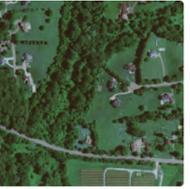 | 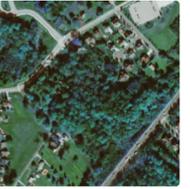 |
| ViT-B/16 | 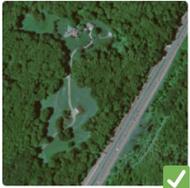 | 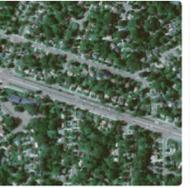 | 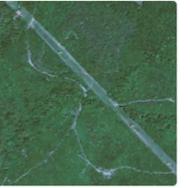 |
| GeoJEPA-T | 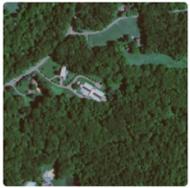 | 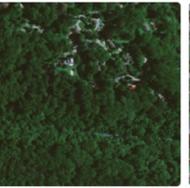 | 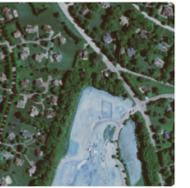 |
| GeoJEPA-TI | 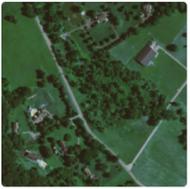 | 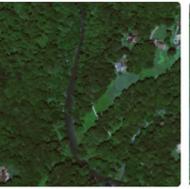 | 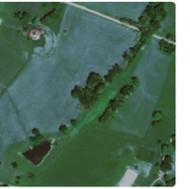 |
| GeoJEPA-GTI | 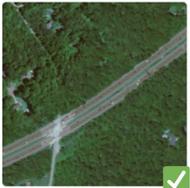 | 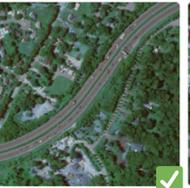 | 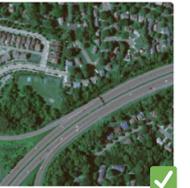 |

**Table 42:** kNN-search with query region featuring fields and a motorway. All samples with motorways have a green checkmark. It can be observed that all image models have difficulties in differentiating between wide dirt roads and a motorway.

| | | | |
|---|---|---|---|
| Query | 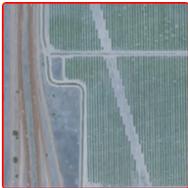 | 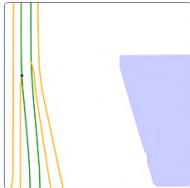 | **6x [lanes=]**<br>**6x [bicycle=no]**<br>**6x [oneway=yes]**<br>**4x [maxspeed=]**<br>**4x [highway=motorway]**<br>**and 12 more..** |
| TagPool | 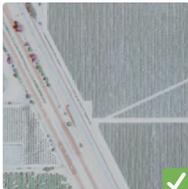 | 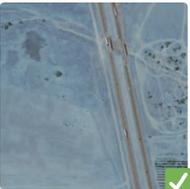 | 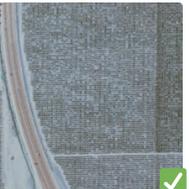 |
| ViT-B/16 | 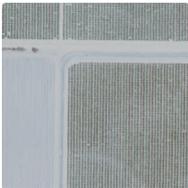 | 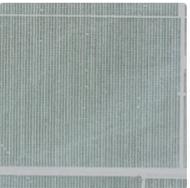 | 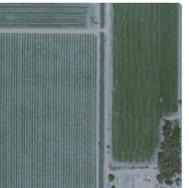 |
| GeoJEPA-T | 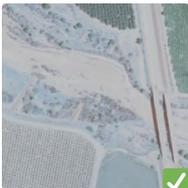 | 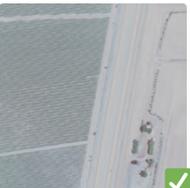 | 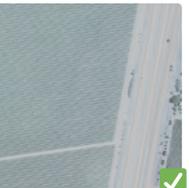 |
| GeoJEPA-TI | 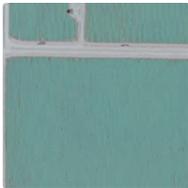 | 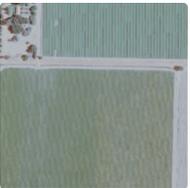 | 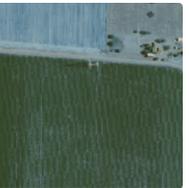 |
| GeoJEPA-GTI | 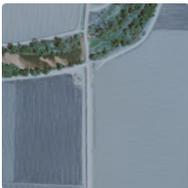 | 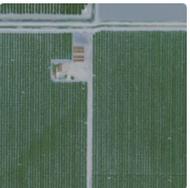 | 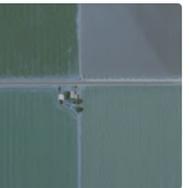 |

# EA  More kNN Queries

These are queries that have been excluded due to being hard to score. There is no single prevailing feature that we might use as an indicator of pass or fail.

**Table 43:** kNN-search over a shopping area with large parking lots. All tiles that prominently feature parking lots have a green checkmark. While most models perform well, looking more closely shows that GeoJEPA-TI and GeoJEPA-GTI perform most consistently across all images.

Query

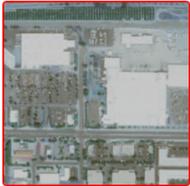 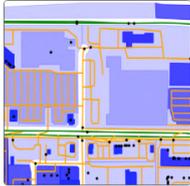

**80x [highway=service]**
**47x [addr:state=CA]**
**46x [addr:street=]**
**46x [addr:housenumber=]**
**44x [addr:city=]**
**and 102 more..**

TagPool

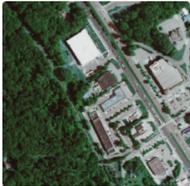 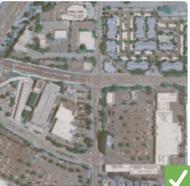 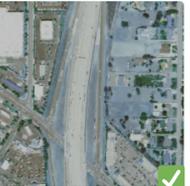

ViT-B/16

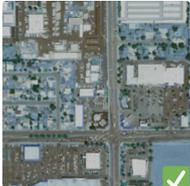 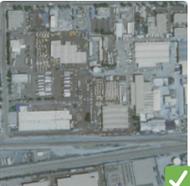 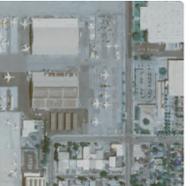

GeoJEPA-T

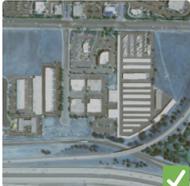 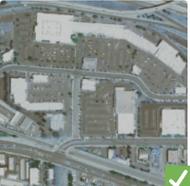 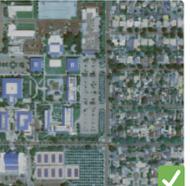

GeoJEPA-TI

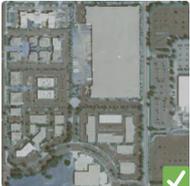 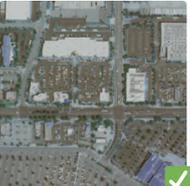 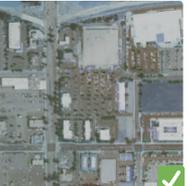

GeoJEPA-GTI

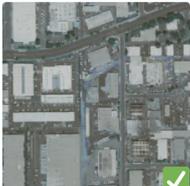 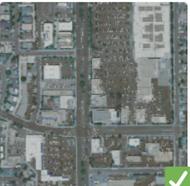 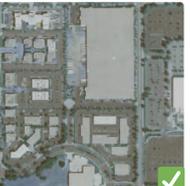

**Table 44:** kNN-search over a crossing area. Similar samples containing crossings have been marked with a green checkmark. All models passed this test. The query was excluded from the evaluation due to being hard to score objectively.

Query

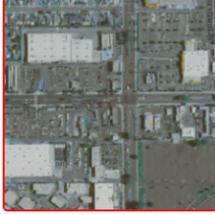 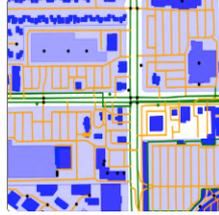

**123x** [highway=service]
**87x** [oneway=yes]
**70x** [addr:postcode=]
**70x** [addr:state=CA]
**69x** [addr:street=]
and 101 more..

TagPool

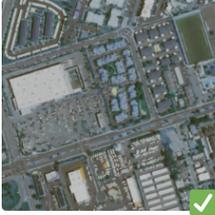 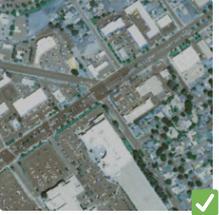 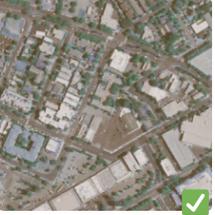

ViTB-16

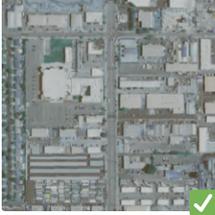 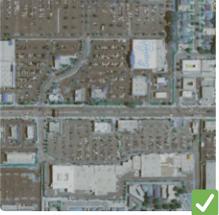 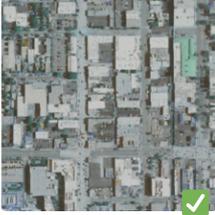

GeoJEPA-T

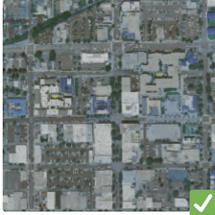 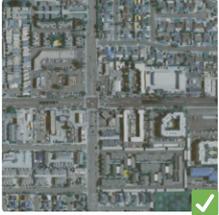 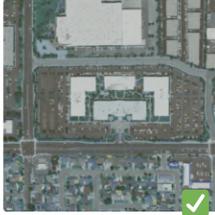

GeoJEPA-TI

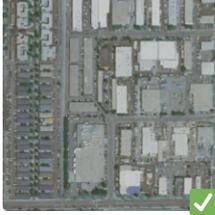 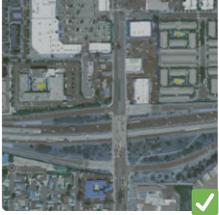 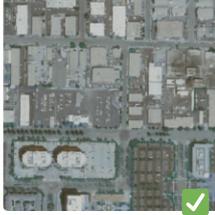

GeoJEPA-GTI

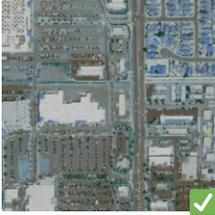 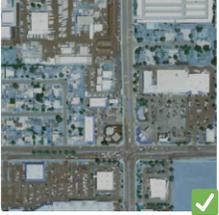 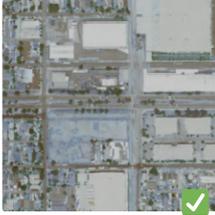

**Table 45:** kNN-search over a wooden area with a moderately sized lake and streams. All samples with lakes have a green checkmark.

| | | |
|---|---|---|
| Query | 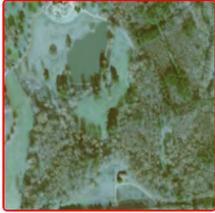 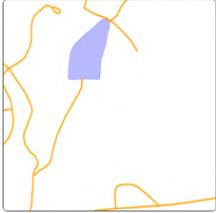 | **4x [attribution=]**<br>**4x [waterway=stream]**<br>**4x [highway=service]**<br>**3x [name=]**<br>**3x [access=private]**<br>**and 12 more..** |
| TagPool | 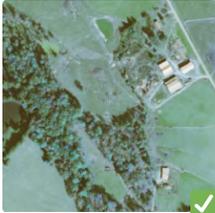 | 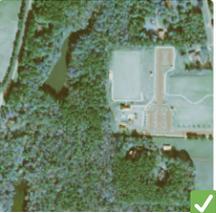 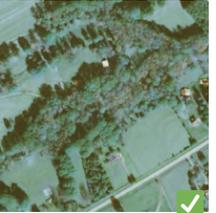 |
| ViT-B/16 | 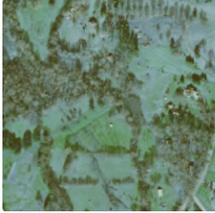 | 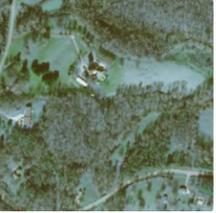 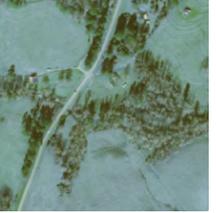 |
| GeoJEPA-T | 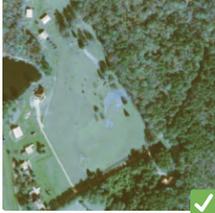 | 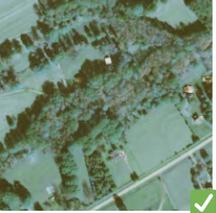 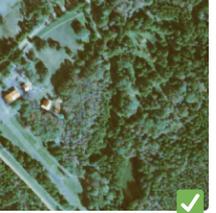 |
| GeoJEPA-TI | 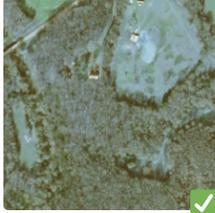 | 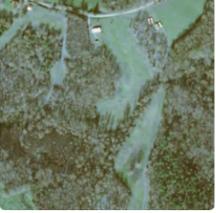 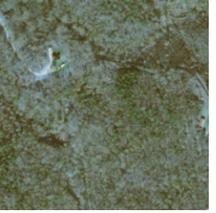 |
| GeoJEPA-GTI | 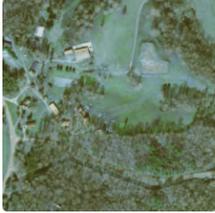 | 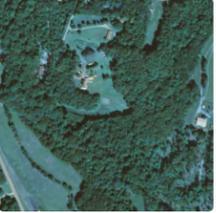 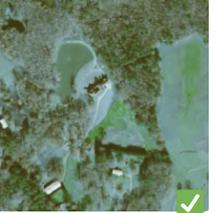 |

**Table 46:** kNN-search over a industrial area.

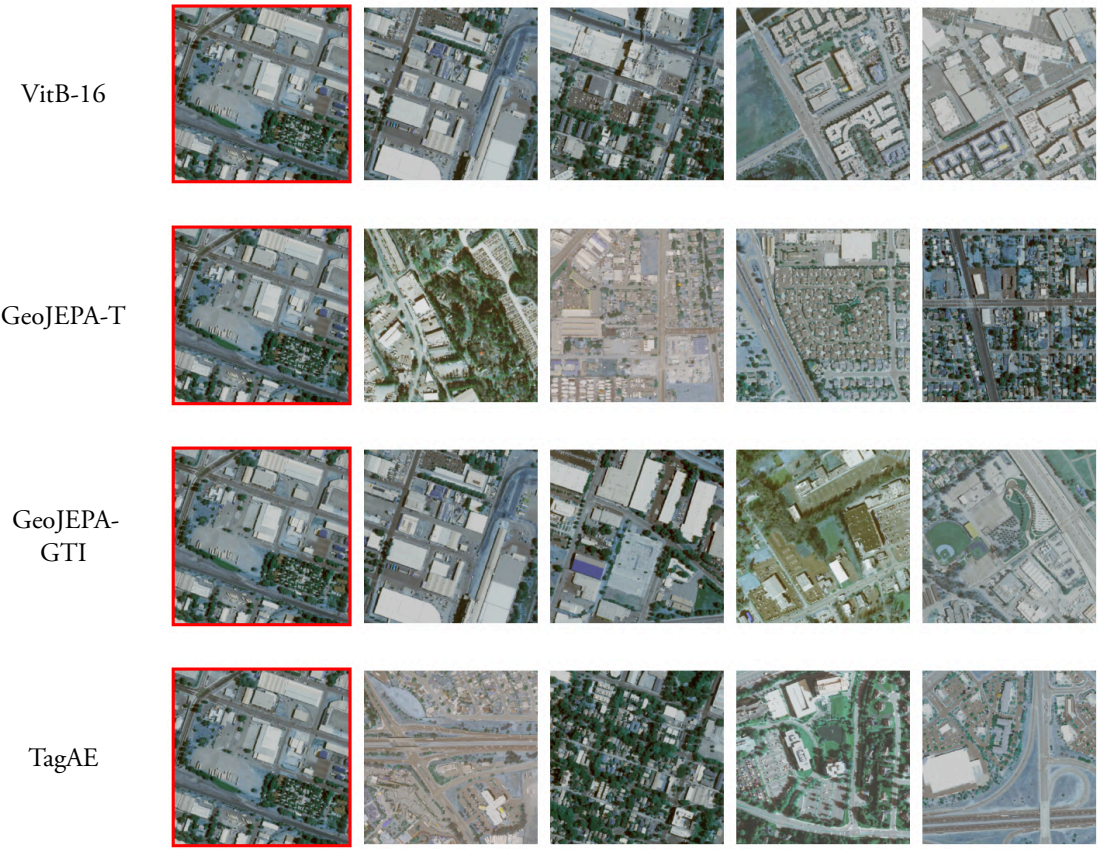

**Table 47:** kNN-search over a residential area.

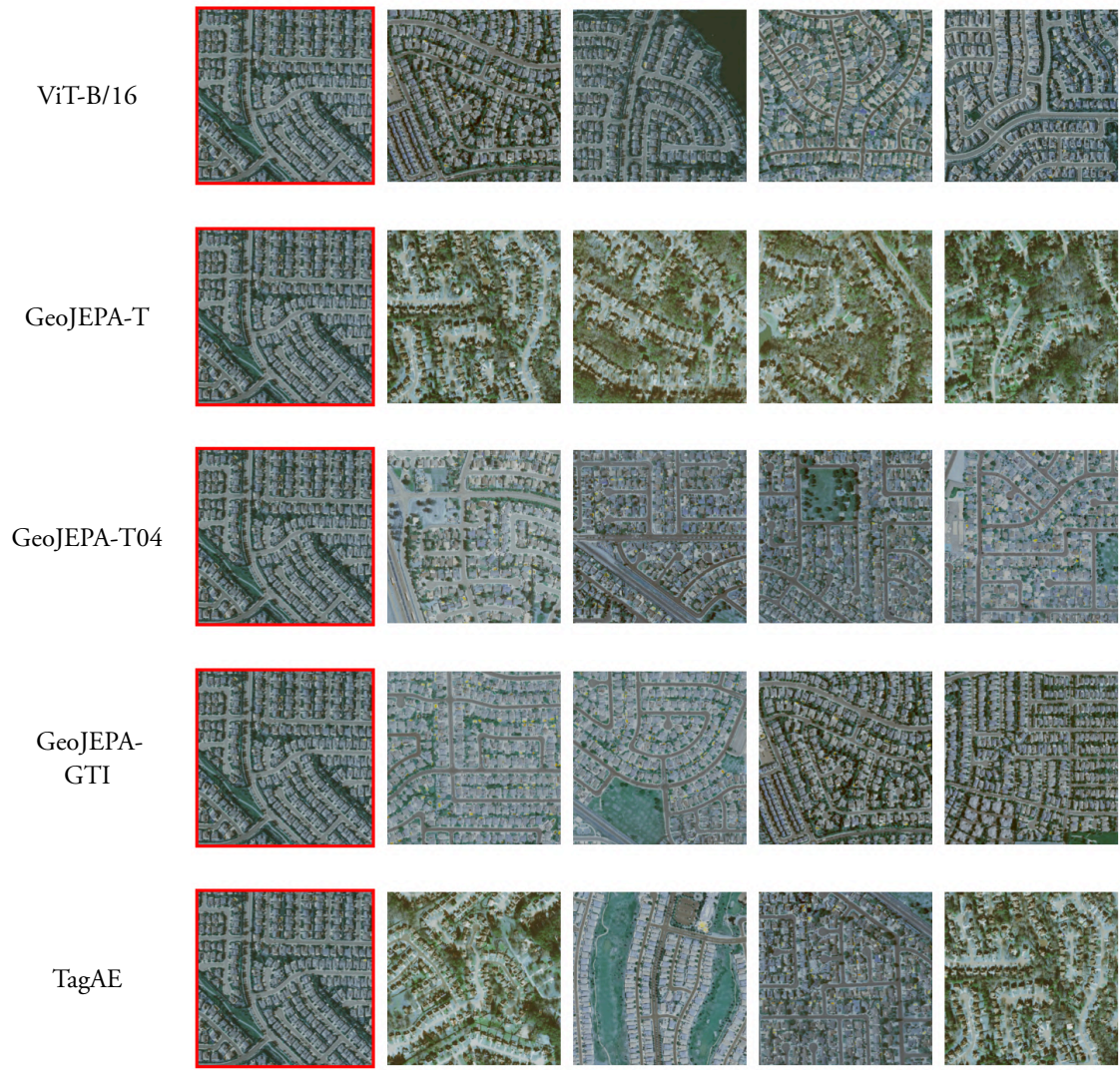

**Table 48:** kNN-search over a city area.

ViT-B/16

GeoJEPA-T

GeoJEPA-T04

GeoJEPA-GTI

TagAE

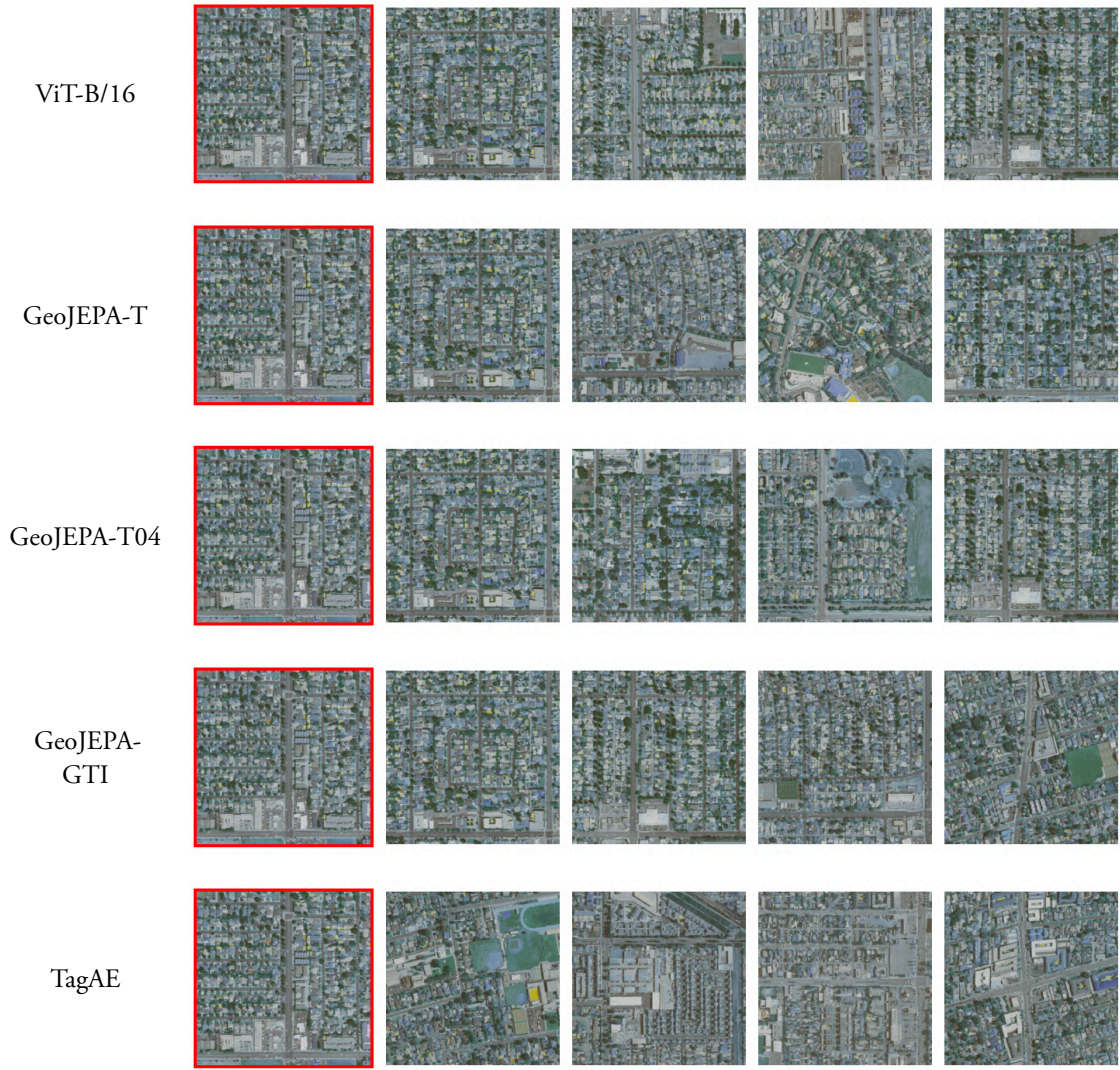

# Att hitta liknande platser och kartföremål

POPULÄRVETENSKAPLIG SAMMANFATTNING  **Theodor Lundqvist, Ludvig Delvret**

Föreställ dig att du äger en resebyrå som säljer resor till exotiska platser, och utforskar var du ska bygga din nästa resort. Du vet att vissa resmål är mer populära hos resenärerna, t.ex. tack vare närhet till stranden, fina restauranger, sevärdheter, och möjlighet till utflykter. Hur ska du hitta liknande resmål, samtidigt som du undviker att kolla igenom varje stad, en efter en?

Digitala kartor spelar en central roll i våra moderna liv, men att representera världen digitalt innebär unika utmaningar. Komplex data, såsom geometrier, bilder, text och trafikinformation måste samverka effektivt. För att kunna leverera avancerade karttjänster krävs algoritmer som inte bara kan hantera denna enorma och olikartade datamängd, utan även förstå dess komplexa beroenden och dra meningsfulla slutsatser.

Vår modell hanterar text, geometrier och relationer från kartdata samt flygbilder. Modellen baseras på en metod för självinlärning (JEPA) som introducerades av Yann LeCun under slutet av 2022. Sedan dess har endast ett fåtal publikationer gjorts i området och enbart två anpassningar kan hantera mer än en datatyp. Vidare är vår modell, GeoJEPA, den första som är utvecklad för kartdata. Modellen skapar konceptuella representationer av områden och föremål i världen, vilka kan användas i rekommendationsalgoritmer eller som underlag för att förutsäga faktorer som bostadspriser, folktäthet, hastighet, eller brottsfrekvens.

Vi utvärderar GeoJEPA på syntetiska uppgifter skapade från kartdata och jämför dess prestanda med våra egna enklare modeller. Resultaten visar att GeoJEPA är särskilt bra på att identifiera liknande föremål och områden, men har svårigheter med att bevara detaljerad information. Dessutom observerar vi, i motsats till vår ursprungliga hypotes, att GeoJEPA presterar sämre ju fler datatyper den hanterar. Detta är särskilt tydligt vid användning av bilddata, vilket vi förklarar med att modellen som bearbetar bilddatan innan den ges till GeoJEPA läcker global information. Sådana representationer verkar inte lämpa sig i JEPA:s träningsmodell, vilket leder till försämrad prestanda. Vi noterar också att vi saknar inblick i, och inte kan styra, hur mycket en representation av ett föremål påverkas av dess omgivning. I flera fall verkar GeoJEPA lägga större vikt vid omgivningen än själva föremålet. Trots detta ser vi JEPA som ett mycket lovande alternativ för geospatial inlärning och vill uppmuntra vidare forskning inom området.

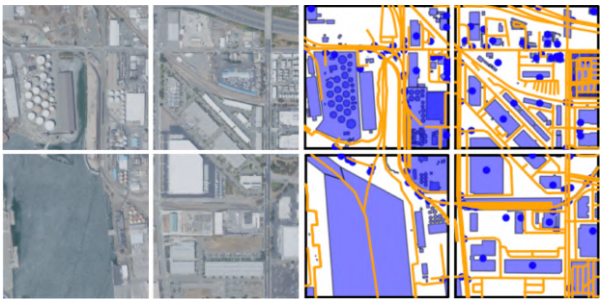

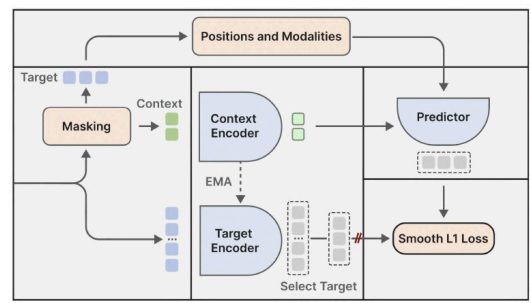